\documentclass{article}

\usepackage[accepted]{icml2022}

\usepackage{microtype}
\usepackage{graphicx}
\usepackage{subcaption}
\usepackage{booktabs}

\usepackage{hyperref}

\usepackage{graphicx}
\usepackage[utf8]{inputenc} 
\usepackage[T1]{fontenc}    
\usepackage{url}            
\usepackage{booktabs}       
\usepackage{amsfonts}       
\usepackage{nicefrac}       
\usepackage{url}            
\usepackage{booktabs}       
\usepackage{amsfonts}       
\usepackage{nicefrac}       
\usepackage{microtype}      
\usepackage{nccmath}
\usepackage{times}
\usepackage{graphicx}
\usepackage{natbib}
\usepackage{algorithm}
\usepackage{algorithmic}
\usepackage{amsfonts}
\usepackage{amsmath}

\usepackage{color-edits}
\addauthor{jm}{purple}
\addauthor{pa}{green}
\addauthor{mm}{orange}
\addauthor{cc}{red}
\usepackage{mathtools}
\usepackage{amsthm}
\usepackage{graphicx}
\usepackage{enumerate}
\usepackage{algorithm}
\usepackage{algorithmic}
\usepackage{thm-restate}

\usepackage{url}
\usepackage{dsfont}
\usepackage[mathscr]{euscript}
\usepackage{booktabs}
\usepackage{makecell}
\usepackage{tabularx}
\usepackage{xcolor, colortbl}
\usepackage[normalem]{ulem}
\usepackage{soul}
\usepackage{prettyref}

\usepackage{MnSymbol}
\DeclareMathAlphabet\mathbb{U}{msb}{m}{n}
\usepackage{xpatch}

\def\Nset{\mathbb{N}}
\def\Rset{\mathbb{R}}

\DeclareMathOperator*{\E}{\mathbb E}

\DeclareMathOperator{\sign}{sign}

\DeclarePairedDelimiter{\abs}{\lvert}{\rvert} 
\DeclarePairedDelimiter{\bracket}{[}{]}
\DeclarePairedDelimiter{\curl}{\{}{\}}
\DeclarePairedDelimiter{\norm}{\|}{\|}
\DeclarePairedDelimiter{\paren}{(}{)}
\DeclarePairedDelimiter{\tri}{\langle}{\rangle}

\newtheorem{theorem}{Theorem}

\newtheorem{corollary}{Corollary}

\newtheorem{definition}{Definition}
\newtheorem*{remark*}{Remark}

\newcommand{\cX}{\mathcal{X}}
\newcommand{\cY}{\mathcal{Y}}

\newcommand{\sC}{{\mathscr C}}
\newcommand{\sD}{{\mathscr D}}

\newcommand{\sH}{{\mathscr H}}

\newcommand{\sM}{{\mathscr M}}
\newcommand{\sP}{{\mathscr P}}

\newcommand{\sR}{{\mathscr R}}

\newcommand{\sT}{{\mathscr T}}

\newcommand{\sX}{{\mathscr X}}
\newcommand{\sY}{{\mathscr Y}}

\newcommand{\h}{\widehat}
\newcommand{\ov}{\overline}
\newcommand{\uv}{\underline}
\newcommand{\wt}{\widetilde}
\newcommand{\e}{\epsilon}

\newcommand{\ignore}[1]{}

\hypersetup{
  colorlinks   = true,
  urlcolor     = blue,
  linkcolor    = blue,
  citecolor    = blue
}

\usepackage[textsize=tiny]{todonotes}

\usepackage[toc, page, header]{appendix}
\icmltitlerunning{H-Consistency Estimation Error of 
Surrogate Loss Minimizers}

\begin{document}

\twocolumn[
\icmltitle{$\sH$-Consistency Estimation Error of 
Surrogate Loss Minimizers}

\begin{icmlauthorlist}
\icmlauthor{Pranjal Awasthi}{google}
\icmlauthor{Anqi Mao}{courant}
\icmlauthor{Mehryar Mohri}{google,courant}
\icmlauthor{Yutao Zhong}{courant}
\end{icmlauthorlist}

\icmlaffiliation{google}{Google Research, New York, NY;}
\icmlaffiliation{courant}{Courant Institute of Mathematical Sciences, 
New York, NY}

\icmlcorrespondingauthor{Anqi Mao}{aqmao@cims.nyu.edu}
\icmlcorrespondingauthor{Yutao Zhong}{yutao@cims.nyu.edu}

\icmlkeywords{consistency, calibration, adversarial loss}

\vskip 0.3in
]
\printAffiliationsAndNotice{}

\begin{abstract}
We present a detailed study of estimation errors \ignore{the
  estimation error with respect to the loss tailored to a task,} in
terms of surrogate loss estimation errors.  We refer to such
guarantees as \emph{$\sH$-consistency estimation error bounds}, since
they account for the hypothesis set $\sH$ adopted. These guarantees
are significantly stronger than $\sH$-calibration or
$\sH$-consistency\ignore{ presented in previous work}. They are also
more informative than similar \emph{excess error bounds} derived in
the literature, \ignore{that is the special case where } when $\sH$ is
the family of all measurable functions.
We prove general theorems providing such guarantees, for both the
distribution-dependent and distribution-independent settings.  We show
that our bounds are tight, modulo a convexity assumption. We also show
that previous excess error bounds can be recovered as special cases of
our general results.

We then present a series of explicit bounds in the case of the
zero-one loss, with multiple choices of the surrogate loss and for
both the family of linear functions and neural networks with one
hidden-layer. We further prove more favorable distribution-dependent
guarantees in that case. We also present a series of explicit bounds
in the case of the adversarial loss, with surrogate losses based on
the supremum of the $\rho$-margin, hinge or sigmoid loss and for the
same two general hypothesis sets. Here too, we prove several
enhancements of these guarantees under natural distributional
assumptions.  Finally, we report the results of simulations
illustrating our bounds and their tightness.

\end{abstract}

\section{Introduction}
\label{sec:intro}

Most learning algorithms rely on optimizing a surrogate loss function
distinct from the \emph{target loss function} tailored to the task
considered. This is typically because the target loss function is
computationally hard to optimize or because it does not admit
favorable properties, such as differentiability or smoothness, crucial
to the convergence of optimization algorithms. But, what guarantees
can we count on for the target loss estimation error, when minimizing
a surrogate loss estimation error?

A desirable property of a surrogate loss function, often referred to
in that context is \emph{Bayes-consistency}. It requires that
asymptotically, near optimal minimizers of the surrogate excess
error also near optimally minimize the target excess error
\citep{steinwart2007compare}. This property holds for a broad family
of convex surrogate losses of the standard binary and
multi-class classification losses
\citep{Zhang2003,bartlett2006convexity,tewari2007consistency,
  steinwart2007compare}.
But, Bayes-consistency is not relevant when learning with a hypothesis
set $\sH$ distinct from the family of all measurable
functions. Instead, the hypothesis-set dependent notion of
$\sH$-consistency should be adopted, as argued by
\citet{long2013consistency} (see also \citep{zhang2020bayes}). Some
recent publications \citep{awasthi2021calibration,bao2020calibrated}
further study $\sH$-consistency guarantees for the \emph{adversarial
loss} \citep{goodfellow2014explaining,madry2017towards,
  tsipras2018robustness,carlini2017towards}.  Nevertheless,
consistency and $\sH$-consistency are both asymptotic properties and 
thus do not provide any guarantee for approximate minimizers learned
from finite samples.

Instead, we will consider upper bounds on the target estimation error
expressed in terms of the surrogate estimation error, which we refer
to as \emph{$\sH$-consistency estimation error bounds}, since they
account for the hypothesis set $\sH$ adopted. These guarantees are
significantly stronger than $\sH$-calibration or $\sH$-consistency
(Section~\ref{sec:adv})\ignore{ bounds presented in previous work} or
some margin-based properties of convex surrogate losses for linear
predictors studied by \citet{ben2012minimizing} and
\citet{long2011learning}. They are also more informative than similar
\emph{excess error bounds} derived in the literature, which correspond
to the special case where $\sH$ is the family of all measurable
functions \citep{Zhang2003,bartlett2006convexity} (see also
\cite{MohriRostamizadehTalwalkar2018}[section~4.7]).
We prove general theorems providing such guarantees, which could be
used in both distribution-dependent and distribution-independent
settings (Section~\ref{sec:general}).  We show that our bounds are
tight, modulo a convexity assumption
(Section~\ref{sec:non-adv-general} and \ref{sec:adv-general}). We also
show that previous excess error bounds can be recovered as special
cases of our general results (Section~\ref{sec:all}).

We then present a series of explicit bounds in the case of the $0/1$
loss (Section~\ref{sec:non-adv}), with multiple choices of the
surrogate loss and for both the family of linear functions
(Section~\ref{sec:non-adv-lin}) and neural networks with one
hidden-layer (Section~\ref{sec:non-adv-NN}). We further prove more
favorable distribution-dependent guarantees in that case
(Section~\ref{sec:noise-non-adv}).

We also present a detailed analysis of the \emph{adversarial loss}
(Section~\ref{sec:adv}). We show that there can be no non-trivial
adversarial $\sH$-consistency estimation error bound for
supremum-based convex loss functions and supremum-based sigmoid loss
function, under mild assumptions that hold for most hypothesis sets
used in practice (Section~\ref{sec:negative}). These results imply
that the loss functions commonly used in practice for optimizing the
adversarial loss cannot benefit from any useful $\sH$-consistency
estimation error guarantee! These are novel results that go beyond the
negative ones given for convex surrogates by
\citet{awasthi2021calibration}.

We present new $\sH$-consistency estimation error bounds for the
adversarial loss with surrogate losses based on the supremum of the
$\rho$-margin loss, for linear hypothesis sets
(Section~\ref{sec:adv-lin}) and the family of neural networks with one
hidden-layer (Section~\ref{sec:adv-NN}). Here too, we prove several
enhancements of these guarantees under some natural distributional
assumptions (Section~\ref{sec:noise-adv}).

Our results help compare different surrogate loss functions of the zero-one loss or adversarial loss, given the specific hypothesis set
used, based on the functional form of their $\sH$-consistency
estimation error. These results, combined with approximation error
properties of surrogate losses, can help select the most suitable
surrogate loss in practice.
In addition to several general theorems, our study required a careful
inspection of the properties of various surrogate loss functions and
hypothesis sets. Our proofs and techniques could be adopted for the
analysis of many other surrogate loss functions and hypothesis sets.

In Section~\ref{sec:simulations}, we report the results of simulations
illustrating our bounds and their tightness. We give a detailed
discussion of related work in Appendix~\ref{app:related}. We start
with some preliminary definitions and notation.

\section{Preliminaries}

Let $\sX$ denote the input space and $\sY = \curl*{-1,+1}$ the binary
label space.
We will denote by $\sD$ a distribution over $\sX \times \sY$, by $\sP$
a set of such distributions and by $\sH$ a hypothesis set of functions
mapping from $\sX$ to $\Rset$. The \emph{generalization error} and
\emph{minimal generalization error} for a loss function $\ell(h,x,y)$
are defined as $\sR_{\ell}(h)=\E_{(x,y)\sim
  \sD}\bracket*{\ell(h,x,y)}$ and $\sR_{\ell,\sH}^* = \inf_{h\in
  \sH}\sR_{\ell}(h)$.  Let $\sH_{\mathrm{all}}$ denote the hypothesis
set of all measurable functions.  The \emph{excess error} of a
hypothesis $h$ is defined as the difference
$\sR_{\ell}(h)-\sR_{\ell,\sH_{\mathrm{all}}}^*$, which can be
decomposed into the sum of two terms, the \emph{estimation error} and
\emph{approximation error}:
\begin{equation}
\label{eq:excess-split}
    \sR_{\ell}(h)-\sR_{\ell,\sH_{\mathrm{all}}}^*
    = \paren*{\sR_{\ell}(h)-\sR_{\ell,\sH}^*}+ \paren*{\sR_{\ell,\sH}^* - \sR_{\ell,\sH_{\mathrm{all}}}^*}.
\mspace{-1mu}
\end{equation}
Given two loss functions $\ell_1$ and $\ell_2$, a fundamental question is whether $\ell_1$ is \emph{consistent} with respect to $\ell_2$ for a hypothesis set $\sH$ and a set of distributions $\sP$ \citep{bartlett2006convexity,steinwart2007compare,long2013consistency,bao2020calibrated,awasthi2021calibration}.
\begin{definition}[\textbf{$(\sP,\sH)$-consistency}]
\ignore{
Let $\sP$ be a set of distributions over $\sX\times\sY$. 
}
We say that $\ell_1$ is \emph{$(\sP,\sH)$-consistent} with respect to $\ell_2$, if for all distributions $\sD\in \sP$ and sequences of $\{h_n\}_{n\in \Nset}\subset \sH$ we have 
\begin{equation}
\label{eq:consistency}
    \mspace{-6mu}
    \lim_{n \to +\infty}
    \mspace{-4mu}
    \sR_{\ell_1}(h_n)-\sR_{\ell_1,\sH}^* = 0
    \Rightarrow 
    \mspace{-10mu}
    \lim_{n \to +\infty}
    \mspace{-4mu}
    \sR_{\ell_2}(h_n)-\sR_{\ell_2,\sH}^* = 0.
    \mspace{-12mu}
\end{equation}
\end{definition}
\ignore{ $\sP$ can be chosen to be any set of interest, such as the
  set of distributions that are
  \emph{realizable}~\citep{long2013consistency,zhang2020bayes,awasthi2021calibration},
  or more generally, the set of distributions that verify certain low
  noise conditions. When $\sP$ consists of a set of distributions all
  supported on a singleton, $(\sP,\sH)$-Consistency reduces to
  \emph{calibration}~\citep{bartlett2006convexity,steinwart2007compare,awasthi2021calibration}.} 
\vspace{-0.3pt}
We will denote by $\Phi$ the margin-based loss and
$\wt{\Phi}\colon=\sup_{x'\colon \|x-x'\|_p\leq \gamma}\Phi\paren*{y
  h(x')}$, $p\in [1,\plus\infty]$ the supremum-based counterpart.  In
the standard binary classification, $\ell_2$ is the $0/1$ loss
$\ell_{0-1}\colon=\mathds{1}_{\sign(h(x))\neq y}$, where
$\sign(\alpha) = \mathds{1}_{\alpha \geq 0} - \mathds{1}_{\alpha < 0}$
and $\ell_1$ is the margin-based loss for some function $\Phi\colon
\Rset \to \Rset_{+}$, typically convex. In the adversarial binary
classification, $\ell_2$ is the adversarial $0/1$ loss
$\ell_{\gamma}\colon=\sup_{x'\colon \|x-x'\|_p\leq
  \gamma}\mathds{1}_{y h(x') \leq 0}$, for some $\gamma \in (0, 1)$
and $\ell_1$ is the supremum-based margin loss $\wt{\Phi}$.

Let $B_p^d(r)$ denote the $d$-dimensional $\ell_p$-ball with radius
$r$: $B_p^d(r) = \big\{z \in \Rset^d \mid \norm*{z}_p\leq r\big\}$.
\ignore{where $\norm*{z}_p\colon =\bracket*{\sum_{i=1}^d
    \abs*{z_i}^p}^{\frac{1}{p}}$, $p\in [1, +\infty]$.}
Without loss of generality, we consider $\sX=B_p^d(1)$\ignore{ and the perturbation size $\gamma \in (0,1)$ throughout the paper}. 
Let $p, q \in[1, \plus\infty]$ be conjugate numbers, that is
$\frac{1}{p} + \frac{1}{q} = 1$. We will specifically study the family
of linear hypotheses $\sH_{\mathrm{lin}}=\big\{x\mapsto w \cdot x + b
\mid \norm*{w}_q\leq W,\abs*{b}\leq B\big\}$ and one-hidden-layer
ReLU networks $\sH_{\mathrm{NN}} = \big\{x\mapsto \sum_{j =
  1}^n u_j(w_j \cdot x+b)_{+} \mid \|u \|_{1}\leq
\Lambda,\|w_j\|_q\leq W, \abs*{b}\leq B\big\}$, where $(\cdot)_+ =
\max(\cdot,0)$.
Finally, for any $\e > 0$, we will denote by $\tri*{t}_{\e}$ the
$\e$-truncation of $t \in \Rset$ defined by $t\mathds{1}_{t>\e}$.

\section{$\sH$-consistency estimation error definitions}

$(\sP,\sH)$-Consistency is an asymptotic relation between two loss
functions. However, we are interested in a more quantitative relation
in many applications. This motivates the study of
\emph{$\sH$-consistency estimation error bound}.

\begin{definition}[\textbf{$\sH$-consistency estimation error bound}]
\ignore{Let $\sP$ be a set of distributions over $\sX\times\sY$.}  If
for some function $f\colon \Rset_{+}\to \Rset_{+}$, a bound of the
following form holds for all $h\in \sH$ and $\sD\in \sP$:
\begin{align}
\label{eq:est-bound}
    \sR_{\ell_2}(h) - \sR_{\ell_2,\sH}^*
    \leq f\paren*{\sR_{\ell_1}(h)-\sR_{\ell_1,\sH}^*},
\end{align}
then we call it an \emph{$\sH$-consistency estimation error
bound}. Furthermore, if $\sP$ consists of all distributions over
$\sX\times\sY$, we say that the bound is
\emph{distribution-independent}.
\end{definition}
When $\sH=\sH_{\mathrm{all}}$ and $\sP$ is the set of all
distributions, a bound of the form \eqref{eq:est-bound} is also called
a \emph{consistency excess error bound}. Note when $f(0)= 0$ and $f$
is continuous at $0$, the $\sH$-consistency bound \eqref{eq:est-bound}
implies $\sH$-consistency \eqref{eq:consistency}. Thus,
$\sH$-consistency estimation error bounds provide stronger results
than consistency and calibration. Furthermore, there is a fundamental
reason to study such bounds from the statistical learning point of
view: they can be turned into more favorable generalization bounds for
the target loss $\ell_2$ than the excess error bound. For example,
when $\sP$ is the set of all distributions, by
\eqref{eq:excess-split}, relation \eqref{eq:est-bound} implies that, for all $h\in \sH$,
\begin{equation}
\label{eq:est-gen}
    \mspace{-6mu}
    \sR_{\ell_2}
    \mspace{-3mu}
    (h)
    \mspace{-1mu}
    -
    \mspace{-1mu}
    \sR_{\ell_2,\sH_{\mathrm{all}}}^*
    \mspace{-4mu}
    \leq
    \mspace{-4mu}
    f\paren*{\sR_{\ell_1}
    \mspace{-3mu}
    (h)
    \mspace{-1mu}
    -
    \mspace{-1mu}
    \sR_{\ell_1,\sH}^*}
    \mspace{-1mu}
    +
    \mspace{-1mu}
    \sR_{\ell_2,\sH}^*
    \mspace{-1mu}
    -
    \mspace{-1mu}
    \sR_{\ell_2,\sH_{\mathrm{all}}}^*.
    \mspace{-15mu}
\end{equation}
Similarly, the excess error bound can be
written as follows:
\begin{equation}
\label{eq:gen}
    \mspace{-6mu}
    \sR_{\ell_2}
    \mspace{-3mu}
    (h)
    \mspace{-1mu}
    -
    \mspace{-1mu}
    \sR_{\ell_2,\sH_{\mathrm{all}}}^*
    \mspace{-4mu}
    \leq
    \mspace{-4mu}
     f\paren*{\sR_{\ell_1}
     \mspace{-3mu}
     (h)
     \mspace{-1mu}
     -
     \mspace{-1mu}
     \sR_{\ell_1,\sH}^*
     \mspace{-1mu}
     +
     \mspace{-1mu}
     \sR_{\ell_l,\sH}^*
     \mspace{-1mu}
     -
     \mspace{-1mu}
     \sR_{\ell_l,\sH_{\mathrm{all}}}^*}. 
     \mspace{-18mu}
\end{equation}
If we further bound the estimation error $[\sR_{\ell_1}(h) -
  \sR_{\ell_1,\sH}^*]$ by the empirical error plus a complexity term,
\eqref{eq:est-gen} and \eqref{eq:gen} both turn into generalization
bounds. However, the generalization bound obtained by
\eqref{eq:est-gen} is linearly dependent on the approximation error of
target loss $\ell_2$, while the one obtained by \eqref{eq:gen} depends
on the approximation error of the surrogate loss $\ell_1$ and can
potentially be worse than linear dependence.
Moreover, \eqref{eq:est-gen} can be easily used to compare different
surrogates by directly comparing the corresponding mapping
$f$. However, only comparing the mapping $f$ for different surrogates
in \eqref{eq:gen} is not sufficient since the approximation errors of
surrogates may differ as well.

\paragraph{Minimizability gap.} We will adopt the standard notation
for the conditional distribution of $Y$ given $X = x$: $\eta(x) =
\sD(Y = 1 \!\mid\! X = x)$ and will also use the shorthand $\Delta
\eta(x) = \eta(x) - \frac{1}{2}$. It is useful to write the
generalization error as
$\sR_{\ell}(h)=\mathbb{E}_{X}\bracket*{\sC_{\ell}(h,x)}$, where
$\sC_{\ell}(h,x)$ is the \emph{conditional $\ell$-risk} defined by
$\sC_{\ell}(h,x) = \eta(x)\ell(h, x, +1) + (1 - \eta(x))\ell(h, x,
-1)$.  The \emph{minimal conditional $\ell$-risk} is denoted by
$\sC_{\ell,\sH}^*(x) = \inf_{h\in \sH}\sC_{\ell}(h,x)$. We also use
the following shorthand for the gap $\Delta\sC_{\ell,\sH}(h,x) =
\sC_{\ell}(h,x)-\sC_{\ell,\sH}^*(x)$. We call
$\tri*{\Delta\sC_{\ell,\sH}(h,x)}_{\e}$ the \emph{conditional
$\e$-regret} for $\ell$. To simplify the notation, we also define for
any $t\in [0,1]$, $\sC_{\ell}(h,x,t) = t\ell(h, x, +1) + (1 -
t)\ell(h, x, -1)$ and $\Delta\sC_{\ell,\sH}(h, x, t) = \sC_{\ell}(h, x, t) - \inf_{h \in \sH}\sC_{\ell}(h, x, t)$. Thus,
$\Delta\sC_{\ell,\sH}(h, x, \eta(x)) = \Delta\sC_{\ell, \sH}(h, x)$.

A key quantity that appears in our bounds is the \emph{$\paren*{\ell,
  \sH}$-minimizability gap} $\sM_{\ell,\sH}$, which is the difference
of the best-in class error and the expectation of the minimal
conditional $\ell$-risk: $ \sM_{\ell,\sH}
 = \sR^*_{\ell,\sH} - \mathbb{E}_{X} \bracket* {\sC^*_{\ell,\sH}(x)}
 $.  This is an inherent property of the hypothesis set $\sH$ and
 distribution $\sD$ that we cannot hope to estimate or minimize. As an
 example, the minimizability gap for the $0/1$ loss and adversarial
 $0/1$ loss with $\sH_{\mathrm{all}}$ can be expressed as follows:
\begin{align*}
\sM_{\ell_{0-1},\sH_{\mathrm{all}}} &= \sR_{\ell_{0-1},\sH_{\mathrm{all}}}^*-\mathbb{E}_X\bracket*{\min\curl*{\eta(x),1-\eta(x)}}=0,\\ \sM_{\ell_{\gamma},\sH_{\mathrm{all}}} &= \sR_{\ell_{\gamma},\sH_{\mathrm{all}}}^*-\mathbb{E}_X\bracket*{\min\curl*{\eta(x),1-\eta(x)}}.
\end{align*}
\citet[Lemma~2.5]{steinwart2007compare} shows that the minimizability
gap vanishes when the loss $\ell$ is
\emph{minimizable}. \citet{awasthi2021calibration} points out that the
minimizability condition does not hold for adversarial loss functions,
and therefore that, in general,
$\sM_{\ell_{\gamma},\sH_{\mathrm{all}}}$ is strictly positive, thereby
presenting additional challenges for adversarial robust
classification. Thus, the minimizability gap is critical in the
study of adversarial surrogate loss functions. The minimizability gaps
for some common loss functions and hypothesis sets are given in
Table~\ref{tab:loss} (Appendix~\ref{app:table}), for completeness.

\section{General theorems}
\label{sec:general} 

We first introduce two main theorems that provide a general
$\sH$-consistency estimation error bound between any target loss and
surrogate loss. These bounds are $\sH$-dependent, taking into
consideration the specific hypothesis set used by a learning
algorithm. To the best of our knowledge, no such guarantee has
appeared in the past. For both theoretical and practical computational
reasons, learning algorithms typically seek a good hypothesis within a
restricted subset $\sH_{\mathrm{all}}$. Thus, in general,
$\sH$-dependent bounds can provide more relevant guarantees than
excess error bounds. Our proposed bounds are also more general in the
sense that $\sH_{\mathrm{all}}$ can be used as a special case.
\ignore { As shown in Section~\ref{sec:all}, when
  $\sH=\sH_{\mathrm{all}}$, $\ell_2=\ell_{0-1}$ and $\epsilon=0$,
  Theorem~\ref{Thm:excess_bounds_Psi} covers the excess error bounds
  in \citep{bartlett2006convexity,MohriRostamizadehTalwalkar2018}.  }
Theorems~\ref{Thm:excess_bounds_Psi} and \ref{Thm:excess_bounds_Gamma}
are counterparts of each other, while the latter may provide a more
explicit form of bounds as in \eqref{eq:est-bound}.

\begin{restatable}[\textbf{Distribution-dependent $\Psi$-bound}]
  {theorem}{ExcessBoundsPsi}
\label{Thm:excess_bounds_Psi}
 Assume that there exists a convex function $\Psi\colon
 \mathbb{R_{+}}\to \Rset$ with $\Psi(0)\geq0$ and $\epsilon\geq0$ such
 that the following holds for all $h\in \sH$ and $x\in \sX$:
\begin{equation}
\label{eq:cond_psi}
\Psi\paren*{\tri*{\Delta\sC_{\ell_2,\sH}(h,x)}_{\e}}\leq \Delta\sC_{\ell_1,\sH}(h,x).
\end{equation}
Then, for any hypothesis $h\in\sH$,
\ifdim\columnwidth=\textwidth
{
\begin{equation}
\label{eq:bound_Psi_general}
     \Psi\paren*{\sR_{\ell_2}(h)- \sR_{\ell_2,\sH}^*+\sM_{\ell_2,\sH}}
     \leq  \sR_{\ell_1}(h)-\sR_{\ell_1,\sH}^* +\sM_{\ell_1,\sH} +\max\curl*{\Psi(0),\Psi(\e)}.
\end{equation}
}
\else
{
\begin{multline}
\label{eq:bound_Psi_general}
     \Psi\paren*{\sR_{\ell_2}(h)- \sR_{\ell_2,\sH}^*+\sM_{\ell_2,\sH}}\\
     \leq  \sR_{\ell_1}(h)-\sR_{\ell_1,\sH}^* +\sM_{\ell_1,\sH} +\max\curl*{\Psi(0),\Psi(\e)}.
\end{multline}
}
\fi
\end{restatable}

\begin{restatable}[\textbf{Distribution-dependent $\Gamma$-bound}]
  {theorem}{ExcessBoundsGamma}
\label{Thm:excess_bounds_Gamma}
Assume that there exists a concave function $\Gamma\colon
\mathbb{R_{+}}\to \Rset$ and $\epsilon\geq0$ such that the following
holds for all $h\in \sH$ and $x\in \sX$:
\begin{equation}
\label{eq:cond_gamma}
\tri*{\Delta\sC_{\ell_2,\sH}(h,x)}_{\e}\leq \Gamma \paren*{\Delta\sC_{\ell_1,\sH}(h,x)}.
\end{equation}
Then, for any hypothesis $h\in\sH$,
\begin{equation}
\label{eq:bound_Gamma_general}
     \mspace{-6mu}
     \sR_{\ell_2}
     \mspace{-3mu}
     (h)
     \mspace{-1mu}
     -
     \mspace{-1mu}
     \sR_{\ell_2,\sH}^*
     \mspace{-3mu}
     \leq
     \mspace{-3mu}
     \Gamma\big(\sR_{\ell_1}
     \mspace{-3mu}
     (h)
     \mspace{-1mu}
     -
     \mspace{-1mu}
     \sR_{\ell_1,\sH}^*
     \mspace{-1mu}
     +
     \mspace{-1mu}
     \sM_{\ell_1,\sH}\big)
     \mspace{-1mu}
     -
     \mspace{-1mu}
     \sM_{\ell_2,\sH}
     \mspace{-1mu}
     +
     \mspace{-1mu}
     \epsilon.
     \mspace{-15mu}
\end{equation}
\end{restatable}
The proofs of Theorems~\ref{Thm:excess_bounds_Psi} and
\ref{Thm:excess_bounds_Gamma} are included in
Appendix~\ref{app:excess_bounds}. Below, we will mainly focus on the
case where $\Psi(0) = 0$ and $\e = 0$. Note that if $\ell_2$ is upper
bounded by $\ell_1$ and
$\sR_{\ell_1,\sH}^*-\sM_{\ell_1,\sH} = \sR_{\ell_2,\sH}^*-\sM_{\ell_2,\sH}$,
then, the following inequality automatically holds for any $h\in \sH$: 
\begin{align*}
  \sR_{\ell_2}(h)- \sR_{\ell_2,\sH}^* + \sM_{\ell_2,\sH}
  \leq  \sR_{\ell_1}(h)-\sR_{\ell_1,\sH}^*+\sM_{\ell_1,\sH}.
\end{align*}
This is a special case of Theorems~\ref{Thm:excess_bounds_Psi} and
\ref{Thm:excess_bounds_Gamma}. Indeed, since
$\sR_{\ell_1,\sH}^*-\sM_{\ell_1,\sH} =
\sR_{\ell_2,\sH}^*-\sM_{\ell_2,\sH}$, we have $\sC_{\ell_2,\sH}^*(x)
\equiv \sC_{\ell_1,\sH}^*(x)$ and thus
$\Delta\sC_{\ell_2,\sH}(h,x)\leq
\Delta\sC_{\ell_1,\sH}(h,x)$. Therefore, $\Phi$ and $\Gamma$ can be
the identity function. We refer to such cases as ``trivial
cases''. They occur when $\sM_{\ell_1,\sH}$ and $\sM_{\ell_2,\sH}$
respectively coincide with the corresponding approximation errors and
$\sR_{\ell_1,\sH_{\mathrm{all}}}^* =
\sR_{\ell_2,\sH_{\mathrm{all}}}^*$. We will later see such cases for
specific loss functions and hypothesis sets (See
\eqref{eq:rho-lin-est-2} in Appendix~\ref{app:rho-lin} and
\eqref{eq:rho-lin-est-adv-3} in Appendix~\ref{app:rho-lin-adv}). Let
us point out, however, that the corresponding $\sH$-consistency
estimation error bounds are still valid and worth studying
since they can be shown to be the tightest 
(Theorems~\ref{Thm:tightness} and \ref{Thm:tightness-adv}).

Theorem~\ref{Thm:excess_bounds_Psi} is distribution-dependent, in the
sense that, for a fixed distribution, if we find a $\Psi$ that
satisfies condition~\eqref{eq:cond_psi}, then the bound
\eqref{eq:bound_Psi_general} only gives guarantee for that same
distribution. Since the distribution $\sD$ of interest is typically
unknown, to obtain guarantees for $\sD$, if the only information given
is that $\sD$ belongs to a set of distributions $\sP$, we need to find
a $\Psi$ that satisfies condition~\eqref{eq:cond_psi} for all the
distributions in $\sP$. The choice of $\Psi$ is critical, since it
determines the form of the bound obtained. We say that $\Psi$ is
\emph{optimal} if any function that makes the bound
\eqref{eq:bound_Psi_general} hold for all distributions in $\sP$ is
everywhere no larger than $\Psi$. The optimal $\Psi$ leads to the
tightest $\sH$-consistency estimation error bound
\eqref{eq:bound_Psi_general} uniform over $\sP$. Specifically, when
$\sP$ consists of all distributions, we say that the bound is
distribution-independent. The above also applies to
Theorem~\ref{Thm:excess_bounds_Gamma}, except that $\Gamma$ is
\emph{optimal} if any function that makes the bound
\eqref{eq:bound_Gamma_general} hold for all distributions in $\sP$ is
everywhere no less than $\Gamma$.

When $\ell_2$ is the $0/1$ loss or the adversarial $0/1$ loss, the
conditional $\epsilon$-regret that appears in
condition~\eqref{eq:cond_psi} has explicit forms for common hypothesis
sets as characterized later in
Lemma~\ref{lemma:explicit_assumption_01} and
\ref{lemma:explicit_assumption_01_adv}, establishing the basis for
introducing non-adversarial and adversarial $\sH$-consistency
estimation error transformation in Section~\ref{sec:non-adv-general}
and \ref{sec:adv-general}. We will see later in these sections that
the transformations introduced are often the optimal $\Psi$ we are
seeking for, which respectively leads to tight non-adversarial and
adversarial distribution-independent guarantees.
In Section~\ref{sec:non-adv} and \ref{sec:adv}, we also apply our
general theorems and tools to loss functions and hypothesis sets
widely used in practice. Each case requires a careful analysis
that we present in detail.

\section{Guarantees for the zero-one loss $\ell_2 = \ell_{0-1}$}
\label{sec:non-adv}

In this section, we discuss guarantees in the non-adversarial scenario
where $\ell_2$ is the zero-one loss, $\ell_{0-1}$.  The lemma stated
next characterizes the minimal conditional $\ell_{0-1}$-risk and the
conditional $\epsilon$-regret, which will be helpful for introducing
the general tools in Section~\ref{sec:non-adv-general}. The proof is
given in Appendix~\ref{app:explicit_assumption}.  For convenience, we
will adopt the following notation: $\ov \sH =
\curl*{h\in\sH:\sign(h(x))\Delta \eta(x)\leq 0}$.

\begin{restatable}{lemma}{ExplicitAssumption}
\label{lemma:explicit_assumption_01}
Assume that $\sH$ satisfies the following condition for any $x\in \sX$:
$\curl*{\sign(h(x)) : h\in \sH} = \curl*{-1, +1}$.
Then, the minimal conditional $\ell_{0-1}$-risk is
\begin{align*}
\sC^*_{\ell_{0-1},\sH}(x)=\sC^*_{\ell_{0-1},\sH_{\mathrm{all}}}(x)=\min\curl*{\eta(x),1-\eta(x)}.
\end{align*}
The conditional $\epsilon$-regret for $\ell_{0-1}$ can be characterized as
\begin{align*}
  \tri*{\Delta\sC_{\ell_{0-1},\sH}(h,x)}_{\e}=\tri*{2 \abs*{\Delta \eta(x)}}_{\e}
  \mathds{1}_{h \in \ov \sH}\,.
\end{align*}
\end{restatable}

\subsection{Hypothesis set of all measurable functions}
\label{sec:all}

Before introducing our general tools, we will consider the case where
$\sH = \sH_{\mathrm{all}}$ and will show that previous excess error
bounds can be recovered as special cases of our results.  As shown in
\citep{steinwart2007compare}, both
$\sM_{\ell_{0-1},\sH_{\mathrm{all}}}$ and
$\sM_{\Phi,\sH_{\mathrm{all}}}$ vanish. Thus by
Lemma~\ref{lemma:explicit_assumption_01}, we obtain the following
corollary of Theorem~\ref{Thm:excess_bounds_Psi} by taking
$\epsilon=0$.

\begin{corollary}
\label{cor:excess_bounds_Psi_01_B}
Assume that there exists a convex function $\Psi\colon
\mathbb{R_{+}}\to \Rset$ with $\Psi(0)=0$ such that for any $x\in
\sX$, $\Psi\paren*{2 \abs*{\Delta \eta(x)}}\leq
\inf_{h\in\ov{\sH_{\mathrm{all}}}}\Delta\sC_{\Phi,\sH_{\mathrm{all}}}(h,x).$
Then, for any hypothesis $h\in\sH_{\mathrm{all}}$,
the following inequality holds:
    \begin{align*}
      \Psi\paren*{\sR_{\ell_{0-1}}(h)
        - \sR_{\ell_{0-1},\sH_{\mathrm{all}}}^*}
      \leq  \sR_{\Phi}(h) - \sR_{\Phi,\sH_{\mathrm{all}}}^*.
    \end{align*}
\end{corollary}
Furthermore, Corollary~\ref{cor:excess_bounds_Psi_01_M} follows from
Corollary~\ref{cor:excess_bounds_Psi_01_B} by taking the convex
function $\Psi(t)=\paren*{t/(2c)}^s$.
\begin{corollary}
\label{cor:excess_bounds_Psi_01_M}
Assume there exist $s\geq 1$ and $c>0$ such that for any $x\in \sX$,
$\abs*{\Delta \eta(x)}\leq c~\inf_{h\in\ov{\sH_{\mathrm{all}}}}\paren*{\Delta\sC_{\Phi,\sH_{\mathrm{all}}}(h,x)}^{\frac1s}$.
Then, for any hypothesis $h\in\sH_{\mathrm{all}}$,
    \begin{align*}
     \sR_{\ell_{0-1}}(h)- \sR_{\ell_{0-1},\sH_{\mathrm{all}}}^*\leq 2c~\paren*{ \sR_{\Phi}(h)-\sR_{\Phi,\sH_{\mathrm{all}}}^*}^{\frac1s}.
    \end{align*}
\end{corollary}
The excess error bound results in the literature are all covered by the above corollaries.  
As shown in Appendix~\ref{app:compare-all-measurable}, Theorem~4.7 in \citep{MohriRostamizadehTalwalkar2018} is a special case of Corollary~\ref{cor:excess_bounds_Psi_01_M} and Theorem 1.1 in \citep{bartlett2006convexity} is a special case of Corollary~\ref{cor:excess_bounds_Psi_01_B}.

\subsection{General hypothesis sets $\sH$}
\label{sec:non-adv-general}

In this section, we provide general tools to study
$\sH$-consistency estimation error bounds when the target loss is the
$0/1$ loss. We will then apply them to study specific hypothesis sets
and surrogates in Section~\ref{sec:non-adv-lin} and
\ref{sec:non-adv-NN}. Lemma~\ref{lemma:explicit_assumption_01}
characterizes the conditional $\epsilon$-regret for $\ell_{0-1}$ with
common hypothesis sets. Thus, Theorems~\ref{Thm:excess_bounds_Psi} and \ref{Thm:excess_bounds_Gamma} can be instantiated as
Theorems~\ref{Thm:excess_bounds_Psi_01_general} and \ref{Thm:excess_bounds_Gamma_01_general} in these cases (see
Appendix~\ref{app:theorems}). They are powerful distribution-dependent
bounds and, as discussed in Section~\ref{sec:general}, the bounds
become distribution-independent if the corresponding conditions can be
verified for all the distributions with some $\Psi$, which is
equivalent to verifying the condition in the following theorem.

\begin{restatable}[\textbf{Distribution-independent $\Psi$-bound}]
  {theorem}{ExcessBoundsPsiUniform}
\label{Thm:excess_bounds_Psi_uniform}
Assume that $\sH$ satisfies the condition of
Lemma~\ref{lemma:explicit_assumption_01}. Assume that there exists a
convex function $\Psi\colon \mathbb{R_{+}} \to \Rset$ with $\Psi(0) =
0$ and $\epsilon\geq0$ such that for any $t\in\left[1/2,1\right]$,
\begin{align*}
  \Psi \paren*{\tri*{2t-1}_{\e}}
  \leq \inf_{x\in \sX,h\in\sH:h(x)<0}\Delta\sC_{\Phi,\sH}(h,x,t).
\end{align*}
Then, for any hypothesis $h\in\sH$ and any distribution,
\ifdim\columnwidth = \textwidth
{
\begin{equation}
\label{eq:bound_Psi_01}
     \Psi\paren*{\sR_{\ell_{0-1}}(h)- \sR_{\ell_{0-1},\sH}^*+\sM_{\ell_{0-1},\sH}}
     \leq  \sR_{\Phi}(h)-\sR_{\Phi,\sH}^*+\sM_{\Phi,\sH}+\max\curl*{0,\Psi(\e)}.
\end{equation}
}
\else
{
\begin{multline}
\label{eq:bound_Psi_01}
     \Psi\paren*{\sR_{\ell_{0-1}}(h)- \sR_{\ell_{0-1},\sH}^*+\sM_{\ell_{0-1},\sH}}\\
     \leq  \sR_{\Phi}(h)-\sR_{\Phi,\sH}^*+\sM_{\Phi,\sH}+\max\curl*{0,\Psi(\e)}.
\end{multline}
}
\fi
\end{restatable}
The counterpart of Theorem~\ref{Thm:excess_bounds_Psi_uniform} is
Theorem~\ref{Thm:excess_bounds_Gamma_uniform}
(distribution-independent $\Gamma$-bound), deferred to
Appendix~\ref{app:theorems} due to space limitations. The proofs for
both theorems are included in
Appendix~\ref{app:uniform}. Theorem~\ref{Thm:excess_bounds_Psi_uniform}
provides the general tool to derive distribution-independent
$\sH$-consistency estimation error bounds. They are in fact
tight if we choose $\Psi$ to be the \emph{$\sH$-estimation error transformation} defined as follows.

\begin{figure}[t]
\begin{center}
\hspace{-.25cm}
\includegraphics[scale=0.29]{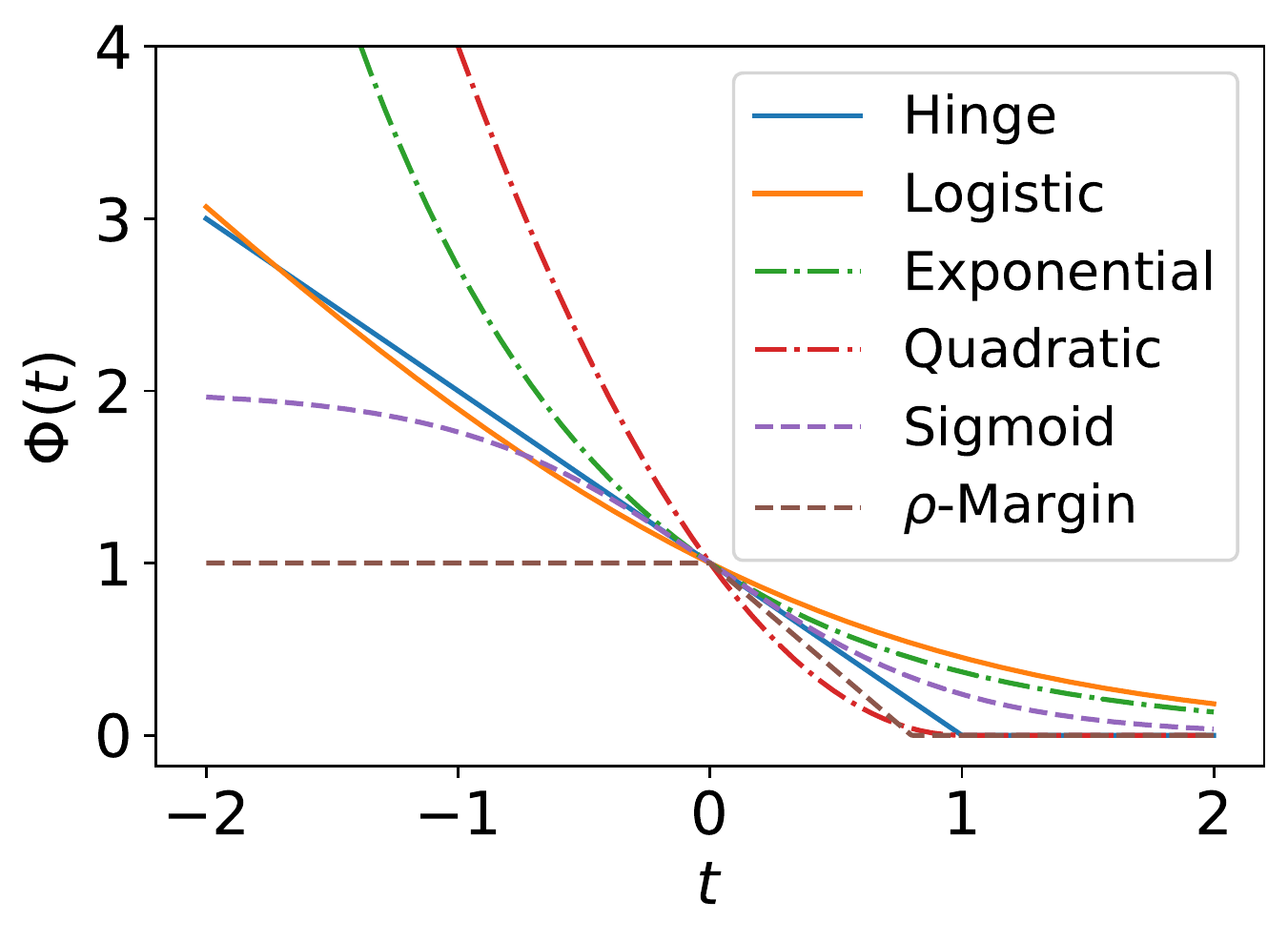}
\includegraphics[scale=0.29]{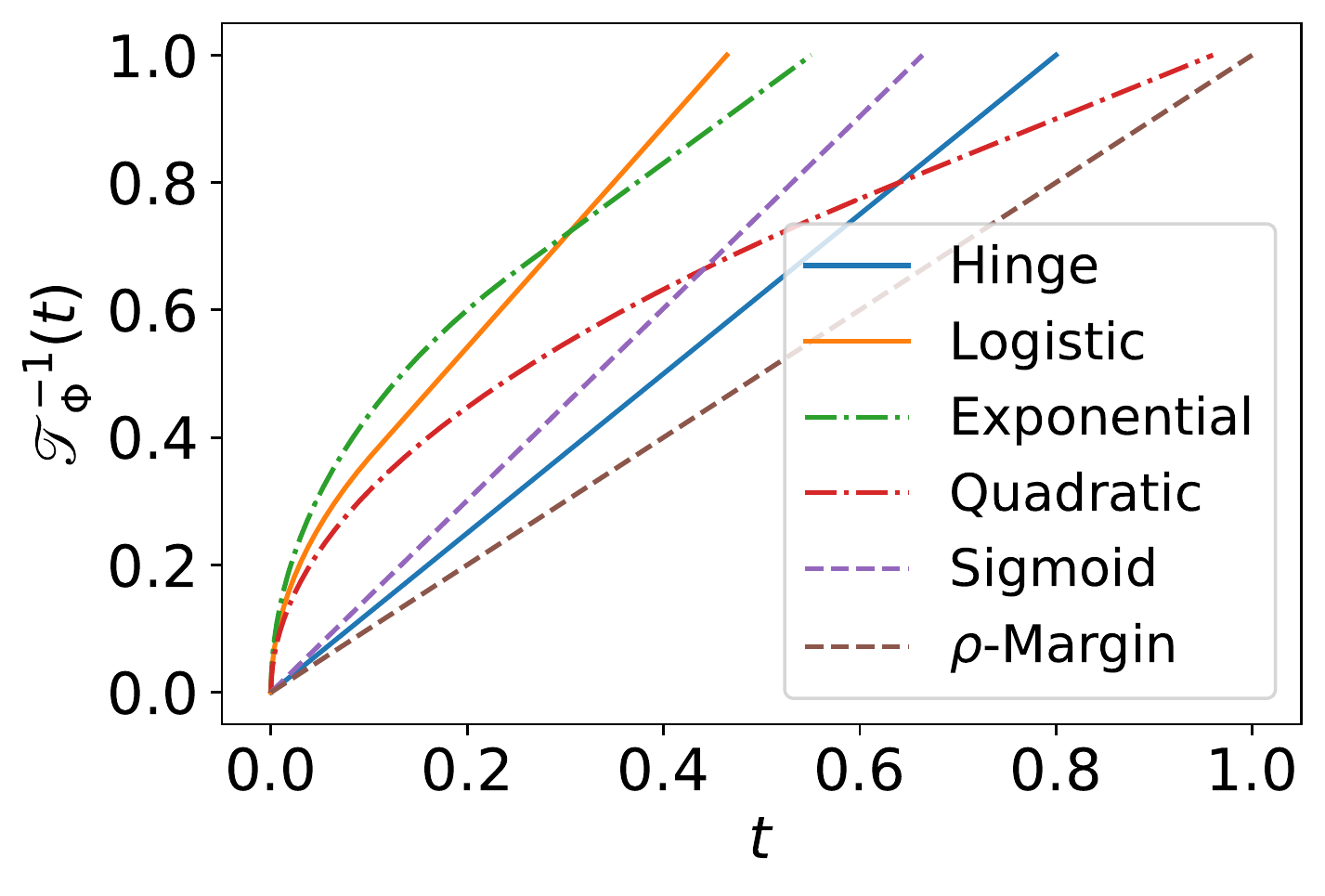}
\vskip -0.15in
\caption{Left: surrogates. Right: $\sH_{\mathrm{lin}}$-est. error trans. inv.}
\label{fig:surrogate}
\end{center}
\vskip -0.3in
\end{figure}
\begin{definition}[\textbf{$\sH$-estimation error transformation}]
\label{def:trans}
The \emph{$\sH$-estimation error transformation} of
$\Phi$ is defined on $t\in \left[0,1\right]$ by
$
\sT_{\Phi}\paren*{t}=\sT(t)\mathds{1}_{t\in
  \left[\epsilon,1\right]}+(\sT(\e)/\e)\,t\mathds{1}_{t\in
  \left[0,\epsilon\right)}
$,
where $\sT(t):=\inf_{x\in \sX,h\in\sH:h(x)<
  0}\Delta\sC_{\Phi,\sH}\paren*{h,x,\frac{t+1}{2}}$.
\end{definition}
Observe that for any $t\in\left[(1+\epsilon)/2,1\right]$,
$
  \sT_{\Phi}\paren*{2t-1}
  = \inf_{x\in \sX,h\in\sH:h(x)<0}\Delta\sC_{\Phi,\sH}(h,x,t)
$.
Taking $\Psi=\sT_{\Phi}$ satisfies the condition in
Theorem~\ref{Thm:excess_bounds_Psi_uniform} if $\sT_{\Phi}$ is convex
with $\sT_{\Phi}(0)=0$. Moreover, as mentioned earlier, it actually
leads to the tightest $\sH$-consistency estimation error bound
\eqref{eq:bound_Psi_01} when $\e=0$.

\begin{restatable}[\textbf{Tightness}]{theorem}{Tightness}
\label{Thm:tightness}
Suppose that $\sH$ satisfies the condition of
Lemma~\ref{lemma:explicit_assumption_01} and that $\e=0$. If
$\sT_{\Phi}$ is convex with $\sT_{\Phi}(0)=0$, then, for any
$t\in[0,1]$ and $\delta>0$, there exist a distribution $\sD$ and a
hypothesis $h\in\sH$ such that $\sR_{\ell_{0-1}}(h)-
\sR_{\ell_{0-1},\sH}^*+\sM_{\ell_{0-1},\sH}=t$ and
$\sT_{\Phi}(t)\leq\sR_{\Phi}(h)-\sR_{\Phi,\sH}^*+\sM_{\Phi,\sH}\leq
\sT_{\Phi}(t) + \delta$.
\end{restatable}
The proof is included in Appendix~\ref{app:tightness}. In other words,
when $\e=0$, if $\sT_{\Phi}$ is convex with $\sT_{\Phi}(0)=0$, it is
optimal for the distribution-independent
bound~\eqref{eq:bound_Psi_01}. Moreover, if $\sT_{\Phi}$ is
additionally invertible and non-increasing, $\sT_{\Phi}^{-1}$ is the
optimal function for the distribution-independent bound in
Theorem~\ref{Thm:excess_bounds_Gamma_uniform}
(Appendix~\ref{app:theorems}) and the two bounds are equivalent.

In the following sections, we will see that all these assumptions hold
for common loss functions with linear and neural network hypothesis
sets. Next, we will apply Theorems~\ref{Thm:excess_bounds_Psi_uniform}
and \ref{Thm:tightness} to the linear models
(Section~\ref{sec:non-adv-lin}) and neural networks
(Section~\ref{sec:non-adv-NN}). Each case requires a detailed analysis
(See Appendix~\ref{app:derivation-lin} and \ref{app:derivation-NN}).

\subsection{Linear hypotheses}
\label{sec:non-adv-lin}
\begin{table}[t]
\caption{$\sH_{\mathrm{lin}}$-estimation error
  transformation and $\sH_{\mathrm{lin}}$-consistency
  estimation error bounds with $\epsilon=0$.}
    \label{tab:compare}
\begin{center}
    \vskip -0.2in
    \begin{small}
\resizebox{\columnwidth}{!}{
    \begin{tabular}{l|lll}
    \toprule
      Surrogates & $\sT_{\Phi}(t),\, t\in [0,1]$   &  Bound \\
    \midrule
      Hinge & $\min \curl*{B, 1} \, t $  & \eqref{eq:hinge-lin-est} \\
      Logistic & $\begin{cases}
\frac{t+1}{2}\log_2(t+1)+\frac{1-t}{2}\log_2(1-t),\quad &  t\leq \frac{e^B-1}{e^B+1},\\
1-\frac{t+1}{2}\log_2(1+e^{-B})-\frac{1-t}{2}\log_2(1+e^B),\quad & t> \frac{e^B-1}{e^B+1}.
\end{cases}$ & \eqref{eq:log-lin-est}\\
      Exponential & $\begin{cases}
1-\sqrt{1-t^2}, & t\leq \frac{e^{2B}-1}{e^{2B}+1},\\
1-\frac{t+1}{2}e^{-B}-\frac{1-t}{2}e^B, & t> \frac{e^{2B}-1}{e^{2B}+1}.
\end{cases}$ & \eqref{eq:exp-lin-est}\\
      Quadratic & $\begin{cases}
t^2, & t\leq B,\\
2B \,t-B^2, & t> B.
\end{cases}$ & \eqref{eq:quad-lin-est}\\
      Sigmoid & $\tanh(kB) \, t$ & \eqref{eq:sig-lin-est}\\
      $\rho$-Margin & $\frac{\min\curl*{B,\rho}}{\rho} \, t$ & \eqref{eq:rho-lin-est}\\
    \bottomrule
    \end{tabular}
    }
    \end{small}
    \end{center}
    \vskip -0.3in
\end{table}
By applying Theorems~\ref{Thm:excess_bounds_Psi_uniform} and \ref{Thm:tightness}, we can derive $\sH_{\mathrm{lin}}$-consistency
estimation error bounds for common loss functions defined in
Table~\ref{tab:loss} of Appendix~\ref{app:table}. Table~\ref{tab:compare} supplies the $\sH_{\mathrm{lin}}$-estimation error transformation $\sT_{\Phi}$ and the corresponding bounds for those loss functions. The inverse $\sT_{\Phi}^{-1}$ is given in Table~\ref{tab:compare_inverse} of Appendix~\ref{app:table}.
Surrogates $\Phi$ and their corresponding $\sT_{\Phi}^{-1}$ ($B=0.8$) are visualized in Figure~\ref{fig:surrogate}. Theorems~\ref{Thm:excess_bounds_Psi_uniform} and \ref{Thm:tightness} apply to all these cases since
$\sT_{\Phi}$ is convex, increasing, invertible and satisfies that
$\sT_{\Phi}(0)=0$. More precisely, taking $\Psi=\sT_{\Phi}$ and $\e=0$
in \eqref{eq:bound_Psi_01} and using the inverse function
$\sT_{\Phi}^{-1}$ directly give the tightest bound. As an example, for
the sigmoid loss,
$\sT_{\Phi_{\mathrm{sig}}}^{-1}(t)=\frac{t}{\tanh(kB)}$. Then the
bound~\eqref{eq:bound_Psi_01} becomes $\sR_{\ell_{0-1}}(h)-
\sR_{\ell_{0-1},\sH_{\mathrm{lin}}}^*\leq
\paren{\sR_{\Phi_{\mathrm{sig}}}(h)-
  \sR_{\Phi_{\mathrm{sig}},\sH_{\mathrm{lin}}}^*
+\sM_{\Phi_{\mathrm{sig}},\sH_{\mathrm{lin}}}}/\tanh(kB)
- \sM_{\ell_{0-1},\sH_{\mathrm{lin}}}$, which is
\eqref{eq:sig-lin-est} in Table~\ref{tab:compare}. Furthermore, after
plugging in the minimizability gaps concluded in Table~\ref{tab:loss},
we will obtain the novel bound $\sR_{\ell_{0-1}}(h)-
\sR_{\ell_{0-1},\sH_{\mathrm{all}}}^*\leq
\paren{\sR_{\Phi_{\mathrm{sig}}}(h)-
  \mathbb{E}_{X}\bracket*{1-\abs*{1-2\eta(x)}
    \tanh\paren*{k\paren*{W\norm*{x}_p+B}}}
}/\tanh(kB)$ (\eqref{eq:sig-lin-est-2} in
Appendix~\ref{app:sig-lin}). The bounds for other surrogates are
similarly derived in Appendix~\ref{app:derivation-lin}. For the
logistic loss and exponential loss, to simplify the expression, the
bounds are obtained by plugging in an upper bound of
$\sT_{\Phi}^{-1}$.

Let us emphasize that these $\sH$-consistency estimation error bounds
are novel in the sense that they are all hypothesis set-dependent and, to our knowledge,
no such guarantee has been presented before. More precisely, the
bounds of Table~\ref{tab:compare} depend directly on the parameter $B$
in the linear models and parameters of the loss function (e.g., $k$ in
sigmoid loss). Thus, for a fixed hypothesis $h\in \sH_{\mathrm{lin}}$,
we may give the tightest bound by choosing the best parameter $B$. As
an example, Appendix~\ref{app:sig-lin} shows that the bound
\eqref{eq:sig-lin-est-2} with $B = \plus \infty$ coincides with the
excess error bound known for the sigmoid loss
\citep{bartlett2006convexity}. However, for a fixed hypothesis $h$, by
varying $B$ (hypothesis set) and $k$ (loss function), we may obtain a
finer bound! Thus studying hypothesis set-dependent bounds can guide
us to select the most suitable hypothesis set and loss
function. Moreover, as shown by Theorem~\ref{Thm:tightness}, all the
bounds obtained by directly using $\sT_{\Phi}^{-1}$ are tight and
cannot be further improved.

\subsection{One-hidden-layer ReLU neural networks}
\label{sec:non-adv-NN}

In this section, we give $\sH$-consistency estimation error bounds for
one-hidden-layer ReLU neural networks
$\sH_{\mathrm{NN}}$. Table~\ref{tab:compare-NN} in
Appendix~\ref{app:table} is the counterpart of Table~\ref{tab:compare}
for $\sH_{\mathrm{NN}}$. Different from the bounds in the linear case,
all the bounds in Table~\ref{tab:compare-NN} not only depend on $B$,
but also depend on $\Lambda$, which is a new parameter in
$\sH_{\mathrm{NN}}$. This further illustrates that our bounds are
hypothesis set-dependent and, as with the linear case, adequately
choosing the parameters $\Lambda$ and $B$ in $\sH_{\mathrm{NN}}$ would
give us better hypothesis set-dependent guarantees than standard
excess error bounds. Our proofs and techniques could also be adopted
for the analysis of multi-layer neural networks.

\subsection{Guarantees under Massart's noise condition}
\label{sec:noise-non-adv}

The distribution-independent $\sH$-consistency estimation error bound
\eqref{eq:bound_Psi_01} cannot be improved, since they are tight as
shown in Theorem~\ref{Thm:tightness}. However, the bounds can be
further improved in the distribution-dependent setting. Indeed, we
will study how $\sH$-consistency estimation error bounds can be
improved under low noise conditions, which impose the restrictions on
the conditional distribution $\eta(x)$. We consider Massart's noise
condition \citep{massart2006risk} which is defined as follows.

\begin{definition}[\textbf{Massart's noise}]
\label{def:massarts-noise}
The distribution $\sD$ over $\cX\times \cY$ satisfies Massart's noise condition if
$\abs*{\Delta \eta(x)} \geq \beta \text{ for almost all } x \in \cX$,
for some constant $\beta \in (0, 1/2]$.
\end{definition}
When it is known that the distribution $\sD$ satisfies Massart's noise condition
with $\beta$, in contrast with the distribution-independent bounds, we
can require the bounds \eqref{eq:bound_Psi_general} and
\eqref{eq:bound_Gamma_general} to hold uniformly only for such
distributions. With Massart's noise condition, we
introduce a modified $\sH$-estimation error
transformation in Proposition~\ref{prop:prop-noise}
(Appendix~\ref{app:derivation-all_noise}), which verifies
condition~\eqref{eq:condition_Psi_general} of
Theorem~\ref{Thm:excess_bounds_Psi_01_general} (the finer distribution
dependent guarantee mentioned before, deferred to
Appendix~\ref{app:theorems}) for all distributions under the noise
condition. Then, using this transformation, we can obtain more
favorable distribution-dependent bounds. As an example, we consider
the quadratic loss $\Phi_{\mathrm{quad}}$, the logistic loss
$\Phi_{\mathrm{log}}$ and the exponential loss $\Phi_{\mathrm{exp}}$
with $\sH_{\mathrm{all}}$. For all distributions and $h\in
\sH_{\mathrm{all}}$, as shown in
\citep{Zhang2003,bartlett2006convexity,MohriRostamizadehTalwalkar2018},
the following holds:
\begin{align*}
  \sR_{\ell_{0-1}}(h)- \sR_{\ell_{0-1},\sH_{\mathrm{all}}}^*&
  \leq \sqrt{2}\paren*{\sR_{\Phi}(h) - \sR_{\Phi,\sH_{\mathrm{all}}}^*}^{1/2},
\end{align*}
when the surrogate loss $\Phi$ is $\Phi_{\mathrm{log}}$ or
$\Phi_{\mathrm{exp}}$. If $\Phi=\Phi_{\mathrm{quad}}$, then the
constant multiplier $\sqrt{2}$ can be removed. For distributions that
satisfy Massart's noise condition with $\beta$, as proven in
Appendix~\ref{app:derivation-all_noise}, for any $h\in
\sH_{\mathrm{all}}$ such that $\sR_{\Phi}(h) \leq
\sR_{\Phi,\sH_{\mathrm{all}}}^*+\sT(2\beta)$, the consistency excess
error bound is improved from the square-root dependency to a linear
dependency:
\begin{align}
\label{eq:non-adv-noise}
    \mspace{-14mu}
    \sR_{\ell_{0-1}}(h)- \sR_{\ell_{0-1},\sH_{\mathrm{all}}}^*
    \mspace{-2mu}
    \leq
    \mspace{-2mu}
    2\beta\paren*{\sR_{\Phi}(h)- \sR_{\Phi,\sH_{\mathrm{all}}}^*} / \sT(2\beta),
    \mspace{-8mu}
\end{align}
where $\sT(t)$ equals to $t^2$,
$\frac{t+1}{2}\log_2(t+1)+\frac{1-t}{2}\log_2(1-t)$ and
$1-\sqrt{1-t^2}$ for $\Phi_{\mathrm{quad}}$, $\Phi_{\mathrm{log}}$ and
$\Phi_{\mathrm{exp}}$ respectively. These linear dependent bounds are
tight, as illustrated in Section~\ref{sec:simulations}.

\section{Guarantees for the adversarial loss $\ell_2 = \ell_{\gamma}$}
\label{sec:adv}

In this section, we discuss the adversarial scenario where $\ell_2$ is
the adversarial $0/1$ loss $\ell_{\gamma}$. We consider
\emph{symmetric} hypothesis sets, which satisfy: $h\in\sH$ if and only
if $-h\in \sH$.
For convenience, we will adopt
the following definitions:
\begin{align*}
\uv h_\gamma(x)  =\inf_{x'\colon \|x - x'\|_p\leq\gamma} h(x') \qquad
\ov h_\gamma(x)  =\sup_{x'\colon \|x - x'\|_p\leq\gamma} h(x').
\end{align*}
We also define $\ov \sH_\gamma = \curl*{h\in\sH:\uv h_\gamma(x)\leq 0
  \leq \ov h_\gamma(x)}$.
The following characterization of the minimal conditional
$\ell_{\gamma}$-risk and conditional $\epsilon$-regret is based on
\citep[Lemma~27]{awasthi2021calibration} and will be helpful in
introducing the general tools in Section~\ref{sec:adv-general}. The
proof is similar and is included in
Appendix~\ref{app:explicit_assumption} for completeness.

\begin{restatable}{lemma}{ExplicitAssumptionAdv}
\label{lemma:explicit_assumption_01_adv}
Assume that $\sH$ is symmetric. Then, the minimal conditional
$\ell_{\gamma}$-risk is
\begin{align*}
  \sC^*_{\ell_{\gamma},\sH}(x)
  = \min\curl*{\eta(x), 1 - \eta(x)}\mathds{1}_{\ov \sH_\gamma\neq \sH}
  + \mathds{1}_{\ov \sH_\gamma=\sH}\,.
\end{align*}
The conditional $\e$-regret for $\ell_{\gamma}$ can be characterized
as
\begin{align*}
\tri*{\Delta\sC_{\ell_{\gamma},\sH}(h,x)}_{\e}
=
\begin{cases}
\tri*{\abs*{\Delta \eta(x)}+\frac12}_{\e} 
&h \in \ov \sH_\gamma\subsetneqq \sH\\
\tri*{2\Delta \eta(x)}_{\e}
& \ov h_\gamma(x)<0\\
\tri*{-2\Delta \eta(x)}_{\e}
&\uv h_\gamma(x)>0 \\
0 
&\text{otherwise}
\end{cases}
\end{align*}
\end{restatable}

\subsection{General hypothesis sets $\sH$}
\label{sec:adv-general}

As with the non-adversarial case, we begin by providing general
theoretical tools to study $\sH$-consistency estimation error bounds
when the target loss is the adversarial $0/1$
loss. Lemma~\ref{lemma:explicit_assumption_01_adv} characterizes the
conditional $\epsilon$-regret for $\ell_{\gamma}$ with symmetric
hypothesis sets. Thus, Theorems~\ref{Thm:excess_bounds_Psi} and \ref{Thm:excess_bounds_Gamma} can be instantiated as
Theorems~\ref{Thm:excess_bounds_Psi_01_general_adv} and
\ref{Thm:excess_bounds_Gamma_01_general_adv} (See
Appendix~\ref{app:theorems}) in these cases. These results are
distribution-dependent and can serve as general tools. For example, we
can use these tools to derive more favorable guarantees under noise
conditions (Section~\ref{sec:noise-adv}). As in the previous section,
we present their distribution-independent version in the following
theorem.

\begin{restatable}[\textbf{Adversarial distribution-independent $\Psi$-bound}]
  {theorem}{ExcessBoundsPsiUniformAdv}
\label{Thm:excess_bounds_Psi_uniform-adv}
Suppose that $\sH$ is symmetric. Assume there exist a convex function
$\Psi\colon \mathbb{R_{+}} \to \Rset$ with $\Psi(0)=0$ and
$\epsilon\geq0$ such that the following holds for any
$t\in\left[1/2,1\right]\colon$
\begin{align*}
  &\Psi\paren*{\tri*{t}_{\e}}
  \leq \inf_{x\in\sX,h\in \ov \sH_\gamma\subsetneqq \sH}\Delta\sC_{\wt{\Phi},\sH}(h,x,t),\\
  &\Psi\paren*{\tri*{2t-1}_{\e}}
  \leq \inf_{x\in \sX,h\in\sH\colon  \ov h_\gamma(x)< 0}\Delta\sC_{\wt{\Phi},\sH}(h,x,t).
\end{align*}
Then, for any hypothesis $h\in\sH$ and any distribution,
\ifdim\columnwidth=\textwidth
{
\begin{equation}
\label{eq:bound_Psi_01_adv}
     \Psi\paren*{\sR_{\ell_{\gamma}}(h) - \sR_{\ell_{\gamma},\sH}^* + \sM_{\ell_{\gamma},\sH}}
     \leq  \sR_{\wt{\Phi}}(h)
     - \sR_{\wt{\Phi},\sH}^* + \sM_{\wt{\Phi},\sH}+\max\curl*{0,\Psi(\e)}.
\end{equation}
}
\else
{
\begin{multline}
\label{eq:bound_Psi_01_adv}
     \Psi\paren*{\sR_{\ell_{\gamma}}(h)- \sR_{\ell_{\gamma},\sH}^*+\sM_{\ell_{\gamma},\sH}}\\
     \leq  \sR_{\wt{\Phi}}(h)-\sR_{\wt{\Phi},\sH}^*
     +\sM_{\wt{\Phi},\sH}+\max\curl*{0,\Psi(\e)}.
\end{multline}
}
\fi
\end{restatable}
The counterpart of Theorem~\ref{Thm:excess_bounds_Psi_uniform-adv} is
Theorem~\ref{Thm:excess_bounds_Gamma_uniform-adv} (adversarial
distribution-independent $\Gamma$-bound), deferred to
Appendix~\ref{app:theorems} due to space limitations. The proofs for
both theorems are included in Appendix~\ref{app:uniform-adv}.  As with
the non-adversarial scenario, the tightest
distribution-independent $\sH$-consistency estimation error bounds
obtained by Theorem~\ref{Thm:excess_bounds_Psi_uniform-adv} can be
achieved by the optimal $\Psi$, which is the \emph{adversarial
$\sH$-estimation error transformation} defined as follows.
\begin{definition}[\textbf{Adversarial $\sH$-estimation error transformation}]
\label{def:trans-adv}
The \emph{adversarial $\sH$-estimation error transformation} of
$\wt{\Phi}$ is defined on $t\in \left[0,1\right]$ by
$\sT_{\wt{\Phi}}\paren*{t}= \min\curl*{\sT_1(t),\sT_2(t)}$,
\begin{align*}
\text{where} \quad &\sT_1(t):=\h{\sT}_1(t)\mathds{1}_{t\in [1/2,1]}+ 2\,\h{\sT}_1(1/2)\, t\mathds{1}_{t\in [0,1/2)}, \\
    &\sT_2(t):=\h{\sT}_2(t)\mathds{1}_{t\in \left[\e,1\right]}+ \big(\h{\sT}_2(\e)/\epsilon\big)\,t\mathds{1}_{t\in \left[0,\e\right)}, \\
\text{with} \quad &\h{\sT}_1(t):= \inf_{x\in \sX,h\in \ov \sH_\gamma\subsetneqq \sH}\Delta\sC_{\wt{\Phi},\sH}(h,x,t),\\[-.25cm]
    &\h{\sT}_2(t):= \inf_{x\in \sX,h\in\sH\colon \ov h_\gamma(x)< 0}\Delta\sC_{\wt{\Phi},\sH}\big(h,x,\frac{t+1}{2}\big).
\end{align*}
\end{definition}
It is clear that $\sT_{\wt{\Phi}}$ satisfies assumptions in
Theorem~\ref{Thm:excess_bounds_Psi_uniform-adv}. The next theorem
shows that it gives the tightest $\sH$-consistency estimation error
bound \eqref{eq:bound_Psi_01_adv} under certain conditions.
\begin{restatable}[\textbf{Adversarial tightness}]{theorem}{TightnessAdv}
\label{Thm:tightness-adv}
Suppose that $\sH$ is symmetric and that $\e=0$. If
$\sT_{\wt{\Phi}}=\min\curl*{\sT_1,\sT_2}$ is convex with
$\sT_{\wt{\Phi}}(0)=0$ and $\sT_2\leq \sT_1$, then, for any
$t\in[0,1]$ and $\delta>0$, there exist a distribution $\sD$ and a
hypothesis $h\in\sH$ such that $\sR_{\ell_{\gamma}}(h)-
\sR_{\ell_{\gamma},\sH}^*+\sM_{\ell_{\gamma},\sH}=t$ and
$\sT_{\wt{\Phi}}(t)
\leq\sR_{\wt{\Phi}}(h)-\sR_{\wt{\Phi},\sH}^*+\sM_{\wt{\Phi},\sH}\leq
\sT_{\wt{\Phi}}(t) + \delta$.
\end{restatable}
The proof is included in Appendix~\ref{app:tightness}. In other words,
when $\e=0$, if $\sT_2\leq \sT_1$ and $\sT_{\wt{\Phi}}$ is convex with
$\sT_{\wt{\Phi}}(0)=0$, $\sT_{\wt{\Phi}}$ is the optimal function for
the distribution-independent
bound~\eqref{eq:bound_Psi_01_adv}. Moreover, if $\sT_{\wt{\Phi}}$ is
additionally invertible and non-increasing, $\sT_{\wt{\Phi}}^{-1}$ is
the optimal function for the distribution-independent bound in
Theorem~\ref{Thm:excess_bounds_Gamma_uniform-adv}
(Appendix~\ref{app:theorems}) and the two bounds will be equivalent.

We will see that all these assumptions hold for cases considered in
Section~\ref{sec:adv-lin} and \ref{sec:adv-NN}. Next, we will apply
Theorem~\ref{Thm:excess_bounds_Psi_uniform-adv} along with the
tightness guarantee Theorem~\ref{Thm:tightness-adv} to study specific
hypothesis sets and adversarial surrogate loss functions in
Section~\ref{sec:negative} for negative results and
Section~\ref{sec:adv-lin} and \ref{sec:adv-NN} for positive results. A
careful analysis is presented in each case (See
Appendix~\ref{app:derivation-adv}).

\subsection{Negative results for adversarial robustness}
\label{sec:negative}

\citet{awasthi2021calibration} show that supremum-based convex
loss functions of the type $\wt{\Phi} = \sup_{x' \colon \|x-x'\|_p\leq
  \gamma}\Phi(y h(x'))$, where $\Phi$ is convex and non-increasing,
are \emph{not $\sH$-calibrated with respect to $\ell_{\gamma}$} for
$\sH$ containing 0, that is \emph{regular for adversarial calibration}
(Definition~\ref{def:regularity} in Appendix~\ref{app:negative_adv}),
e.g., $\sH_{\mathrm{lin}}$ and $\sH_{\mathrm{NN}}$.  Similarly, we
show that there are no non-trivial adversarial $\sH$-consistency
estimation error bounds with respect to $\ell_{\gamma}$ for
supremum-based convex loss functions and supremum-based sigmoid loss
with such hypothesis sets. Note that \citet{awasthi2021calibration} do
not study the sigmoid loss, which is non-convex. Thus, our results go
beyond their results for convex adversarial surrogates.

\begin{restatable}[\textbf{Negative results for robustness}]{theorem}{NegativeConvexAdv}
\label{Thm:negative_convex_adv}
Suppose that $\sH$ contains $0$ and is regular for adversarial
calibration. Let $\ell_1$ be supremum-based convex loss or
supremum-based sigmoid loss and $\ell_2=\ell_{\gamma}$. Then, $f\geq
1/2$ are the only non-decreasing functions $f$ such that
\eqref{eq:est-bound} holds.
\end{restatable}
The proof is given in
Appendix~\ref{app:negative_adv}. In other words, the function $f$ in bound \eqref{eq:est-bound} must be lower bounded by $1/2$ for such adversarial surrogates. Theorem~\ref{Thm:negative_convex_adv}
implies that the loss functions commonly used in practice for
optimizing the adversarial loss cannot benefit from any useful
$\sH$-consistency estimation error guarantees. Instead, we show in
Section~\ref{sec:adv-lin} and \ref{sec:adv-NN} that the supremum-based
$\rho$-margin loss $\wt{\Phi}_{\rho}=\sup_{x'\colon \|x-x'\|_p\leq\gamma}\Phi_{\mathrm{\rho}}(y h(x'))$ proposed by
\citep{awasthi2021calibration} admits favorable adversarial
$\sH$-consistency estimation error bounds. These bounds would also
imply significantly stronger results than the asymptotic
$\sH$-consistency guarantee in \citep{awasthi2021calibration}.


\subsection{Linear hypotheses}
\label{sec:adv-lin}

In this section, by applying
Theorems~\ref{Thm:excess_bounds_Psi_01_general_adv} and
\ref{Thm:excess_bounds_Gamma_01_general_adv}, we derive the
adversarial $\sH_{\mathrm{lin}}$-consistency estimation error
bound~\eqref{eq:rho-lin-est-adv} in Table~\ref{tab:compare-adv} of
Appendix~\ref{app:table} for supremum-based $\rho$-margin loss. This is a completely new consistency estimation
error bound in the adversarial setting. As with the non-adversarial
case, the bound is dependent on the parameter $B$ in linear hypothesis
set and $\rho$ in the loss function. This helps guide the choice of
loss functions once the hypothesis set is fixed. More precisely, if
$B>0$ is known, we can always choose $\rho<B$ such that the bound is
the tightest. Moreover, the bound can turn into more significant
$\e$-consistency results in adversarial setting than the
$\sH$-consistency result in \citep{awasthi2021calibration}.
\begin{corollary}
\label{cor:lin-adv-stonger}
 Let $\sD$ be a distribution over $\sX\times\sY$ such that
 $\sM_{\wt{\Phi}_{\rho},\sH_{\mathrm{lin}}}\leq\e$ for some $\e\geq
 0$.  Then, the following holds:
\begin{align*}
\mspace{-3mu}
 \sR_{\ell_{\gamma}}
 \mspace{-3mu}
 (h)
 \mspace{-1mu}
 - 
 \mspace{-1mu}
 \sR_{\ell_{\gamma},\sH_{\mathrm{lin}}}^* 
 \mspace{-4mu}
  \leq 
  \mspace{-4mu}
  \rho\paren*{\sR_{\wt{\Phi}_{\rho}}
  \mspace{-3mu}
  (h)
  \mspace{-1mu}
  -
  \mspace{-1mu}
  \sR_{\wt{\Phi}_{\rho},\sH_{\mathrm{lin}}}^*
  \mspace{-1mu}
  +
 \mspace{-1mu}
 \e}/\min\curl*{B,\rho}.
\end{align*}
\end{corollary}
\citet{awasthi2021calibration} show that $\wt{\Phi}_{\rho}$ is
$\sH_{\mathrm{lin}}$-consistent with respect to $\ell_{\gamma}$ when
$\sM_{\wt{\Phi}_{\rho},\sH_{\mathrm{lin}}}=0$. This result can be
immediately implied by Corollary~\ref{cor:lin-adv-stonger}. Moreover,
Corollary~\ref{cor:lin-adv-stonger} provides guarantees for more
general cases where $\sM_{\wt{\Phi}_{\rho},\sH_{\mathrm{lin}}}$ can be
nonzero.

\subsection{One-hidden-layer ReLU neural networks}
\label{sec:adv-NN}

For the one-hidden-layer ReLU neural networks $\sH_{\mathrm{NN}}$ and
$\wt{\Phi}_{\rho}$, we have the $\sH_{\mathrm{NN}}$-estimation error
bound \eqref{eq:rho-NN-est-adv} in Table~\ref{tab:compare-adv}.
Note $\inf_{x\in\sX}\sup_{h\in\sH_{\mathrm{NN}}}\uv h_\gamma(x)$ does
not have an explicit expression. However, \eqref{eq:rho-NN-est-adv}
can be further relaxed to be \eqref{eq:rho-NN-est-adv-2} in
Appendix~\ref{app:derivation-NN-adv}, which is identical to the bound
in the linear case modulo the replacement of $B$ by $\Lambda B$. As in
the linear case, the bound is new and also
implies stronger $\e$-consistency results as follows:
\begin{corollary}
Let $\sD$ be a distribution over $\sX\times\sY$ such that
$\sM_{\wt{\Phi}_{\rho},\sH_{\mathrm{NN}}}\leq \e$ for some $\e\geq 0$.
Then,
\begin{align*}
\mspace{-6mu}
 \sR_{\ell_{\gamma}}
 \mspace{-3mu}
 (h)
 \mspace{-1mu}
 - 
 \mspace{-1mu}
 \sR_{\ell_{\gamma},\sH_{\mathrm{NN}}}^* 
 \mspace{-4mu}
  \leq 
  \mspace{-4mu}
  \rho\paren*{\sR_{\wt{\Phi}_{\rho}}
  \mspace{-3mu}
  (h)
  \mspace{-1mu}
  -
  \mspace{-1mu}
  \sR_{\wt{\Phi}_{\rho},\sH_{\mathrm{NN}}}^*
  \mspace{-10mu}
  +
 \mspace{-4mu}
 \e}
 \mspace{-1mu}
 /
 \mspace{-5mu}
 \min\curl*{\Lambda B,\rho}
 \mspace{-4mu}.
\end{align*}
\end{corollary}
Besides the bounds for $\wt{\Phi}_{\rho}$, Table~\ref{tab:compare-adv}
gives a series of results that are all new in the adversarial
setting. Like the bounds in Table~\ref{tab:compare} and
\ref{tab:compare-NN}, they are all hypothesis set dependent and very
useful. For example, the improved bounds for
$\wt{\Phi}_{\mathrm{hinge}}$ and $\wt{\Phi}_{\mathrm{sig}}$ under
noise conditions in the table can also turn into meaningful
consistency results under Massart's noise condition, as shown in
Section~\ref{sec:noise-adv}.

\subsection{Guarantees under Massart's noise condition}
\label{sec:noise-adv}

Section~\ref{sec:negative} shows that non-trivial
distribution-independent bounds for supremum-based hinge loss and
supremum-based sigmoid loss do not exist. However, under Massart's
noise condition (Definition~\ref{def:massarts-noise}), we will show
that there exist non-trivial adversarial $\sH$-consistency estimation
error bounds for the two loss functions. Furthermore, we will see that
the bounds are linear dependent as those in
Section~\ref{sec:noise-non-adv}.

As with the non-adversarial scenario, we introduce a modified
adversarial $\sH$-estimation error transformation in
Proposition~\ref{prop-adv-noise}
(Appendix~\ref{app:derivation-adv_noise}). Using this tool, we derive
adversarial $\sH$-consistency estimation error bounds for
$\wt{\Phi}_{\mathrm{hinge}}$ and $\wt{\Phi}_{\mathrm{sig}}$ under
Massart's noise condition in Table~\ref{tab:compare-adv}.
From the bounds~\eqref{eq:hinge-lin-est-adv},
\eqref{eq:sig-lin-est-adv}, \eqref{eq:hinge-NN-est-adv}, and
\eqref{eq:sig-NN-est-adv}, we can also obtain novel $\e$-consistency
results for $\wt{\Phi}_{\mathrm{hinge}}$ and
$\wt{\Phi}_{\mathrm{sig}}$ with linear models and neural networks
under Massart's noise condition.
\begin{corollary}
Let $\sH$ be $\sH_{\mathrm{lin}}$ or $\sH_{\mathrm{NN}}$. Let $\sD$ be
a distribution over $\sX\times\sY$ which satisfies Massart's noise
condition with $\beta$ such that $\sM_{\wt{\Phi},\sH}\leq\e$ for some
$\e\geq 0$. Then,
\begin{align*}
\sR_{\ell_{\gamma}}(h)- \sR_{\ell_{\gamma},\sH}^* \leq 
     \frac{1+2\beta}{4\beta}\paren{\sR_{\wt{\Phi}}(h)-\sR_{\wt{\Phi},\sH}^*+\e}/\sT(B),
\end{align*}
where $\sT(t)$ equals to $\min\curl*{t,1}$ and $\tanh\paren*{k t}$ for
$\wt{\Phi}_{\mathrm{hinge}}$ and $\wt{\Phi}_{\mathrm{sig}}$
respectively, $B$ is replaced by $\Lambda B$ for
$\sH=\sH_{\mathrm{NN}}$.
\end{corollary}
In Section~\ref{sec:simulations}, we will further show that these
linear dependency bounds in adversarial setting are tight, along with
the non-adversarial bounds we discussed earlier in
Section~\ref{sec:noise-non-adv}.

\section{Simulations}
\label{sec:simulations}
\begin{figure}[t]
\vskip -0.06in
\begin{center}
\hspace{-.25cm}
\includegraphics[width=0.5\columnwidth]{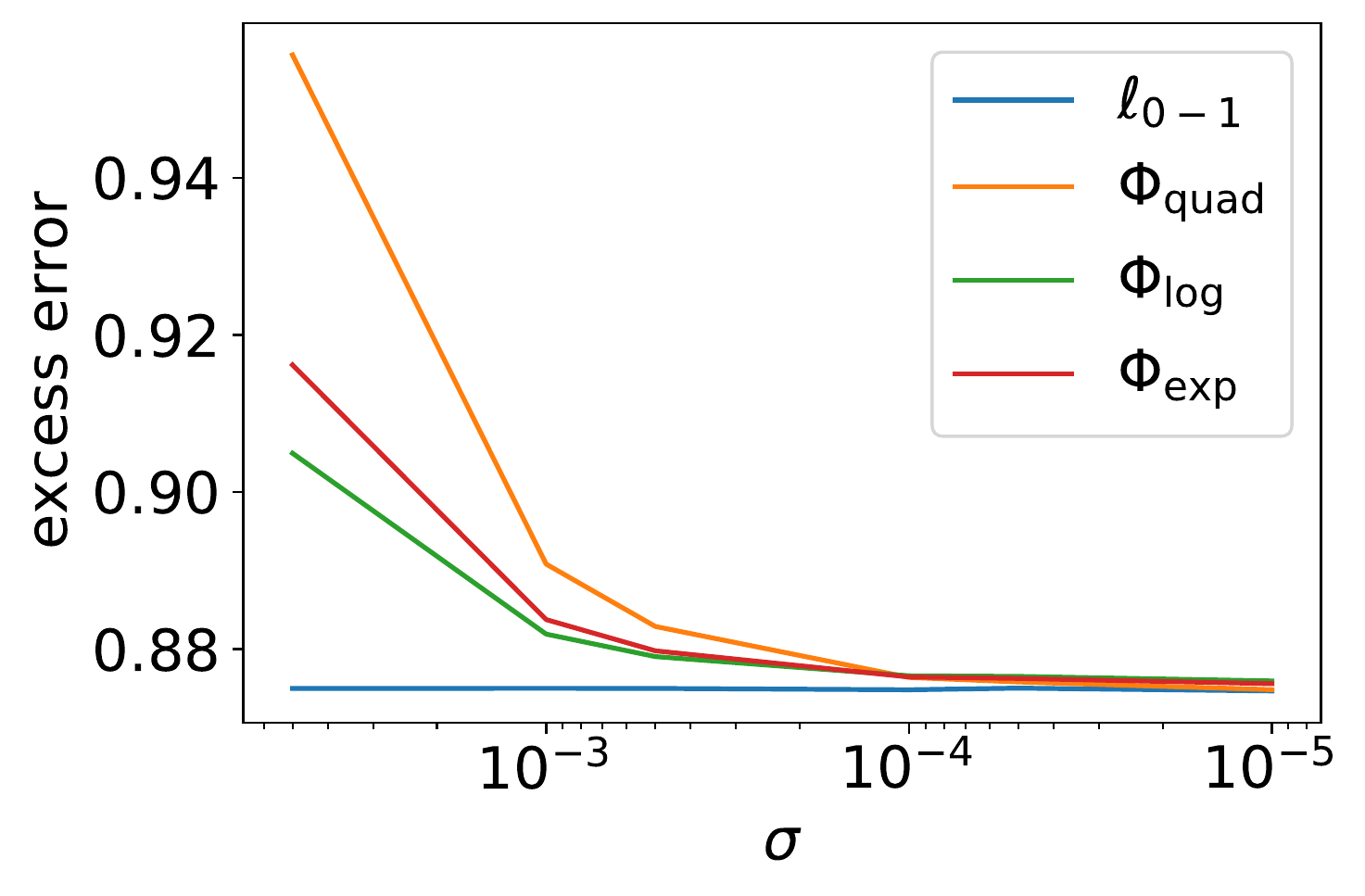}
\includegraphics[width=0.5\columnwidth]{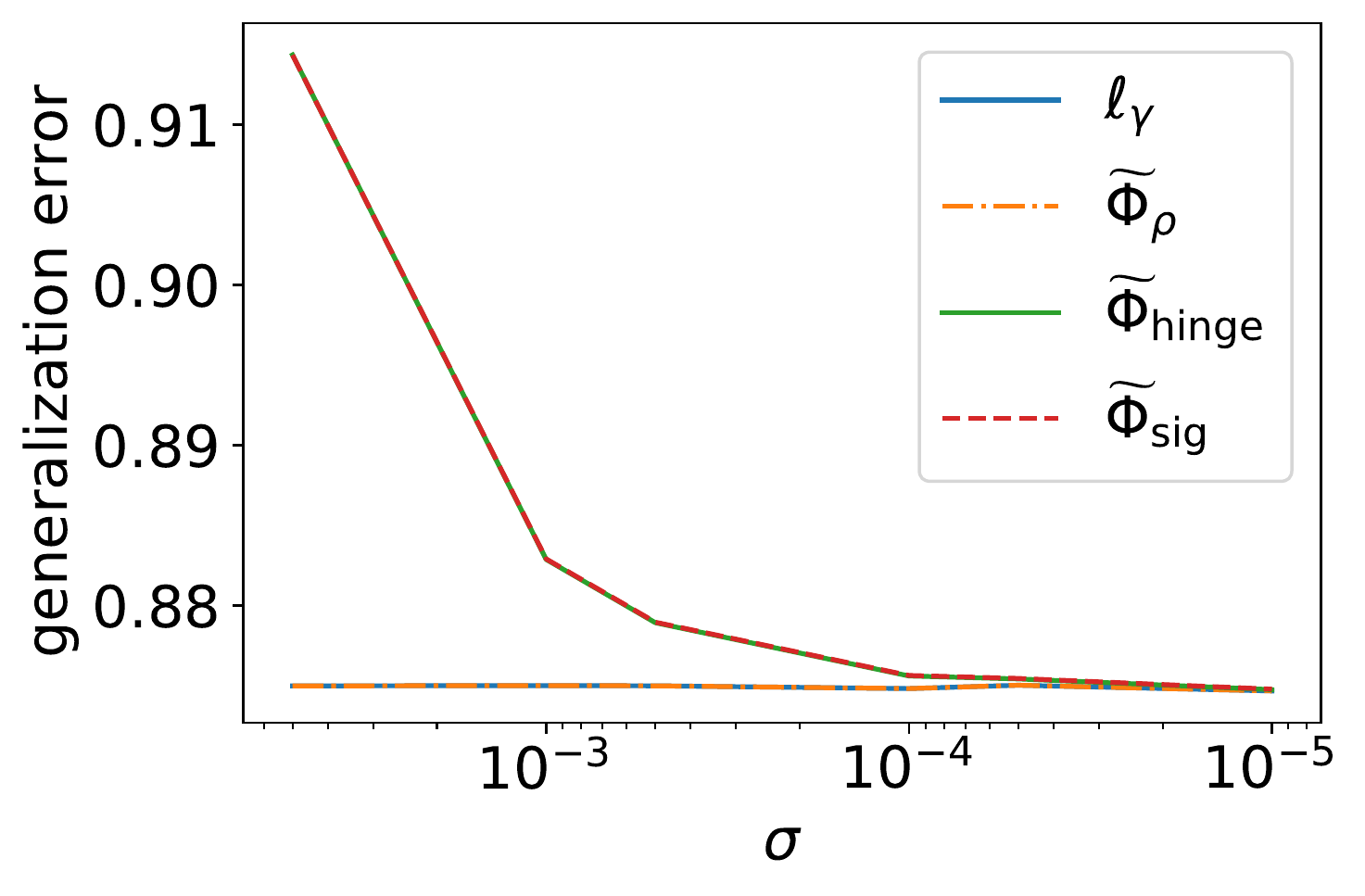}
\vskip -0.15in
\caption{Left: tightness of bound~\eqref{eq:non-adv-noise} in
  Section~\ref{sec:noise-non-adv}. Right: tightness of
  bound~\eqref{eq:rho-lin-est-adv},\eqref{eq:hinge-lin-est-adv} and
  \eqref{eq:sig-lin-est-adv} in Section~\ref{sec:adv-lin} and
  \ref{sec:noise-adv}.}
\label{fig:simulation}
\end{center}
\vskip -0.3in
\end{figure}
Here, we present experiments on simulated data to illustrate our
bounds and their tightness. We generate data points $x \in \mathbb{R}$
on $[-1, +1]$. All risks are approximated by their empirical
counterparts computed over $10^7$
i.i.d.\ samples.

\textbf{Non-adversarial.} To demonstrate the
tightness of our non-adversarial bounds, we consider a scenario where
the marginal distribution is symmetric about $x = 0$ with labels
flipped. With probability $\frac{1}{16}$, $(x, y)=(1, -1)$; with
probability $\frac{7}{16}$, the label is $+1$ and the data follows the
truncated normal distribution on $[\sigma, 1]$ with both mean and
standard deviation $\sigma$. We consider
$\Phi_{\mathrm{quad}}$, $\Phi_{\mathrm{log}}$ and
$\Phi_{\mathrm{exp}}$ defined in Table~\ref{tab:loss} of
Appendix~\ref{app:table}.  The distribution considered satisfies
Massart's noise condition with $\beta = \frac{1}{2}$. Thus, our
bound \eqref{eq:non-adv-noise} in Section~\ref{sec:noise-non-adv}
becomes $\sR_{\ell_{0-1}}(h)- \sR_{\ell_{0-1},\sH_{\mathrm{all}}}^*
\leq \sR_{\Phi}(h)- \sR_{\Phi,\sH_{\mathrm{all}}}^*$, for any $h\in
\sH_{\mathrm{all}}$ such that $\sR_{\Phi}(h) \leq
\sR_{\Phi,\sH_{\mathrm{all}}}^*+1$. All the minimal generalization
errors vanish in this case. As shown in Figure~\ref{fig:simulation},
for $h(x) = -5x$, the bounds corresponding to $\Phi_{\mathrm{quad}}$,
$\Phi_{\mathrm{log}}$ and $\Phi_{\mathrm{exp}}$ are all tight as
$\sigma \to 0$.

\textbf{Adversarial}.  To demonstrate the tightness of
our adversarial bounds, the distribution is modified as follows: with
probability $\frac{1}{16}$, $(x,y)=(1,-1)$; with probability
$\frac{1}{16}$, $(x,y)=(-1,+1)$; with probability $\frac{7}{8}$, the
label is $-1$ and the data follows the truncated normal distribution
on $[-1, \gamma-\sigma]$ with mean $\gamma-\sigma$ and standard
deviation $\sigma$. We set $\gamma=0.1$ and consider
$\wt{\Phi}_{\rho}$ with $\rho=1$, $\wt{\Phi}_{\mathrm{hinge}}$ and
$\wt{\Phi}_{\mathrm{sig}}$ with $k=1$.  The distribution considered
satisfies Massart's noise condition with $\beta = \frac{1}{2}$. Thus, our
bounds \eqref{eq:rho-lin-est-adv}, \eqref{eq:hinge-lin-est-adv} and
\eqref{eq:sig-lin-est-adv} in Table~\ref{tab:compare-adv} become
$\sR_{\ell_{\gamma}}(h) \leq \sR_{ \ \wt{\Phi}}(h)$, for any $h\in
\sH_{\mathrm{lin}}$. As shown in Figure~\ref{fig:simulation}, for
$h(x) = -5x$, the bounds corresponding to $\wt{\Phi}_{\rho}$,
$\wt{\Phi}_{\mathrm{hinge}}$ and $\wt{\Phi}_{\mathrm{sig}}$ are all
tight as $\sigma \to 0$.

\section{Conclusion}

We presented an exhaustive study of $\sH$-consistency estimation error
bounds, including a series of new guarantees for both the
non-adversarial zero-one loss function and the adversarial zero-one
loss function. Our hypothesis-dependent guarantees are significantly
stronger than the consistency or calibration ones.  Our results
include a series of theoretical and conceptual tools helpful for the
analysis of other loss functions and other hypothesis sets, including
multi-class classification or ranking losses.

\ignore{
\section*{Acknowledgements}
}

\bibliography{hcb}
\bibliographystyle{icml2022}

\newpage
\appendix
\onecolumn

\renewcommand{\contentsname}{Contents of Appendix}
\tableofcontents
\addtocontents{toc}{\protect\setcounter{tocdepth}{3}} 
\clearpage

\section{Related Work}
\label{app:related}
Bayes-consistency (also known as consistency) and excess error bounds between margin-based loss functions and the zero-one loss have been widely studied in the
literature
\citep{Zhang2003,bartlett2006convexity,steinwart2007compare,MohriRostamizadehTalwalkar2018}. Consistency studies the asymptotic relation between the surrogate excess error and the target excess error while excess error bounds study the quantitative relation between them and thus is stronger. They both consider the  hypothesis set of all measurable functions. The works of \citet{Zhang2003}, \citet{bartlett2006convexity}, and \citet{steinwart2007compare} studied consistency via the lens of calibration 
and showed that calibration and consistency are equivalent in the standard binary classification when considering the hypothesis set of all measurable functions.

The work of \citet{Zhang2003} analyzed how close to the optimal excess error of the zero-one loss can one reach via 
minimizers of convex surrogates.
\citet{bartlett2006convexity} extended the results in \citep{Zhang2003} and developed a general methodology for finding
quantitative bounds between the excess error associated with the zero-one loss and the excess error of margin-based surrogate loss functions for all distributions.
In a recent work, \citet{MohriRostamizadehTalwalkar2018} simplified these results and
provided different proofs for the excess error bounds of various
loss functions widely used in practice. Calibration and consistency analysis have also been extended to multi-class settings \citep{zhang2004statistical,tewari2007consistency} and to ranking problems \citep{uematsu2011theoretically,gao2015consistency}.

Bayes-consistency is not an appropriate notion when studying learning with a hypothesis set $\sH$ that is distinct from the family of all measurable functions. Therefore, a new hypothesis set-dependent notion namely, $\sH$-consistency, has been proposed and explored in the more recent literature \citep{long2013consistency,kuznetsov2014multi,zhang2020bayes}. In particular, \citet{long2013consistency} argued that $\sH$-consistency is a more useful notion than consistency by empirically showing that certain loss functions that are $\sH$-consistent but not Bayes consistent can perform significantly better than a loss function known to be Bayes consistent. The work of \citet{kuznetsov2014multi} extended the $\sH$-consistency results in \citep{long2013consistency} to the case of structured prediction and provided positive results for $\sH$-consistency of several multi-class ensemble algorithms.

In a recent work \citet{zhang2020bayes} investigated the empirical phenomenon in \citep{long2013consistency} and designed a class of piecewise linear scoring functions such that minimizing a surrogate that is not $\sH$-consistent over this larger class yields $\sH$-consistency of linear models. For linear predictors, more general margin-based properties of convex surrogate losses are also studied in \citep{long2011learning,ben2012minimizing}. Aiming for such margin based error guarantees, \citet{ben2012minimizing} argued that the hinge loss is optimal among convex losses.

Most recently, the notion of $\sH$-consistency along with $\sH$-calibration have also been studied in the context of adversarially robust classification \citep{bao2020calibrated,awasthi2021calibration}. In the adversarial scenario, in contrast to standard classification, the target loss is the adversarial zero-one loss \citep{goodfellow2014explaining,madry2017towards,carlini2017towards,tsipras2018robustness,shafahi2019adversarial,wong2020fast}. This corresponds to the worst zero-one loss incurred over an adversarial perturbation of $x$ within a $\gamma$-ball as measured in a norm, typically $\ell_p$ for $p\in[1,\plus\infty]$. The adversarial loss presents new challenges and makes the consistency analysis significantly more complex. 

The work of \citet{bao2020calibrated} initiated the study of $\sH$-calibration with respect to the adversarial zero-one loss for the linear models. They showed that convex surrogates are not calibrated and introduced a class of nonconvex margin-based surrogate losses. They then provided sufficient conditions for such nonconvex losses to be calibrated in the linear case. The work of \citet{awasthi2021calibration} extended the results in \citep{bao2020calibrated} to the general nonlinear hypothesis sets and pointed out that although $\sH$-calibration is a necessary condition of $\sH$-consistency, it is not sufficient in the adversarial scenario. They then proposed sufficient conditions which guarantee calibrated losses to be consistent in the setting of adversarially robust classification. 

All the above mentioned works either studied asymptotic properties (Bayes-consistency  or $\sH$-consistency) or studied quantitative relations when $\sH$ is the family of all measurable functions (excess error bounds). Instead, our work considers a hypothesis set-dependent quantitative relation between the surrogate estimation error and the target estimation error. This is significantly stronger than $\sH$-calibration or $\sH$-consistency and is also more informative than excess error bounds which correspond to a special case of our results with $\sH=\sH_{\mathrm{all}}$. As a by-product, our theory contributes more significant consistency results for the poorly understood setting of adversarial robustness. There have also been recent works on different theoretical aspects of adversarial robustness such as tension between the zero-one loss and the adversarial zero-one loss \citep{tsipras2018robustness,zhang2019theoretically}, computational bottlenecks for adversarial loss \citep{bubeck2018adversarial2,bubeck2018adversarial,awasthi2019robustness}, adversarial examples \citep{bartlett2021adversarial,bubeck2021single}, sample complexity of adversarial surrogate losses \citep{khim2018adversarial,cullina2018pac,YinRamchandranBartlett2019,montasser2019vc,awasthi2020adversarial},
computational complexity of adversarially robust linear classifiers \citep{diakonikolas2020complexity},
connections with PAC learning
\citep{montasser2020reducing,viallard2021pac}, perturbations beyond $\ell_p$ norm\citep{feige2015learning,feige2018robust,attias2018improved}, adversarial robustness optimization \citep{robey2021adversarial}, overparametrization \citep{bubeck2021universal} and Bayes optimality \citep{awasthi2021existence}.

\newpage
\section{Deferred Tables}
\label{app:table}
\begin{table*}[ht]
\caption{Loss functions and their minimizability gaps. In some cases,
  the minimizability gap coincides with the approximation error. For
  example, $\sM_{\Phi_{\mathrm{sig}},\sH_{\mathrm{lin}}}=
  \sR_{\Phi_{\mathrm{sig}},\sH_{\mathrm{lin}}}^*-\mathbb{E}_{X}\bracket*{1-\abs*{1-2\eta(x)}\tanh\paren*{k\paren*{W\norm*{x}_p+B}}}$
  coincides with the
  $\paren*{\Phi_{\mathrm{sig}},\sH_{\mathrm{lin}}}$-approximation
  error
  $\sR_{\Phi_{\mathrm{sig}},\sH_{\mathrm{lin}}}^*-\mathbb{E}_{X}\bracket*{1-\abs*{1-2\eta(x)}}$
  for $B = \plus \infty$;
  $\sM_{\Phi_{\mathrm{hinge}},\sH_{\mathrm{NN}}} =
  \sR_{\Phi_{\mathrm{hinge}},\sH_{\mathrm{NN}}}^*-\mathbb{E}_{X}\bracket*{1-\abs*{2\eta(x)-1}\min\curl*{\Lambda
      W\norm*{x}_p+\Lambda B,1}}$ coincides with the
  $\paren*{\Phi_{\mathrm{hinge}},\sH_{\mathrm{NN}}}$-approximation
  error
  $\sR_{\Phi_{\mathrm{hinge}},\sH_{\mathrm{NN}}}^*-\mathbb{E}_{X}\bracket*{1-\abs*{1-2\eta(x)}}$
  for $\Lambda B \geq 1$.  The detailed derivation is included in
  Appendix~\ref{app:derivation-non-adv}, \ref{app:derivation-adv}.}
    \label{tab:loss}
    \vskip -0.2in
\begin{center}
    \begin{small}
    \begin{tabular}{l|lll}
    \toprule
      Loss Functions & Definitions & $\sM_{\ell,\sH_{\mathrm{lin}}}$ & $\sM_{\ell,\sH_{\mathrm{NN}}}$\\
    \midrule
     Hinge & $\Phi_{\mathrm{hinge}}(t)=\max\curl*{0,1-t}$ & \eqref{eq:M-hinge-lin} & \eqref{eq:M-hinge-NN}\\
     Logistic & $\Phi_{\mathrm{log}}(t)=\log_2(1+e^{-t})$ & \eqref{eq:M-log-lin} & \eqref{eq:M-log-NN}\\
     Exponential & $\Phi_{\mathrm{exp}}(t)=e^{-t}$ & \eqref{eq:M-exp-lin} & \eqref{eq:M-exp-NN} \\ Quadratic  & $\Phi_{\mathrm{quad}}(t)=(1-t)^2\mathds{1}_{t\leq 1}$ & \eqref{eq:M-quad-lin} & \eqref{eq:M-quad-lin} \\
     Sigmoid & $\Phi_{\mathrm{sig}}(t)=1-\tanh(kt),~k>0$ & \eqref{eq:M-sig-lin} & \eqref{eq:M-sig-NN}\\
     $\rho$-Margin & $\Phi_{\rho}(t)=\min\curl*{1,\max\curl*{0,1-\frac{t}{\rho}}},~\rho>0$ & \eqref{eq:M-rho-lin} & \eqref{eq:M-hinge-NN}\\
     Sup-$\rho$-Margin & $\wt{\Phi}_{\rho}=\sup_{x'\colon \|x-x'\|_p\leq \gamma}\Phi_{\rho}(y h(x'))$ & \eqref{eq:M-rho-lin-adv} & \eqref{eq:M-rho-NN-adv}\\
     Zero-One & $\ell_{0-1}=\mathds{1}_{\sign(h(x))\neq y}$ & \eqref{eq:M-01-lin} & \eqref{eq:M-01-NN}\\
     Adversarial Zero-One & $\ell_{\gamma}=\sup_{x'\colon \|x-x'\|_p\leq
  \gamma}\mathds{1}_{y h(x') \leq 0}$ & \eqref{eq:M-01-lin-adv} & \eqref{eq:M-01-NN-adv}\\
    \bottomrule
    \end{tabular}
    \end{small}
    \end{center}
    \vskip -0.1in
\end{table*}

\begin{table*}[ht]
\caption{Non-adversarial $\sH_{\mathrm{lin}}$-estimation error
  transformation ($\epsilon=0)$ and $\sH_{\mathrm{lin}}$-consistency
  estimation error bounds. All the bounds are hypothesis set-dependent (parameter $B$ in $\sH_{\mathrm{lin}}$) and provide
  novel guarantees as discussed in Section~\ref{sec:non-adv-lin}. The
  minimizability gaps appearing in the bounds for the surrogates are
  concluded in Table~\ref{tab:loss}. The detailed derivation is
  included in Appendix~\ref{app:derivation-lin}.}
    \label{tab:compare_inverse}
    \vskip -0.2in
\begin{center}
    \begin{small}
    \begin{tabular}{l|lll}
    \toprule
      Surrogates & $\sT_{\Phi}(t),\, t\in [0,1]$   & $\sT_{\Phi}^{-1}(t),\, t
      \in\Rset_{+}$ & Bound \\
    \midrule
      Hinge & $\min \curl*{B, 1} \, t $  & $\frac{t}{\min\curl*{B, 1}}$ & \eqref{eq:hinge-lin-est} \\
      Logistic & $\begin{cases}
\frac{t+1}{2}\log_2(t+1)+\frac{1-t}{2}\log_2(1-t),\quad &  t\leq \frac{e^B-1}{e^{B}+1},\\
1-\frac{t+1}{2}\log_2(1+e^{-B})-\frac{1-t}{2}\log_2(1+e^B),\quad & t> \frac{e^B-1}{e^B+1}.
\end{cases}$ & upper bounded by $\begin{cases}
\sqrt{2t}, & t\leq \frac{1}{2}\paren*{\frac{e^B-1}{e^B+1}}^2,\\
2\paren*{\frac{e^B+1}{e^B-1}}\, t, & t> \frac{1}{2}\paren*{\frac{e^B-1}{e^B+1}}^2.
\end{cases}$ & \eqref{eq:log-lin-est}\\
      Exponential & $\begin{cases}
1-\sqrt{1-t^2}, & t\leq \frac{e^{2B}-1}{e^{2B}+1},\\
1-\frac{t+1}{2}e^{-B}-\frac{1-t}{2}e^B, & t> \frac{e^{2B}-1}{e^{2B}+1}.
\end{cases}$ & upper bounded by $\begin{cases}
\sqrt{2t}, & t\leq \frac{1}{2}\paren*{\frac{e^{2B}-1}{e^{2B}+1}}^2,\\
2\paren*{\frac{e^{2B}+1}{e^{2B}-1}}\, t, & t> \frac{1}{2}\paren*{\frac{e^{2B}-1}{e^{2B}+1}}^2.
\end{cases}$ & \eqref{eq:exp-lin-est}\\
      Quadratic & $\begin{cases}
t^2, & t\leq B,\\
2B \,t-B^2, & t> B.
\end{cases}$ & $\begin{cases}
\sqrt{t}, & t \leq B^2, \\
\frac{t}{2B}+\frac{B}{2}, & t > B^2.
\end{cases}$ & \eqref{eq:quad-lin-est}\\
      Sigmoid & $\tanh(kB) \, t$ & $\frac{t}{\tanh(kB)}$ & \eqref{eq:sig-lin-est}\\
      $\rho$-Margin & $\frac{\min\curl*{B,\rho}}{\rho} \, t$ & $\frac{\rho }{\min\curl*{B,\rho}} \, t$ & \eqref{eq:rho-lin-est}\\
    \bottomrule
    \end{tabular}
    \end{small}
    \end{center}
    \vskip -0.1in
\end{table*}

\begin{table*}[ht]
    \caption{Non-adversarial $\sH_{\mathrm{NN}}$-estimation error
      transformation ($\epsilon=0)$ and
      $\sH_{\mathrm{NN}}$-consistency estimation error bounds. All the
      bounds are hypothesis set-dependent (parameter $\Lambda$
      and $B$ in $\sH_{\mathrm{NN}}$) and provide novel guarantees as
      dicussed in Section~\ref{sec:non-adv-NN}. The minimizability
      gaps appearing in the bounds for the surrogates are concluded in
      Table~\ref{tab:loss}. The detailed derivation is included in
      Appendix~\ref{app:derivation-NN}.}
    \label{tab:compare-NN}
    \vskip 0.15in
\begin{center}
    \begin{small}
    \begin{tabular}{l|lll}
    \toprule
      Surrogates & $\sT_{\Phi}(t),\, t\in [0,1]$   & $\sT_{\Phi}^{-1}(t),\, t
      \in\Rset_{+}$ & Bound \\
    \midrule
      Hinge & $\min \curl*{\Lambda B, 1} \, t $  & $\frac{t}{\min\curl*{\Lambda B, 1}}$ & \eqref{eq:hinge-NN-est} \\
      Logistic & $\begin{cases}
\frac{t+1}{2}\log_2(t+1)+\frac{1-t}{2}\log_2(1-t),\quad & t\leq \frac{e^{\Lambda B}-1}{e^{\Lambda B}+1},\\
1-\frac{t+1}{2}\log_2(1+e^{-\Lambda B})-\frac{1-t}{2}\log_2(1+e^{\Lambda B}),\quad & t> \frac{e^{\Lambda B}-1}{e^{\Lambda B}+1}.
\end{cases}$ & upper bounded by $\begin{cases}
\sqrt{2t}, & t\leq \frac{1}{2}\paren*{\frac{e^{\Lambda B}-1}{e^{\Lambda B}+1}}^2,\\
2\paren*{\frac{e^{\Lambda B}+1}{e^{\Lambda B}-1}}\, t, & t> \frac{1}{2}\paren*{\frac{e^{\Lambda B}-1}{e^{\Lambda B}+1}}^2.
\end{cases}$ & \eqref{eq:log-NN-est}\\
      Exponential & $\begin{cases}
1-\sqrt{1-t^2}, & t\leq \frac{e^{2\Lambda B}-1}{e^{2\Lambda B}+1},\\
1-\frac{t+1}{2}e^{-\Lambda B}-\frac{1-t}{2}e^{\Lambda B}, & t> \frac{e^{2\Lambda B}-1}{e^{2\Lambda B}+1}.
\end{cases}$ & upper bounded by $\begin{cases}
\sqrt{2t}, & t\leq \frac{1}{2}\paren*{\frac{e^{2\Lambda B}-1}{e^{2B}+1}}^2,\\
2\paren*{\frac{e^{2\Lambda B}+1}{e^{2\Lambda B}-1}}\, t, & t> \frac{1}{2}\paren*{\frac{e^{2\Lambda B}-1}{e^{2\Lambda B}+1}}^2.
\end{cases}$ & \eqref{eq:exp-NN-est}\\
      Quadratic & $\begin{cases}
t^2,~t\leq \Lambda B,\\
2\Lambda B t-(\Lambda B)^2,~t> \Lambda B.
\end{cases}$ & $\begin{cases}
\sqrt{t}, & t \leq (\Lambda B)^2 \\
\frac{t}{2\Lambda B}+\frac{\Lambda B}{2}, & t > (\Lambda B)^2
\end{cases}$ & \eqref{eq:quad-NN-est}\\
      Sigmoid & $\tanh(k\Lambda B) \, t$ & $\frac{t}{\tanh(k\Lambda B)}$ & \eqref{eq:sig-NN-est}\\
      $\rho$-Margin & $\frac{\min\curl*{\Lambda B,\rho}}{\rho} \, t$ & $\frac{\rho }{\min\curl*{\Lambda B,\rho}} \, t$ & \eqref{eq:rho-NN-est}\\
    \bottomrule
    \end{tabular}
    \end{small}
    \end{center}
    \vskip -0.1in
\end{table*}

\begin{table}[t]
\caption{Adversarial $\sH$-consistency estimation error bounds. They are completely new consistency estimation error bounds in the adversarial setting and can turn into more significant $\e$-consistency results. The minimizability gaps appearing in the bounds for the surrogates are concluded in Table~\ref{tab:loss}. The detailed derivation is included in Appendix~\ref{app:derivation-adv}, \ref{app:derivation-adv_noise}.}
    \label{tab:compare-adv}
    \vskip -0.4in
\begin{center}
    \begin{small}
    \begin{tabular}{l|lll}
    \toprule
      Surrogates & Bound ($\sH_{\mathrm{lin}}$) & Bound ($\sH_{\mathrm{NN}}$) & Distribution set\\
    \midrule
     $\wt{\Phi}_{\rho}$ & \eqref{eq:rho-lin-est-adv} & \eqref{eq:rho-NN-est-adv} & All distributions \\
     $\wt{\Phi}_{\mathrm{hinge}}$ & \eqref{eq:hinge-lin-est-adv} & \eqref{eq:hinge-NN-est-adv} & Massart’s noise\\
     $\wt{\Phi}_{\mathrm{sig}}$ & \eqref{eq:sig-lin-est-adv} & \eqref{eq:sig-NN-est-adv} & Massart’s noise\\
    \bottomrule
    \end{tabular}
    \end{small}
    \end{center}
    \vskip -0.3in
\end{table}

\newpage
\section{Deferred Theorems}
\label{app:theorems}

\begin{theorem}[\textbf{Non-adversarial distribution-dependent $\Psi$-bound}]
\label{Thm:excess_bounds_Psi_01_general}
Suppose that $\sH$ satisfies the condition of
Lemma~\ref{lemma:explicit_assumption_01} and that $\Phi$ is a
margin-based loss function. Assume there exist a convex function
$\Psi\colon \mathbb{R_{+}} \to \Rset$ with $\Psi(0)=0$ and
$\epsilon\geq0$ such that the following holds for any $x\in \sX$:
\begin{align}
\label{eq:condition_Psi_general}
    &\Psi\paren*{\tri*{2 \abs*{\Delta \eta(x)}}_{\e}}
    \leq \inf_{h \in \ov \sH}\Delta\sC_{\Phi,\sH}(h,x).
\end{align}
Then, for any hypothesis $h \in \sH$,
\ifdim\columnwidth=\textwidth
{
\begin{equation}
     \Psi\paren*{\sR_{\ell_{0-1}}(h)- \sR_{\ell_{0-1},\sH}^*+\sM_{\ell_{0-1},\sH}}
     \leq  \sR_{\Phi}(h)-\sR_{\Phi,\sH}^*+\sM_{\Phi,\sH}+\max\curl*{0,\Psi(\e)}.
\end{equation}
}
\else
{
\begin{multline}
     \Psi\paren*{\sR_{\ell_{0-1}}(h)- \sR_{\ell_{0-1},\sH}^*+\sM_{\ell_{0-1},\sH}}\\
     \leq  \sR_{\Phi}(h)-\sR_{\Phi,\sH}^*+\sM_{\Phi,\sH}+\max\curl*{0,\Psi(\e)}.
\end{multline}
}
\fi
\end{theorem}
\begin{theorem}[\textbf{Non-adversarial distribution-dependent $\Gamma$-bound}]
\label{Thm:excess_bounds_Gamma_01_general}
Suppose that $\sH$ satisfies the condition of
Lemma~\ref{lemma:explicit_assumption_01} and that $\Phi$ is a
margin-based loss function. Assume there exist a non-negative and
non-decreasing concave function $\Gamma\colon \mathbb{R_{+}}\to \Rset$
and $\epsilon\geq0$ such that the following holds for any $x\in \sX$:
\begin{align}
\label{eq:condition_Gamma_general}
    \tri*{2 \abs*{\Delta \eta(x)}}_{\e} \leq \Gamma\paren*{\inf_{h\in \ov \sH}\Delta\sC_{\Phi,\sH}(h,x)}.
\end{align}
Then, for any hypothesis $h\in\sH$,
\ifdim\columnwidth=\textwidth
{
\begin{equation}
     \sR_{\ell_{0-1}}(h)- \sR_{\ell_{0-1},\sH}^*
     \leq  \Gamma\paren*{\sR_{\Phi}(h)-\sR_{\Phi,\sH}^*+\sM_{\Phi,\sH}}
     -\sM_{\ell_{0-1},\sH}+\epsilon.
\end{equation}
}
\else
{
\begin{multline}
     \sR_{\ell_{0-1}}(h)- \sR_{\ell_{0-1},\sH}^*
     \leq  \Gamma\paren*{\sR_{\Phi}(h)-\sR_{\Phi,\sH}^*+\sM_{\Phi,\sH}}\\
     -\sM_{\ell_{0-1},\sH}+\epsilon.
\end{multline}
}
\fi
\end{theorem}
\begin{theorem}[\textbf{Adversarial distribution-dependent $\Psi$-bound}]
\label{Thm:excess_bounds_Psi_01_general_adv}
Suppose that $\sH$ is symmetric and that $\wt{\Phi}$ is a
supremum-based margin loss function. Assume there exist a convex
function $\Psi\colon \mathbb{R_{+}}\to \Rset$ with $\Psi(0)=0$ and
$\epsilon\geq0$ such that the following holds for any $x\in \sX$:
\begin{equation}
\mspace{-8mu}
\label{eq:condition_Psi_general_adv}
\begin{aligned}
&\Psi\paren*{\tri*{\abs*{\Delta \eta(x)}+1/2}_{\e}}\leq
\inf_{h\in \ov \sH_\gamma}
\Delta\sC_{\Phi,\sH}(h,x),\\
&\Psi\paren*{\tri*{2\Delta \eta(x)}_{\e}}\leq
\inf_{h\in\sH: \ov h_\gamma(x)<0}
\Delta\sC_{\Phi,\sH}(h,x),\\
&\Psi\paren*{\tri*{-2\Delta \eta(x)}_{\e}}\leq
\inf_{h\in\sH: \uv h_\gamma(x)>0}
\Delta\sC_{\Phi,\sH}(h,x).
\end{aligned}
\mspace{-40mu}
\end{equation}
Then, for any hypothesis $h\in\sH$,
\ifdim\columnwidth=\textwidth
{
\begin{equation}
     \Psi\paren*{\sR_{\ell_{\gamma}}(h)- \sR_{\ell_{\gamma},\sH}^*+\sM_{\ell_{\gamma},\sH}}
     \leq  \sR_{\wt{\Phi}}(h)-\sR_{\wt{\Phi},\sH}^*+\sM_{\wt{\Phi},\sH}+\max\curl*{0,\Psi(\e)}.
\end{equation}
}
\else
{
\begin{multline}
\label{eq:bound_Psi_01_adv}
     \Psi\paren*{\sR_{\ell_{\gamma}}(h)- \sR_{\ell_{\gamma},\sH}^*+\sM_{\ell_{\gamma},\sH}}\\
     \leq  \sR_{\wt{\Phi}}(h)-\sR_{\wt{\Phi},\sH}^*+\sM_{\wt{\Phi},\sH}+\max\curl*{0,\Psi(\e)}.
\end{multline}
}
\fi
\end{theorem}
\begin{theorem}[\textbf{Adversarial distribution-dependent $\Gamma$-bound}]
\label{Thm:excess_bounds_Gamma_01_general_adv}
Suppose that $\sH$ is symmetric and that $\wt{\Phi}$ is a
supremum-based margin loss function. Assume there exist a non-negative
and non-decreasing concave function $\Gamma\colon \mathbb{R_{+}}\to
\Rset$ and $\epsilon\geq0$ such that the following holds for any $x\in
\sX$:
\begin{equation}
\mspace{-8mu}
\label{eq:condition_Gamma_general_adv}
\begin{aligned}
&\tri*{\abs*{\Delta \eta(x)}+1/2}_{\e}\leq \Gamma\paren*{
\inf_{h\in \ov \sH_\gamma}
\Delta\sC_{\Phi,\sH}(h,x)},\\
&\tri*{2\Delta \eta(x)}_{\e}\leq \Gamma\paren*{
\inf_{h\in\sH: \ov h_\gamma(x)<0}
\Delta\sC_{\Phi,\sH}(h,x)},\\
&\tri*{-2\Delta \eta(x)}_{\e}\leq \Gamma\paren*{
\inf_{h\in\sH: \uv h_\gamma(x)>0}
\Delta\sC_{\Phi,\sH}(h,x)}.
\end{aligned}
\mspace{-40mu}
\end{equation}
Then, for any hypothesis $h\in\sH$,
\ifdim\columnwidth=\textwidth
{
\begin{equation}
     \sR_{\ell_{\gamma}}(h)- \sR_{\ell_{\gamma},\sH}^*
     \leq  \Gamma\paren*{\sR_{\wt{\Phi}}(h)-\sR_{\wt{\Phi},\sH}^*+\sM_{\wt{\Phi},\sH}}-\sM_{\ell_{\gamma},\sH}+\epsilon.
\end{equation}
}
\else
{
\begin{multline}
\label{eq:bound_Gamma_01_adv}
     \sR_{\ell_{\gamma}}(h)- \sR_{\ell_{\gamma},\sH}^*\\
     \leq  \Gamma\paren*{\sR_{\wt{\Phi}}(h)-\sR_{\wt{\Phi},\sH}^*+\sM_{\wt{\Phi},\sH}}-\sM_{\ell_{\gamma},\sH}+\epsilon.
\end{multline}
}
\fi
\end{theorem}

\begin{restatable}[\textbf{Distribution-independent $\Gamma$-bound}]{theorem}{ExcessBoundsGammaUniform}
\label{Thm:excess_bounds_Gamma_uniform}
Suppose that $\sH$ satisfies the condition of
Lemma~\ref{lemma:explicit_assumption_01} and that $\Phi$ is a
margin-based loss function. Assume there exist a non-negative and
non-decreasing concave function $\Gamma\colon \mathbb{R_{+}}\to \Rset$
and $\epsilon\geq0$ such that the following holds for any for any
$t\in\left[1/2,1\right]\colon$
\begin{align*}
\tri*{2t-1}_{\e}\leq \Gamma\paren*{\inf_{x\in \sX,h\in\sH:h(x)< 0}\Delta\sC_{\Phi,\sH}(h,x,t)}. 
\end{align*}
Then, for any hypothesis $h\in\sH$ and any distribution,
\ifdim\columnwidth=\textwidth
{
\begin{equation}
\label{eq:bound_Gamma_01}
     \sR_{\ell_{0-1}}(h)- \sR_{\ell_{0-1},\sH}^*
     \leq  \Gamma\paren*{\sR_{\Phi}(h)-\sR_{\Phi,\sH}^*+\sM_{\Phi,\sH}}
     -\sM_{\ell_{0-1},\sH}+\epsilon.
\end{equation}
}
\else
{
\begin{multline}
\label{eq:bound_Gamma_01}
     \sR_{\ell_{0-1}}(h)- \sR_{\ell_{0-1},\sH}^*
     \leq  \Gamma\paren*{\sR_{\Phi}(h)-\sR_{\Phi,\sH}^*+\sM_{\Phi,\sH}}\\
     -\sM_{\ell_{0-1},\sH}+\epsilon.
\end{multline}
}
\fi
\end{restatable}

\begin{restatable}[\textbf{Adversarial distribution-independent $\Gamma$-bound}]{theorem}{ExcessBoundsGammaUniformAdv}
\label{Thm:excess_bounds_Gamma_uniform-adv}
Suppose that $\sH$ is symmetric and that $\wt{\Phi}$ is a supremum-based margin loss function. Assume there exist a non-negative and non-decreasing concave function $\Gamma\colon \mathbb{R_{+}}\to \Rset$ and $\epsilon\geq0$ such that the following holds for any for any $t\in\left[1/2,1\right]\colon$
\begin{align*}
&\tri*{t}_{\e}  \leq \Gamma\paren*{\inf_{x\in \sX,h\in \ov \sH_\gamma\subsetneqq \sH}\Delta\sC_{\wt{\Phi},\sH}(h,x,t)},\\
&\tri*{2t-1}_{\e} \leq \Gamma\paren*{\inf_{x\in \sX,h\in\sH\colon \ov h_\gamma(x)< 0}\Delta\sC_{\wt{\Phi},\sH}(h,x,t)}.
\end{align*}
Then, for any hypothesis $h\in\sH$ and any distribution,
\ifdim\columnwidth=\textwidth
{
\begin{equation}
\label{eq:bound_Gamma_01_adv}
     \sR_{\ell_{\gamma}}(h)- \sR_{\ell_{\gamma},\sH}^*
     \leq  \Gamma\paren*{\sR_{\wt{\Phi}}(h)-\sR_{\wt{\Phi},\sH}^*+\sM_{\wt{\Phi},\sH}}-\sM_{\ell_{\gamma},\sH}+\epsilon.
\end{equation}
}
\else
{
\begin{multline}
\label{eq:bound_Gamma_01_adv}
     \sR_{\ell_{\gamma}}(h)- \sR_{\ell_{\gamma},\sH}^*\\
     \leq  \Gamma\paren*{\sR_{\wt{\Phi}}(h)-\sR_{\wt{\Phi},\sH}^*+\sM_{\wt{\Phi},\sH}}-\sM_{\ell_{\gamma},\sH}+\epsilon.
\end{multline}
}
\fi
\end{restatable}

\section{Proof of Theorem~\ref{Thm:excess_bounds_Psi} and Theorem~\ref{Thm:excess_bounds_Gamma}}
\label{app:excess_bounds}

\ExcessBoundsPsi*
\begin{proof}
For any $h\in \sH$, since $\Psi\paren*{\Delta\sC_{\ell_2,\sH}(h,x)\mathds{1}_{\Delta\sC_{\ell_2,\sH}(h,x)>\epsilon}}\leq \Delta\sC_{\ell_1,\sH}(h,x)$ for all $x\in \sX$, we have
    \begin{align*}
       &\Psi\paren*{\sR_{\ell_2}(h)-\sR_{\ell_2,\sH}^*+\sM_{\ell_2,\sH}}\\
       &=\Psi\paren*{\mathbb{E}_{X}  \bracket*{\sC_{\ell_2}(h,x)-\sC^*_{\ell_2,\sH}(x)}}\\
       &=\Psi\paren*{\mathbb{E}_{X}  \bracket*{\Delta\sC_{\ell_2,\sH}(h,x)}}\\
       &\leq\mathbb{E}_{X} \bracket*{\Psi\paren*{\Delta\sC_{\ell_2,\sH}(h,x)}} & (\text{Jensen's ineq.}) \\
        &=\mathbb{E}_{X}  \bracket*{\Psi\paren*{\Delta\sC_{\ell_2,\sH}(h,x)\mathds{1}_{\Delta\sC_{\ell_2,\sH}(h,x)>\epsilon}+\Delta\sC_{\ell_2,\sH}(h,x)\mathds{1}_{\Delta\sC_{\ell_2,\sH}(h,x)\leq\epsilon}}}\\ 
       &\leq\mathbb{E}_{X} \bracket*{\Psi\paren*{\Delta\sC_{\ell_2,\sH}(h,x)\mathds{1}_{\Delta\sC_{\ell_2,\sH}(h,x)>\epsilon}}+\Psi\paren*{\Delta\sC_{\ell_2,\sH}(h,x)\mathds{1}_{\Delta\sC_{\ell_2,\sH}(h,x)\leq\epsilon}}}  &\paren*{\Psi(0)\geq 0}\\ 
       &\leq\mathbb{E}_{X} \bracket*{\Delta\sC_{\ell_1,\sH}(h,x)}+\sup_{t\in[0,\epsilon]}\Psi(t) &\paren*{\text{assumption}}\\
       &=\sR_{\ell_1}(h)-\sR_{\ell_1,\sH}^*+\sM_{\ell_1,\sH}+\max\curl*{\Psi(0),\Psi(\e)}, &\paren*{\text{convexity of $\Psi$}}
    \end{align*}
    which proves the theorem.
\end{proof}

\ExcessBoundsGamma*
\begin{proof}
For any $h\in \sH$, since $\Delta\sC_{\ell_2,\sH}(h,x)\mathds{1}_{\Delta\sC_{\ell_2,\sH}(h,x)>\epsilon}\leq \Gamma\paren*{\Delta\sC_{\ell_1,\sH}(h,x)}$ for all $ x\in \sX$, we have
    \begin{align*}
       &\sR_{\ell_2}(h)-\sR_{\ell_2,\sH}^*+\sM_{\ell_2,\sH}\\
       &=\mathbb{E}_{X}  \bracket*{\sC_{\ell_2}(h,x)-\sC^*_{\ell_2,\sH}(x)}\\
       &=\mathbb{E}_{X}  \bracket*{\Delta\sC_{\ell_2,\sH}(h,x)}\\
        &=\mathbb{E}_{X}  \bracket*{\Delta\sC_{\ell_2,\sH}(h,x)\mathds{1}_{\Delta\sC_{\ell_2,\sH}(h,x)>\epsilon}+\Delta\sC_{\ell_2,\sH}(h,x)\mathds{1}_{\Delta\sC_{\ell_2,\sH}(h,x)\leq\epsilon}}\\ 
        &\leq\mathbb{E}_{X}  \bracket*{\Gamma\paren*{\Delta\sC_{\ell_1,\sH}(h,x)}}+\epsilon &\paren*{\text{assumption}}\\ 
        &\leq\Gamma\paren*{\mathbb{E}_{X}  \bracket*{\Delta\sC_{\ell_1,\sH}(h,x)}}+\epsilon &\paren*{\text{concavity of $\Gamma$}}\\ 
       &=\Gamma\paren*{\sR_{\ell_1}(h)-\sR_{\ell_1,\sH}^*+\sM_{\ell_1,\sH}}+\epsilon,
    \end{align*}
    which proves the theorem.
\end{proof}

\section{Proof of Lemma~\ref{lemma:explicit_assumption_01} and Lemma~\ref{lemma:explicit_assumption_01_adv}}
\label{app:explicit_assumption}
\ExplicitAssumption*
\begin{proof}
By the definition, the conditional $\ell_{0-1}$-risk is 
\begin{align*}
\sC_{\ell_{0-1}}(h,x) 
& = \eta(x)\mathds{1}_{h(x)< 0}+(1-\eta(x))\mathds{1}_{h(x)\geq 0} \\
& = 
\begin{cases}
\eta(x) & \text{if} ~ h(x)<0,\\
1-\eta(x) & \text{if} ~ h(x)\geq 0.
\end{cases}
\end{align*}
By the assumption, for any $x\in \sX$, there exists $h^*\in \sH$ such that $\sign(h^*(x))=\sign(\Delta \eta(x))$, where $\Delta \eta(x)$ is the Bayes classifier such that $\sC_{\ell_{0-1}}\paren*{\Delta \eta(x),x}=\sC^*_{\ell_{0-1},\sH_{\mathrm{all}}}(x)=\min\curl*{\eta(x),1-\eta(x)}$. Therefore,  the optimal conditional $\ell_{0-1}$-risk is
\begin{align*}
\sC^*_{\ell_{0-1},\sH}(x)= \sC_{\ell_{0-1}}\paren*{h^*,x}=\sC_{\ell_{0-1}}\paren*{\Delta \eta(x),x}=\min\curl*{\eta(x),1-\eta(x)}
\end{align*}
which proves the first part of lemma. By the definition,
\begin{align*}
\Delta\sC_{\ell_{0-1},\sH}(h,x)
&=\sC_{\ell_{0-1}}(h,x) - \sC^*_{\ell_{0-1},\sH}(x)\\
& = \eta(x)\mathds{1}_{h(x)< 0}+(1-\eta(x))\mathds{1}_{h(x)\geq 0} - \min\curl*{\eta(x),1-\eta(x)}\\
& = 
\begin{cases}
2 \abs*{\Delta \eta(x)},  & h \in \ov \sH, \\
0, & \text{otherwise} .
\end{cases}
\end{align*}
This leads to
\begin{align*}
\tri*{\Delta\sC_{\ell_{0-1},\sH}(h,x)}_{\e}=\tri*{2 \abs*{\Delta \eta(x)}}_{\e}\mathds{1}_{h \in \ov \sH}\,.
\end{align*}
\end{proof}

\ExplicitAssumptionAdv*
\begin{proof}
By the definition, the conditional $\ell_{\gamma}$-risk is
\begin{align*}
  \sC_{\ell_{\gamma}}(h,x)
  &=\eta(x) \mathds{1}_{\left\{\uv h_\gamma(x)\leq 0\right\}}+(1-\eta(x)) \mathds{1}_{\left\{\ov h_\gamma(x)\geq 0\right\}}\\
  &=\begin{cases}
   1 & \text{if} ~ h\in \ov \sH_\gamma,\\
   \eta(x) & \text{if} ~ \ov h_\gamma(x)<0,\\
   1-\eta(x) & \text{if} ~ \uv h_\gamma(x)> 0.\\
  \end{cases}
\end{align*}
Since $\sH$ is symmetric, for any $x \in \sX$, either there exists $h\in\sH$ such that $\uv h_\gamma(x)>0$, or $\ov \sH_\gamma=\sH$. When $\ov \sH_\gamma=\sH$, $\{h\in \sH:\ov h_\gamma(x)<0\}$ and $\{h\in \sH:\uv h_\gamma(x)>0\}$ are both empty sets. Thus $\sC^*_{\ell_{\gamma},\sH}(x)=1$.  When $\ov \sH_\gamma\neq\sH$, there exists $h\in \sH$ such that $\sC_{\ell_{\gamma}}(h,x)=\min\curl*{\eta(x),1-\eta(x)}=\sC^*_{\ell_{\gamma},\sH}(x)$. Therefore, the minimal conditional $\ell_{\gamma}$-risk is
\begin{align*}
     \sC^*_{\ell_{\gamma},\sH}(x)=\begin{cases}
        1, & \ov \sH_\gamma=\sH\,,\\
     	\min\curl*{\eta(x),1-\eta(x)}, & \ov \sH_\gamma\neq\sH\,.
     \end{cases}
\end{align*}
When $\ov \sH_\gamma=\sH$, $\sC_{\ell_{\gamma}}(h,x)\equiv1$, which implies that $\Delta\sC_{\ell_{\gamma},\sH}(h,x)\equiv0$. For $h\in \ov \sH_\gamma\subsetneqq \sH$, $\Delta\sC_{\ell_{\gamma},\sH}(h,x)=1-\min\curl*{\eta(x),1-\eta(x)}=\abs*{\Delta\eta(x)}+1/2$; for $h\in\sH$ such that $\ov h_\gamma(x)<0$, we have $\Delta\sC_{\ell_{\gamma},\sH}(h,x)=\eta(x)-\min\curl*{\eta(x),1-\eta(x)}=\max\curl*{0,2\Delta \eta(x)}$; for $h\in \sH$ such that $\uv h_\gamma(x)>0$, $\Delta\sC_{\ell_{\gamma},\sH}(h,x)=1-\eta(x)-\min\curl*{\eta(x),1-\eta(x)}=\max\curl*{0,-2\Delta \eta(x)}$. Therefore,
\begin{align*}
  \Delta\sC_{\ell_{\gamma},
    \sH}(h,x)=
    \begin{cases}
    \abs*{\Delta\eta(x)}+1/2 &  h\in \ov \sH_\gamma\subsetneqq \sH,\\
    \max\curl*{0,2\Delta \eta(x)} & \ov h_\gamma(x)<0,\\
    \max\curl*{0,-2\Delta \eta(x)} & \uv h_\gamma(x)>0,\\
    0 & \text{otherwise}.
    \end{cases}
\end{align*}
This leads to
\begin{align*}
\tri*{\Delta\sC_{\ell_{\gamma},\sH}(h,x)}_{\e}
=
\begin{cases}
\tri*{\abs*{\Delta \eta(x)}+\frac12}_{\e} 
&h \in \ov \sH_\gamma\subsetneqq \sH\\
\tri*{2\Delta \eta(x)}_{\e}
& \ov h_\gamma(x)<0\\
\tri*{-2\Delta \eta(x)}_{\e}
&\uv h_\gamma(x)>0 \\
0 
&\text{otherwise}
\end{cases}
\end{align*}
\end{proof}

\section{Comparison with Previous Results when \texorpdfstring{$\sH=\sH_{\mathrm{all}}$}{all}}
\label{app:compare-all-measurable}
\subsection{Comparison with \texorpdfstring{\citep[Theorem 4.7]{MohriRostamizadehTalwalkar2018}}{ref}}
Assume $\Phi$ is convex and non-increasing. For any $x\in \sX$, by the convexity, we have
\begin{align}
\label{eq:convex}
\sC_{\Phi}(h,x)=\eta(x)\Phi(h(x))+(1-\eta(x))\Phi(-h(x))\geq \Phi(2\Delta \eta(x)h(x)).
\end{align}
Then,
\begin{align*}
\inf_{h\in \ov{\sH_{\mathrm{all}}}}\Delta\sC_{\Phi,\sH_{\mathrm{all}}}(h,x)
&\geq
\inf_{h\in\sH_{\mathrm{all}}:2\Delta \eta(x)h(x)\leq 0}\Delta\sC_{\Phi,\sH_{\mathrm{all}}}(h,x) & (h\in \ov{\sH_{\mathrm{all}}}  \implies h(x)\Delta \eta(x)\leq 0)\\
& \geq \inf_{h\in\sH_{\mathrm{all}}:2\Delta \eta(x)h(x)\leq 0}\Phi(2\Delta \eta(x)h(x))-\sC^*_{\Phi,\sH_{\mathrm{all}}}(x) & (\text{eq.~}\eqref{eq:convex})\\
&= \sC_{\Phi}(0,x)- \sC^*_{\Phi,\sH_{\mathrm{all}}}(x) & (\text{$\Phi$ is non-increasing}).
\end{align*}
Thus the condition of Theorem~4.7 in \citep{MohriRostamizadehTalwalkar2018} implies the condition in Corollary~\ref{cor:excess_bounds_Psi_01_M}:
\begin{align*}
\abs*{\Delta \eta(x)}\leq c~\bracket*{\sC_{\Phi}(0,x)- \sC^*_{\Phi,\sH_{\mathrm{all}}}(x)}^{\frac1s},\; \forall x\in \sX \implies
\abs*{\Delta \eta(x)}\leq c~\inf_{h\in \ov{\sH_{\mathrm{all}}}}\bracket*{\Delta\sC_{\Phi,\sH_{\mathrm{all}}}(h,x)}^{\frac1s},\; \forall x\in \sX.
\end{align*}
Therefore, Theorem~4.7 in \citep{MohriRostamizadehTalwalkar2018} is a special case of Corollary~\ref{cor:excess_bounds_Psi_01_M}.

\subsection{Comparison with \texorpdfstring{\citep[Theorem 1.1]{bartlett2006convexity}}{ref}}
We show that the $\psi$-transform in \citep{bartlett2006convexity} verifies the condition in Corollary~\ref{cor:excess_bounds_Psi_01_B} for all distributions.
First, by Definition 2 in \citep{bartlett2006convexity}, we know that $\psi$ is convex, $\psi(0)=0$ and $\psi\leq \wt{\psi}$. 
Then,
\begin{align*}
\psi\paren*{2 \abs*{\Delta \eta(x)}}
&\leq \wt{\psi} \paren*{2 \abs*{\Delta \eta(x)}} & \paren*{\psi\leq \wt{\psi}} \\
&= \inf_{\alpha\leq0}\paren*{\max\curl*{\eta(x),1-\eta(x)}\Phi(\alpha)+\min\curl*{\eta(x),1-\eta(x)}\Phi(-\alpha)}\\
&-\inf_{\alpha\in \Rset}\paren*{\max\curl*{\eta(x),1-\eta(x)}\Phi(\alpha)+\min\curl*{\eta(x),1-\eta(x)}\Phi(-\alpha)} &\paren*{\text{def. of } \wt{\psi}}\\
&= \inf_{\alpha\Delta \eta(x)\leq0}\paren*{\eta(x)\Phi(\alpha)+\paren*{1-\eta(x)}\Phi(-\alpha)}-\inf_{\alpha\in \Rset}\paren*{\eta(x)\Phi(\alpha)+\paren*{1-\eta(x)}\Phi(-\alpha)} & \paren*{\text{symmetry}}\\
&= \inf_{h\in\sH_{\mathrm{all}}:h(x)\Delta \eta(x)\leq 0}\Delta\sC_{\Phi,\sH_{\mathrm{all}}}(h,x)\\
&\leq\inf_{h\in \ov{\sH_{\mathrm{all}}}}\Delta\sC_{\Phi,\sH_{\mathrm{all}}}(h,x) & (h\in \ov{\sH_{\mathrm{all}}}  \implies h(x)\Delta \eta(x)\leq 0)
\end{align*}
Therefore, Theorem 1.1 in \citep{bartlett2006convexity} is a special case of Corollary~\ref{cor:excess_bounds_Psi_01_B}.

\section{Proof of Theorem~\ref{Thm:excess_bounds_Psi_uniform} and Theorem~\ref{Thm:excess_bounds_Gamma_uniform}}
\label{app:uniform}
\ExcessBoundsPsiUniform*
\begin{proof}
Note the condition~\eqref{eq:condition_Psi_general} in Theorem~\ref{Thm:excess_bounds_Psi_01_general} is symmetric about $\Delta \eta(x)=0$. Thus, condition~\eqref{eq:condition_Psi_general} uniformly holds for all distributions is equivalent to the following holds for any $t\in\left[1/2,1\right]\colon$
\begin{align*}
\Psi \paren*{\tri*{2t-1}_{\e}}\leq \inf_{x\in \sX,h\in\sH:h(x)<0}\Delta\sC_{\Phi,\sH}(h,x,t),
\end{align*}
which proves the theorem.
\end{proof}

\ExcessBoundsGammaUniform*
\begin{proof}
Note the condition~\eqref{eq:condition_Gamma_general} in Theorem~\ref{Thm:excess_bounds_Gamma_01_general} is symmetric about $\Delta \eta(x)=0$. Thus, condition~\eqref{eq:condition_Gamma_general} uniformly holds for all distributions is equivalent to the following holds for any $t\in\left[1/2,1\right]\colon$
\begin{align*}
\Psi \paren*{\tri*{2t-1}_{\e}}\leq \inf_{x\in \sX,h\in\sH:h(x)<0}\Delta\sC_{\Phi,\sH}(h,x,t),
\end{align*}
which proves the theorem.
\end{proof}

\section{Proof of Theorem~\ref{Thm:excess_bounds_Psi_uniform-adv} and Theorem~\ref{Thm:excess_bounds_Gamma_uniform-adv}}
\label{app:uniform-adv}
\ExcessBoundsPsiUniformAdv*
\begin{proof}
Note the condition~\eqref{eq:condition_Psi_general_adv} in Theorem~\ref{Thm:excess_bounds_Psi_01_general_adv} is symmetric about $\Delta \eta(x)=0$. Thus, condition~\eqref{eq:condition_Psi_general_adv} uniformly holds for all distributions is equivalent to the following holds for any $t\in\left[1/2,1\right]\colon$
\begin{align*}
&\Psi\paren*{\tri*{t}_{\e}}  \leq \inf_{x\in\sX,h\in \ov \sH_\gamma\subsetneqq \sH}\Delta\sC_{\wt{\Phi},\sH}(h,x,t),\\
&\Psi\paren*{\tri*{2t-1}_{\e}} \leq \inf_{x\in \sX,h\in\sH\colon  \ov h_\gamma(x)< 0}\Delta\sC_{\wt{\Phi},\sH}(h,x,t),
\end{align*}
which proves the theorem.
\end{proof}

\ExcessBoundsGammaUniformAdv*
\begin{proof}
Note the condition~\eqref{eq:condition_Gamma_general_adv} in Theorem~\ref{Thm:excess_bounds_Gamma_01_general_adv} is symmetric about $\Delta \eta(x)=0$. Thus, condition~\eqref{eq:condition_Gamma_general_adv} uniformly holds for all distributions is equivalent to the following holds for any $t\in\left[1/2,1\right]\colon$
\begin{align*}
&\tri*{t}_{\e}  \leq \Gamma\paren*{\inf_{x\in \sX,h\in \ov \sH_\gamma\subsetneqq \sH}\Delta\sC_{\wt{\Phi},\sH}(h,x,t)},\\
&\tri*{2t-1}_{\e} \leq \Gamma\paren*{\inf_{x\in \sX,h\in\sH\colon \ov h_\gamma(x)< 0}\Delta\sC_{\wt{\Phi},\sH}(h,x,t)},
\end{align*}
which proves the theorem.
\end{proof}

\section{Proof of Theorem~\ref{Thm:tightness} and Theorem~\ref{Thm:tightness-adv} }
\label{app:tightness}
\Tightness*
\begin{proof}
By Theorem~\ref{Thm:excess_bounds_Psi_uniform}, if $\sT_{\Phi}$ is convex with $\sT_{\Phi}(0)=0$, the first inequality holds. For any $t\in [0,1]$, consider the distribution that supports on a singleton $\curl*{x_0}$ and satisfies that $\eta(x_0)=\frac12+\frac{t}{2}$. Thus
\begin{align*}
\inf_{x\in\sX,h\in\sH:h(x)<0}\Delta\sC_{\Phi,\sH}\paren*{h,x,\eta(x_0)}=\inf_{h\in\sH:h(x_0)<0}\Delta\sC_{\Phi,\sH}\paren*{h,x_0,\eta(x_0)}= \inf_{h\in\sH:h(x_0)<0}\Delta\sC_{\Phi,\sH}(h,x_0).
\end{align*}
For any $\delta>0$, take $h_0\in \sH$ such that $h_0(x_0)< 0$ and
\begin{align*}
\Delta\sC_{\Phi,\sH}(h_0,x_0)\leq\inf_{h\in\sH:h(x_0)<0}\Delta\sC_{\Phi,\sH}(h,x_0)+\delta= \inf_{x\in\sX,h\in\sH:h(x)<0}\Delta\sC_{\Phi,\sH}\paren*{h,x,\eta(x_0)}+\delta.    
\end{align*}
Then, we have
\begin{align*}
\sR_{\ell_{0-1}}(h_0)- \sR_{\ell_{0-1},\sH}^*+\sM_{\ell_{0-1},\sH} & = \sR_{\ell_{0-1}}(h_0) - \mathbb{E}_{X} \bracket* {\sC^*_{\ell_{0-1},\sH}(x)}\\
& =\Delta\sC_{\ell_{0-1},\sH}(h_0,x_0)\\ 
& =2\eta(x_0)-1\\
& = t,\\
\sR_{\Phi}(h_0)-\sR_{\Phi,\sH}^*+\sM_{\Phi,\sH} 
& = \sR_{\Phi}(h_0) - \mathbb{E}_{X} \bracket* {\sC^*_{\Phi,\sH}(x)}\\
& = \Delta\sC_{\Phi,\sH}(h_0,x_0)\\
& \leq \inf_{x\in\sX,h\in\sH:h(x)<0}\Delta\sC_{\Phi,\sH}\paren*{h,x,\eta(x_0)} + \delta \\
& = \sT_{\Phi}\paren*{2\eta(x_0)-1}+\delta\\
& = \sT_{\Phi}(t) + \delta,
\end{align*}
which completes the proof.
\quad
\end{proof}
\TightnessAdv*
\begin{proof}
By Theorem~\ref{Thm:excess_bounds_Psi_uniform-adv}, if $\sT_{\wt{\Phi}}$ is convex with $\sT_{\wt{\Phi}}(0)=0$, the first inequality holds. For any $t\in [0,1]$, consider the distribution that supports on a singleton $\curl*{x_0}$, which satisfies that $\eta(x_0)=\frac12+\frac{t}{2}$ and $\ov \sH_{\gamma}\neq \sH$.
Thus
\begin{align*}
\inf_{x\in\sX,h\in\sH:\ov h_\gamma(x)<0}\Delta\sC_{\wt{\Phi},\sH}\paren*{h,x,\eta(x_0)}=\inf_{h\in\sH:\ov h_\gamma(x_0)<0}\Delta\sC_{\wt{\Phi},\sH}\paren*{h,x_0,\eta(x_0)}= \inf_{h\in\sH:\ov h_\gamma(x_0)<0}\Delta\sC_{\wt{\Phi},\sH}(h,x_0).
\end{align*}
For any $\delta>0$, take $h\in \sH$ such that $\ov h_\gamma(x_0)< 0$ and
\begin{align*}
\Delta\sC_{\wt{\Phi},\sH}(h,x_0)\leq \inf_{h\in\sH:\ov h_\gamma(x_0)<0}\Delta\sC_{\wt{\Phi},\sH}(h,x_0)+\delta=\inf_{x\in\sX,h\in\sH:\ov h_\gamma(x)<0}\Delta\sC_{\wt{\Phi},\sH}\paren*{h,x,\eta(x_0)}+\delta.    
\end{align*}
Then, we have
\begin{align*}
\sR_{\ell_{\gamma}}(h)- \sR_{\ell_{\gamma},\sH}^*+\sM_{\ell_{\gamma},\sH} & = \sR_{\ell_{\gamma}}(h) - \mathbb{E}_{X} \bracket* {\sC^*_{\ell_{\gamma},\sH}(x)}\\
& =\Delta\sC_{\ell_{\gamma},\sH}(h,x_0)\\
& =2\eta(x_0)-1\\
& = t,\\
\sR_{\wt{\Phi}}(h)-\sR_{\wt{\Phi},\sH}^*+\sM_{\wt{\Phi},\sH} 
& = \sR_{\wt{\Phi}}(h) - \mathbb{E}_{X} \bracket* {\sC^*_{\wt{\Phi},\sH}(x)}\\
& = \Delta\sC_{\wt{\Phi},\sH}(h,x_0)\\
& \leq \inf_{x\in\sX,h\in\sH:\ov h_\gamma(x)< 0}\Delta\sC_{\wt{\Phi},\sH}\paren*{h,x,\eta(x_0)} + \delta \\
& = \sT_{2}\paren*{2\eta(x_0)-1}+\delta\\
& = \sT_{\wt{\Phi}}(2\eta(x_0)-1) + \delta & (\sT_2\leq \sT_1)\\
& =\sT_{\wt{\Phi}}(t) + \delta
\end{align*}
which completes the proof.
\quad
\end{proof}


\section{Proof of Theorem~\ref{Thm:negative_convex_adv}}
\label{app:negative_adv}
\begin{definition}[\textbf{Regularity for adversarial calibration}]\emph{[Definition~5 in \citep{awasthi2021calibration}]}
\label{def:regularity}
We say that a hypothesis set $\sH$ is \emph{regular for adversarial calibration}
if there exists a \emph{distinguishing $x$} in $\sX$, that is if there exist $f, g \in \sH$ 
such that $\inf_{\| x' - x \|_p \leq \gamma } f(x')>0$ and 
$\sup_{\| x' - x \|_p \leq \gamma }g(x')<0$.
\end{definition}
\NegativeConvexAdv*
\begin{proof}
Assume $x_0\in \sX$ is distinguishing. Consider the distribution that supports on $\curl*{x_0}$. Let $\eta(x_0)=1/2$ and $h_0=0\in \sH$. Then, for any $h\in \sH$,
\begin{align*}
\sR_{\ell_{\gamma}}(h)=\sC_{\ell_{\gamma}}(h,x_0)
=1/2\mathds{1}_{\uv h_\gamma(x_0)\leq 0} +1/2 \mathds{1}_{\ov h_\gamma(x_0)\geq 0} \geq 1/2,
\end{align*}
where the equality can be achieved for some $h\in \sH$ since $x_0$ is distinguishing. Therefore, 
\begin{align*}
\sR_{\ell_{\gamma},\sH}^*=\sC^*_{\ell_{\gamma},\sH}(x_0)=\inf_{h\in \sH}\sC_{\ell_{\gamma}}(h,x_0)=1/2.
\end{align*}
Note $\sR_{\ell_{\gamma}}(h_0)=1/2 +1/2 =1$. 
For the supremum-based convex loss $\wt{\Phi}$, for any $h\in \sH$,
\begin{align*}
\sR_{\wt{\Phi}}(h)=\sC_{\wt{\Phi}}(h,x_0)
&=1/2\Phi\paren*{\uv h_\gamma(x_0)} +1/2\Phi\paren*{-\ov h_\gamma(x_0)}\\  
& \geq \Phi\paren*{1/2 \uv h_\gamma(x_0) -1/2 \ov h_\gamma(x_0)} &\quad \paren*{\text{convexity of }\Phi}\\
& \geq \Phi(0),  &\quad \paren*{\Phi \text{ is non-increasing}}
\end{align*}
where both equality can be achieved by $h_0=0$.
Therefore,
\begin{align*}
\sR_{\wt{\Phi},\sH}^*=\sC^*_{\wt{\Phi},\sH}(x_0)=\sR_{\wt{\Phi}}(h_0)=\Phi(0).
\end{align*}
If \eqref{eq:est-bound} holds for some non-decreasing function $f$, then, we obtain for any $h\in \sH$,
\begin{align*}
\sR_{\ell_{\gamma}}(h)-1/2\leq  f\paren*{\sR_{\wt{\Phi}}(h) - \Phi(0)}.
\end{align*}
Let $h=h_0$, then $f(0)\geq 1/2$. Since $f$ is non-decreasing, for any $t\in [0,1]$, $f(t)\geq 1/2$.

For the supremum-based sigmoid loss $\wt{\Phi}_{\mathrm{sig}}$, for any $h\in \sH$,
\begin{align*}
\sR_{\wt{\Phi}_{\mathrm{sig}}}(h)=\sC_{\wt{\Phi}_{\mathrm{sig}}}(h,x_0)
&=1/2\Phi_{\mathrm{sig}}\paren*{\uv h_\gamma(x_0)} +1/2\Phi_{\mathrm{sig}}\paren*{-\ov h_\gamma(x_0)}\\
& = 1 +1/2 \bracket*{\tanh(k \ov h_\gamma(x_0))-\tanh(k \uv h_\gamma(x_0))} \\
& \geq 1
\end{align*}
where the equality can be achieved by $h_0=0$.
Therefore,
\begin{align*}
\sR_{\wt{\Phi}_{\mathrm{sig}},\sH}^*=\sC^*_{\wt{\Phi}_{\mathrm{sig}},\sH}(x_0)=\sR_{\wt{\Phi}_{\mathrm{sig}}}(h_0)=1
\end{align*}
If \eqref{eq:est-bound} holds for some non-decreasing function $f$, then, we obtain for any $h\in \sH$,
\begin{align*}
\sR_{\ell_{\gamma}}(h)-1/2\leq  f\paren*{\sR_{\wt{\Phi}_{\mathrm{sig}}}(h) - 1}.
\end{align*}
Let $h=h_0$, then $f(0)\geq 1/2$. Since $f$ is non-decreasing, for any $t\in [0,1]$, $f(t)\geq 1/2$.
\end{proof}

\section{Derivation of Non-Adversarial \texorpdfstring{$\sH$}{H}-Estimation Error Bounds}
\label{app:derivation-non-adv}
\subsection{Linear Hypotheses}
\label{app:derivation-lin}
Since $\sH_{\mathrm{lin}}$ satisfies the condition of Lemma~\ref{lemma:explicit_assumption_01}, by Lemma~\ref{lemma:explicit_assumption_01} the $\paren*{\ell_{0-1},\sH_{\mathrm{lin}}}$-minimizability gap can be expressed as follows:
\begin{equation}
\begin{aligned}
\label{eq:M-01-lin}
\sM_{\ell_{0-1},\sH_{\mathrm{lin}}}
& = \sR_{\ell_{0-1},\sH_{\mathrm{lin}}}^*-\mathbb{E}_{X}\bracket*{\min\curl*{\eta(x),1-\eta(x)}} \\
&= \sR_{\ell_{0-1}, \sH_{\mathrm{lin}}}^* - \sR_{\ell_{0-1}, \sH_{\mathrm{all}}}^*.
\end{aligned}
\end{equation}
Therefore, the $\paren*{\ell_{0-1},\sH_{\mathrm{lin}}}$-minimizability gap coincides with
the $\paren*{\ell_{0-1},\sH_{\mathrm{lin}}}$-approximation error.
By the definition of $\sH_{\mathrm{lin}}$, for any $x \in \sX$, $\curl[\big]{h(x) \mid h \in \sH_{\mathrm{lin}}} = \bracket*{-W \norm*{x}_p-B, W\norm*{x}_p + B}$.
\subsubsection{Hinge Loss}
For the hinge loss $\Phi_{\mathrm{hinge}}(\alpha)\colon=\max\curl*{0,1 - \alpha}$, for all $h\in \sH_{\mathrm{lin}}$ and $x\in \sX$:
\begin{equation*}
\begin{aligned}
\sC_{\Phi_{\mathrm{hinge}}}(h,x,t)
& =t \Phi_{\mathrm{hinge}}(h(x))+(1-t)\Phi_{\mathrm{hinge}}(-h(x))\\
& =t\max\curl*{0,1-h(x)}+(1-t)\max\curl*{0,1+h(x)}.\\
\inf_{h\in\sH_{\mathrm{lin}}}\sC_{\Phi_{\mathrm{hinge}}}(h,x,t)
& = 1-\abs*{2t-1}\min\curl*{W\norm*{x}_p+B,1}.
\end{aligned}
\end{equation*}
Therefore, the $\paren*{\Phi_{\mathrm{hinge}},\sH_{\mathrm{lin}}}$-minimizability gap can be expressed as follows:
\begin{equation}
\begin{aligned}
\label{eq:M-hinge-lin}
\sM_{\Phi_{\mathrm{hinge}},\sH_{\mathrm{lin}}}
& = \sR_{\Phi_{\mathrm{hinge}},\sH_{\mathrm{lin}}}^*-\mathbb{E}_{X}\bracket*{1-\inf_{h\in\sH_{\mathrm{lin}}}\sC_{\Phi_{\mathrm{hinge}}}(h,x,\eta(x))}.\\
& = \sR_{\Phi_{\mathrm{hinge}},\sH_{\mathrm{lin}}}^*-\mathbb{E}_{X}\bracket*{1-\abs*{2\eta(x)-1}\min\curl*{W\norm*{x}_p+B,1}}.
\end{aligned}
\end{equation}
Note the $\paren*{\Phi_{\mathrm{hinge}},\sH_{\mathrm{lin}}}$-minimizability gap coincides with
the $\paren*{\Phi_{\mathrm{hinge}},\sH_{\mathrm{lin}}}$-approximation error $\sR_{\Phi_{\mathrm{hinge}},\sH_{\mathrm{lin}}}^*-\mathbb{E}_{X}\bracket*{1-\abs*{2\eta(x)-1}}$ for $B \geq 1$.

For $\frac{1}2< t\leq1$, we have
\begin{align*}
\inf_{h\in\sH_{\mathrm{lin}}:h(x)<0}\sC_{\Phi_{\mathrm{hinge}}}(h,x,t)
& = t\max\curl*{0,1-0}+(1-t)\max\curl*{0,1+0}\\
& =1.\\
\inf_{x\in \sX} \inf_{h\in\sH_{\mathrm{lin}}:h(x)<0} \Delta\sC_{\Phi_{\mathrm{hinge}},\sH_{\mathrm{lin}}}(h,x,t)
& = \inf_{x\in \sX} \curl*{\inf_{h\in\sH_{\mathrm{lin}}:h(x)<0}\sC_{\Phi_{\mathrm{hinge}}}(h,x,t)-\inf_{h\in\sH_{\mathrm{lin}}}\sC_{\Phi_{\mathrm{hinge}}}(h,x,t)}\\
&=\inf_{x\in \sX}\paren*{2t-1}\min\curl*{W\norm*{x}_p+B,1}\\
&=(2t-1)\min\curl*{B,1}\\
&=\sT(2t - 1),
\end{align*}
where $\sT$ is the increasing and convex function on $[0,1]$ defined by
\begin{align*}
\forall t \in [0,1], \quad \sT(t) = \min \curl*{B, 1} \, t .
\end{align*}
By Definition~\ref{def:trans}, for any $\epsilon\geq 0$, the $\sH_{\mathrm{lin}}$-estimation error transformation of the hinge loss is as follows:
\begin{align*}
\sT_{\Phi_{\mathrm{hinge}}}= \min \curl*{B, 1} \, t, \quad t \in [0,1],
\end{align*}
Therefore, $\sT_{\Phi_{\mathrm{hinge}}}$ is convex, non-decreasing, invertible and satisfies that $\sT_{\Phi_{\mathrm{hinge}}}(0)=0$. By Theorem~\ref{Thm:tightness}, we can choose $\Psi(t)=\min\curl*{B,1} \, t$ in Theorem~\ref{Thm:excess_bounds_Psi_uniform}, or, equivalently, $\Gamma(t) = \frac{t}{\min\curl*{B, 1}}$ in Theorem~\ref{Thm:excess_bounds_Gamma_uniform}, which are optimal when $\e=0$. Thus, by Theorem~\ref{Thm:excess_bounds_Psi_uniform} or Theorem~\ref{Thm:excess_bounds_Gamma_uniform}, setting $\e = 0$ yields the $\sH_{\mathrm{lin}}$-consistency estimation error bound for the hinge loss, valid for all $h \in \sH_{\mathrm{lin}}$:
\begin{align}
\label{eq:hinge-lin-est}
     \sR_{\ell_{0-1}}(h)- \sR_{\ell_{0-1},\sH_{\mathrm{lin}}}^*
     \leq \frac{\sR_{\Phi_{\mathrm{hinge}}}(h)- \sR_{\Phi_{\mathrm{hinge}},\sH_{\mathrm{lin}}}^*+\sM_{\Phi_{\mathrm{hinge}},\sH_{\mathrm{lin}}}}{\min\curl*{B,1}}-\sM_{\ell_{0-1}, \sH_{\mathrm{lin}}}.
\end{align}
Since the $\paren*{\ell_{0-1},\sH_{\mathrm{lin}}}$-minimizability gap coincides with
the $\paren*{\ell_{0-1},\sH_{\mathrm{lin}}}$-approximation error and 
$\paren*{\Phi_{\mathrm{hinge}},\sH_{\mathrm{lin}}}$-minimizability gap coincides with
the $\paren*{\Phi_{\mathrm{hinge}},\sH_{\mathrm{lin}}}$-approximation error for $B \geq 1$,
the inequality can be rewritten as follows:
\begin{align*}
     \sR_{\ell_{0-1}}(h)- \sR_{\ell_{0-1},\sH_{\mathrm{all}}}^*
     \leq 
     \begin{cases}
     \sR_{\Phi_{\mathrm{hinge}}}(h) - \sR_{\Phi_{\mathrm{hinge}},\sH_{\mathrm{all}}}^* & \text{if } B \geq 1\\
     \frac{1}{B} \bracket[\Big]{\sR_{\Phi_{\mathrm{hinge}}}(h)
     - \mathbb{E}_{X}\bracket*{1-\abs*{2\eta(x)-1}\min\curl*{W\norm*{x}_p+B,1}} } & \text{otherwise}.
     \end{cases}
\end{align*}
The inequality for $B \geq 1$ coincides with the consistency excess error bound
known for the hinge loss \citep{Zhang2003,bartlett2006convexity,MohriRostamizadehTalwalkar2018} but the one for $B < 1$ is distinct and novel. For $B<1$, we have
\begin{align*}
\mathbb{E}_{X}\bracket*{1-\abs*{2\eta(x)-1}\min\curl*{W\norm*{x}_p+B,1}}> \mathbb{E}_{X}\bracket*{1-\abs*{2\eta(x)-1}}= 2\mathbb{E}_X\bracket*{\min\curl*{\eta(x), 1 - \eta(x)}}= \sR_{\Phi_{\mathrm{hinge}},\sH_{\mathrm{all}}}^*.
\end{align*}
Therefore for $B<1$,
\begin{align*}
\sR_{\Phi_{\mathrm{hinge}}}(h) - \mathbb{E}_{X}\bracket*{1-\abs*{2\eta(x)-1}\min\curl*{W\norm*{x}_p+B,1}} < \sR_{\Phi_{\mathrm{hinge}}}(h) - \sR_{\Phi_{\mathrm{hinge}},\sH_{\mathrm{all}}}^*.
\end{align*}
Note that: $\sR_{\Phi_{\mathrm{hinge}},\sH_{\mathrm{all}}}^* = 2 \sR_{\ell_{0-1},\sH_{\mathrm{all}}}^* =2\mathbb{E}_X\bracket*{\min\curl*{\eta(x), 1 - \eta(x)}}$. Thus, the first
inequality (case $B \geq 1$) can be equivalently written as follows:
\begin{align*}
    \forall h \in \sH_{\mathrm{lin}},\; \sR_{\ell_{0-1}}(h) 
     \leq \sR_{\Phi_{\mathrm{hinge}}}(h) - \mathbb{E}_X\bracket*{\min\curl*{\eta(x), 1 - \eta(x)}},
\end{align*}
which is a more informative upper bound than the standard
inequality $\sR_{\ell_{0-1}}(h) 
     \leq \sR_{\Phi_{\mathrm{hinge}}}(h)$.
     
\subsubsection{Logistic Loss}
For the logistic loss $\Phi_{\mathrm{log}}(\alpha)\colon=\log_2(1+e^{-\alpha})$, for all $h\in \sH_{\mathrm{lin}}$ and $x\in \sX$:
\begin{equation*}
\begin{aligned}
\sC_{\Phi_{\mathrm{log}}}(h,x,t)
& = t \Phi_{\mathrm{log}}(h(x))+(1-t)\Phi_{\mathrm{log}}(-h(x)),\\
& = t\log_2\paren*{1+e^{-h(x)}}+(1-t)\log_2\paren*{1+e^{h(x)}}.\\
\inf_{h\in\sH_{\mathrm{lin}}}\sC_{\Phi_{\mathrm{log}}}(h,x,t)
&=\begin{cases}
-t\log_2(t)-(1-t)\log_2(1-t) & \text{if }\log\abs*{\frac{t}{1-t}}\leq W\norm*{x}_p+B,\\
\max\curl*{t,1-t}\log_2\paren*{1+e^{-(W\norm*{x}_p+B)}}+\min\curl*{t,1-t}\log_2\paren*{1+e^{W\norm*{x}_p+B}} &\text{if }\log\abs*{\frac{t}{1-t}}> W\norm*{x}_p+B.
\end{cases}
\end{aligned}
\end{equation*}
Therefore, the $\paren*{\Phi_{\mathrm{log}},\sH_{\mathrm{lin}}}$-minimizability gap can be expressed as follows:
\begin{equation}
\begin{aligned}
\label{eq:M-log-lin}
\sM_{\Phi_{\mathrm{log}},\sH_{\mathrm{lin}}}
& = \sR_{\Phi_{\mathrm{log}},\sH_{\mathrm{lin}}}^*-
\mathbb{E}_{X}\bracket*{\inf_{h\in\sH_{\mathrm{lin}}}\sC_{\Phi_{\mathrm{log}}}(h,x,\eta(x))}\\
& = \sR_{\Phi_{\mathrm{log}},\sH_{\mathrm{lin}}}^*-
\mathbb{E}_{X}\bracket*{-\eta(x)\log_2(\eta(x))-(1-\eta(x))\log_2(1-\eta(x))\mathds{1}_{\log\abs*{\frac{\eta(x)}{1-\eta(x)}}\leq W\norm*{x}_p+B}}\\
& - \mathbb{E}_{X}\bracket*{\max\curl*{\eta(x),1-\eta(x)}\log_2\paren*{1+e^{-(W\norm*{x}_p+B)}}\mathds{1}_{\log\abs*{\frac{\eta(x)}{1-\eta(x)}}> W\norm*{x}_p+B}}\\
&-\mathbb{E}_{X}\bracket*{\min\curl*{\eta(x),1-\eta(x)}\log_2\paren*{1+e^{W\norm*{x}_p+B}}\mathds{1}_{\log\abs*{\frac{\eta(x)}{1-\eta(x)}}> W\norm*{x}_p+B}}
\end{aligned}
\end{equation}
Note $\paren*{\Phi_{\mathrm{log}},\sH_{\mathrm{lin}}}$-minimizability gap coincides with
the $\paren*{\Phi_{\mathrm{log}},\sH_{\mathrm{lin}}}$-approximation error $\sR_{\Phi_{\mathrm{log}},\sH_{\mathrm{lin}}}^*-
\mathbb{E}_{X}\bracket*{-\eta(x)\log_2(\eta(x))-(1-\eta(x))\log_2(1-\eta(x))}$ for $B =\plus\infty$.

For $\frac{1}2< t\leq1$, we have
\begin{align*}
\inf_{h\in\sH_{\mathrm{lin}}:h(x)<0}\sC_{\Phi_{\mathrm{log}}}(h,x,t)
& = t\log_2\paren*{1+e^{-0}}+(1-t)\log_2\paren*{1+e^{0}} \\
& = 1, \\
\inf_{x\in \sX}\inf_{h\in\sH_{\mathrm{lin}}:h(x)<0}\Delta\sC_{\Phi_{\mathrm{log}},\sH_{\mathrm{lin}}}(h,x,t)
&=\inf_{x\in \sX}\paren*{\inf_{h\in\sH_{\mathrm{lin}}:h(x)<0}\sC_{\Phi_{\mathrm{log}}}(h,x,t)-\inf_{h\in\sH_{\mathrm{lin}}}\sC_{\Phi_{\mathrm{log}}}(h,x,t)}\\
&=\inf_{x\in \sX}\begin{cases}
1+t\log_2(t)+(1-t)\log_2(1-t)\\
\text{if }\log\frac{t}{1-t}\leq W\norm*{x}_p+B,\\
1-t\log_2\paren*{1+e^{-(W\norm*{x}_p+B)}}-(1-t)\log_2\paren*{1+e^{W\norm*{x}_p+B}}\\
\text{if }\log\frac{t}{1-t}> W\norm*{x}_p+B.
\end{cases}\\
&=\begin{cases}
1+t\log_2(t)+(1-t)\log_2(1-t) & \text{if }\log\frac{t}{1-t}\leq B,\\
1-t\log_2\paren*{1+e^{-B}}-(1-t)\log_2\paren*{1+e^{B}} & \text{if }\log\frac{t}{1-t}> B.
\end{cases}\\
&=\sT(2t-1),
\end{align*}
where $\sT$ is the increasing and convex function on $[0,1]$ defined by
\begin{align*}
\forall t\in[0,1], \quad
\sT(t)=\begin{cases}
\frac{t+1}{2}\log_2(t+1)+\frac{1-t}{2}\log_2(1-t),\quad &  t\leq \frac{e^B-1}{e^B+1},\\
1-\frac{t+1}{2}\log_2(1+e^{-B})-\frac{1-t}{2}\log_2(1+e^B),\quad & t> \frac{e^B-1}{e^B+1}.
\end{cases}
\end{align*}
By Definition~\ref{def:trans}, for any $\epsilon\geq 0$, the $\sH_{\mathrm{lin}}$-estimation error transformation of the logistic loss is as follows:
\begin{align*}
\sT_{\Phi_{\mathrm{log}}}= 
\begin{cases}
\sT(t), & t\in \left[\epsilon,1\right], \\
\frac{\sT(\epsilon)}{\epsilon}\, t, &  t\in \left[0,\epsilon\right).
\end{cases}
\end{align*}
Therefore, when $\epsilon=0$, $\sT_{\Phi_{\mathrm{log}}}$ is convex, non-decreasing, invertible and satisfies that $\sT_{\Phi_{\mathrm{log}}}(0)=0$. By Theorem~\ref{Thm:tightness}, we can choose $\Psi(t)=\sT_{\Phi_{\mathrm{log}}}(t)$ in Theorem~\ref{Thm:excess_bounds_Psi_uniform}, or equivalently $\Gamma(t)=\sT_{\Phi_{\mathrm{log}}}^{-1}(t)$ in Theorem~\ref{Thm:excess_bounds_Gamma_uniform}, which are optimal. To simplify the expression, using the fact that
\begin{align*}
\frac{t+1}{2}\log_2(t+1)+\frac{1-t}{2}\log_2(1-t)
&=
1-\paren*{-\frac{t+1}{2}\log_2\paren*{\frac{t+1}{2}}-\frac{1-t}{2}\log_2\paren*{\frac{1-t}{2}}}\\
&\geq 1 - \sqrt{4\frac{1-t}{2} \frac{t+1}{2}}\\
& = 1- \sqrt{1-t^2}\\
& \geq \frac{t^2}{2},\\
1-\frac{t+1}{2}\log_2(1+e^{-B})-\frac{1-t}{2}\log_2(1+e^B)
&=\frac{1}{2}\log_2\paren*{\frac{4}{2+e^{-B}+e^B}}+1/2\log_2\paren*{\frac{1+e^B}{1+e^{-B}}}\, t,
\end{align*}
$\sT_{\Phi_{\mathrm{log}}}$ can be lower bounded by
\begin{align*}
\wt{\sT}_{\Phi_{\mathrm{log}}}(t)= \begin{cases}
\frac{t^2}{2},& t\leq \frac{e^B-1}{e^B+1},\\
\frac{1}{2}\paren*{\frac{e^B-1}{e^B+1}}\, t, & t> \frac{e^B-1}{e^B+1}.
\end{cases}   
\end{align*}
Thus, we adopt an upper bound of $\sT_{\Phi_{\mathrm{log}}}^{-1}$ as follows:
\begin{align*}
\wt{\sT}_{\Phi_{\mathrm{log}}}^{-1}(t)=\begin{cases}
\sqrt{2t}, & t\leq \frac{1}{2}\paren*{\frac{e^B-1}{e^B+1}}^2,\\
2\paren*{\frac{e^B+1}{e^B-1}}\, t, & t> \frac{1}{2}\paren*{\frac{e^B-1}{e^B+1}}^2.
\end{cases}
\end{align*}
Therefore, by Theorem~\ref{Thm:excess_bounds_Psi_uniform} or Theorem~\ref{Thm:excess_bounds_Gamma_uniform}, setting $\e = 0$ yields the $\sH_{\mathrm{lin}}$-consistency estimation error bound for the logistic loss, valid for all $h \in \sH_{\mathrm{lin}}$:
\begin{multline}
\label{eq:log-lin-est}
     \sR_{\ell_{0-1}}(h)-\sR_{\ell_{0-1},\sH_{\mathrm{lin}}}^*+\sM_{\ell_{0-1}, \sH_{\mathrm{lin}}}\\
     \leq 
     \begin{cases}
     \sqrt{2}\,\paren*{\sR_{\Phi_{\mathrm{log}}}(h)- \sR_{\Phi_{\mathrm{log}},\sH_{\mathrm{lin}}}^*+\sM_{\Phi_{\mathrm{log}},\sH_{\mathrm{lin}}}}^{\frac12}, & \text{if } \sR_{\Phi_{\mathrm{log}}}(h)- \sR_{\Phi_{\mathrm{log}},\sH_{\mathrm{lin}}}^*\leq \frac{1}{2}\paren*{\frac{e^B-1}{e^B+1}}^2-\sM_{\Phi_{\mathrm{log}},\sH_{\mathrm{lin}}}\\
     2\paren*{\frac{e^B+1}{e^B-1}}\paren*{\sR_{\Phi_{\mathrm{log}}}(h)- \sR_{\Phi_{\mathrm{log}},\sH_{\mathrm{lin}}}^*+\sM_{\Phi_{\mathrm{log}},\sH_{\mathrm{lin}}}}, & \text{otherwise}
     \end{cases}
\end{multline}
Since the $\paren*{\ell_{0-1},\sH_{\mathrm{lin}}}$-minimizability gap coincides with
the $\paren*{\ell_{0-1},\sH_{\mathrm{lin}}}$-approximation error and 
$\paren*{\Phi_{\mathrm{log}},\sH_{\mathrm{lin}}}$-minimizability gap coincides with
the $\paren*{\Phi_{\mathrm{log}},\sH_{\mathrm{lin}}}$-approximation error for $B =\plus\infty$,
the inequality can be rewritten as follows:
\begin{align*}
     &\sR_{\ell_{0-1}}(h)- \sR_{\ell_{0-1},\sH_{\mathrm{all}}}^*\\
     &\quad \leq 
     \begin{cases}
      \sqrt{2}\,\bracket*{\sR_{\Phi_{\mathrm{log}}}(h) - \sR_{\Phi_{\mathrm{log}},\sH_{\mathrm{all}}}^*}^{\frac12} & \text{if } B = \plus\infty \\
     \begin{cases}
    \sqrt{2}\,\bracket*{\sR_{\Phi_{\mathrm{log}}}(h)- \sR_{\Phi_{\mathrm{log}},\sH_{\mathrm{lin}}}^*+\sM_{\Phi_{\mathrm{log}},\sH_{\mathrm{lin}}}}^{\frac12}  & \text{if } \sR_{\Phi_{\mathrm{log}}}(h)- \sR_{\Phi_{\mathrm{log}},\sH_{\mathrm{lin}}}^*\leq \frac{1}{2}\paren*{\frac{e^B-1}{e^B+1}}^2-\sM_{\Phi_{\mathrm{log}},\sH_{\mathrm{lin}}} \\
    2\paren*{\frac{e^B+1}{e^B-1}}\paren*{\sR_{\Phi_{\mathrm{log}}}(h)- \sR_{\Phi_{\mathrm{log}},\sH_{\mathrm{lin}}}^*+\sM_{\Phi_{\mathrm{log}},\sH_{\mathrm{lin}}}} & \text{otherwise}
    \end{cases} & \text{otherwise}
     \end{cases}
\end{align*}
where the $\paren*{\Phi_{\mathrm{log}},\sH_{\mathrm{lin}}}$-minimizability gap $\sM_{\Phi_{\mathrm{log}},\sH_{\mathrm{lin}}}$ is characterized as below, which is less than
the $\paren*{\Phi_{\mathrm{log}},\sH_{\mathrm{lin}}}$-approximation error when $B<\plus \infty$:
\begin{align*}
\sM_{\Phi_{\mathrm{log}},\sH_{\mathrm{lin}}}
& = \sR_{\Phi_{\mathrm{log}},\sH_{\mathrm{lin}}}^*-
\mathbb{E}_{X}\bracket*{-\eta(x)\log_2(\eta(x))-(1-\eta(x))\log_2(1-\eta(x))\mathds{1}_{\log\abs*{\frac{\eta(x)}{1-\eta(x)}}\leq W\norm*{x}_p+B}}\\
& - \mathbb{E}_{X}\bracket*{\max\curl*{\eta(x),1-\eta(x)}\log_2\paren*{1+e^{-(W\norm*{x}_p+B)}}\mathds{1}_{\log\abs*{\frac{\eta(x)}{1-\eta(x)}}> W\norm*{x}_p+B}}\\
&-\mathbb{E}_{X}\bracket*{\min\curl*{\eta(x),1-\eta(x)}\log_2\paren*{1+e^{W\norm*{x}_p+B}}\mathds{1}_{\log\abs*{\frac{\eta(x)}{1-\eta(x)}}> W\norm*{x}_p+B}}\\
& < \sR_{\Phi_{\mathrm{log}},\sH_{\mathrm{lin}}}^* - \mathbb{E}_{X}\bracket*{-\eta(x)\log_2(\eta(x))-(1-\eta(x))\log_2(1-\eta(x))}\\
& = \sR_{\Phi_{\mathrm{log}},\sH_{\mathrm{lin}}}^* - \sR_{\Phi_{\mathrm{log}},\sH_{\mathrm{all}}}^* .
\end{align*}
Therefore, the inequality for $B = \plus \infty$ coincides with the consistency excess error bound
known for the logistic loss \citep{Zhang2003,MohriRostamizadehTalwalkar2018} but the one for $B< \plus \infty$ is distinct and novel.

\subsubsection{Exponential Loss}
For the exponential loss $\Phi_{\mathrm{exp}}(\alpha)\colon=e^{-\alpha}$, for all $h\in \sH_{\mathrm{lin}}$ and $x\in \sX$:
\begin{equation*}
\begin{aligned}
\sC_{\Phi_{\mathrm{exp}}}(h,x,t)
&=t \Phi_{\mathrm{exp}}(h(x))+(1-t)\Phi_{\mathrm{exp}}(-h(x))\\
&=te^{-h(x)}+(1-t)e^{h(x)}.\\
\inf_{h\in\sH_{\mathrm{lin}}}\sC_{\Phi_{\mathrm{exp}}}(h,x,t)
&=\begin{cases}
2\sqrt{t(1-t)} & \text{if }1/2\log\abs*{\frac{t}{1-t}}\leq W\norm*{x}_p+B\\
\max\curl*{t,1-t}e^{-(W\norm*{x}_p+B)}+\min\curl*{t,1-t}e^{W\norm*{x}_p+B} & \text{if }1/2\log\abs*{\frac{t}{1-t}}> W\norm*{x}_p+B.
\end{cases}
\end{aligned}
\end{equation*}
Therefore, the $\paren*{\Phi_{\mathrm{exp}},\sH_{\mathrm{lin}}}$-minimizability gap can be expressed as follows:
\begin{equation}
\begin{aligned}
\label{eq:M-exp-lin}
\sM_{\Phi_{\mathrm{exp}},\sH_{\mathrm{lin}}}
& = \sR_{\Phi_{\mathrm{exp}},\sH_{\mathrm{lin}}}^*-
\mathbb{E}_{X}\bracket*{\inf_{h\in\sH_{\mathrm{lin}}}\sC_{\Phi_{\mathrm{exp}}}(h,x,\eta(x))}\\
& = \sR_{\Phi_{\mathrm{exp}},\sH_{\mathrm{lin}}}^*-
\mathbb{E}_{X}\bracket*{2\sqrt{\eta(x)(1-\eta(x))}\mathds{1}_{1/2\log\abs*{\frac{\eta(x)}{1-\eta(x)}}\leq W\norm*{x}_p+B}}\\
& - \mathbb{E}_{X}\bracket*{\max\curl*{\eta(x),1-\eta(x)}e^{-(W\norm*{x}_p+B)}\mathds{1}_{1/2\log\abs*{\frac{\eta(x)}{1-\eta(x)}}> W\norm*{x}_p+B}}\\
&-\mathbb{E}_{X}\bracket*{\min\curl*{\eta(x),1-\eta(x)}e^{W\norm*{x}_p+B}\mathds{1}_{1/2\log\abs*{\frac{\eta(x)}{1-\eta(x)}}> W\norm*{x}_p+B}}.
\end{aligned}
\end{equation}
Note $\paren*{\Phi_{\mathrm{exp}},\sH_{\mathrm{lin}}}$-minimizability gap coincides with
the $\paren*{\Phi_{\mathrm{exp}},\sH_{\mathrm{lin}}}$-approximation error $\sR_{\Phi_{\mathrm{exp}},\sH_{\mathrm{lin}}}^*-
\mathbb{E}_{X}\bracket*{2\sqrt{\eta(x)(1-\eta(x))}}$ for $B =\plus\infty$.

For $\frac{1}2< t\leq1$, we have
\begin{align*}
\inf_{h\in\sH_{\mathrm{lin}}:h(x)<0}\sC_{\Phi_{\mathrm{exp}}}(h,x,t)
&=te^{-0}+(1-t)e^{0}\\
&=1.\\
\inf_{x\in \sX}\inf_{h\in\sH_{\mathrm{lin}}:h(x)<0}\Delta\sC_{\Phi_{\mathrm{exp}},\sH_{\mathrm{lin}}}(h,x,t)
&=\inf_{x\in \sX}\paren*{\inf_{h\in\sH_{\mathrm{lin}}:h(x)<0}\sC_{\Phi_{\mathrm{exp}}}(h,x,t)-\inf_{h\in\sH_{\mathrm{lin}}}\sC_{\Phi_{\mathrm{exp}}}(h,x,t)}\\
&=\inf_{x\in \sX}\begin{cases}
1-2\sqrt{t(1-t)} & \text{if }1/2\log\frac{t}{1-t}\leq W\norm*{x}_p+B,\\
1-te^{-(W\norm*{x}_p+B)}-(1-t)e^{W\norm*{x}_p+B} & \text{if }1/2\log\frac{t}{1-t}> W\norm*{x}_p+B.
\end{cases}\\
&=\begin{cases}
1-2\sqrt{t(1-t)}, & 1/2\log\frac{t}{1-t}\leq B\\
1-te^{-B}-(1-t)e^{B}, & 1/2\log\frac{t}{1-t}> B
\end{cases}\\
&=\sT(2t-1),
\end{align*}
where $\sT$ is the increasing and convex function on $[0,1]$ defined by
\begin{align*}
\forall t\in[0,1], \quad 
\sT(t)=\begin{cases}
1-\sqrt{1-t^2}, & t\leq \frac{e^{2B}-1}{e^{2B}+1},\\
1-\frac{t+1}{2}e^{-B}-\frac{1-t}{2}e^B, & t> \frac{e^{2B}-1}{e^{2B}+1}.
\end{cases}
\end{align*}
By Definition~\ref{def:trans}, for any $\epsilon\geq 0$, the $\sH_{\mathrm{lin}}$-estimation error transformation of the exponential loss is as follows:
\begin{align*}
\sT_{\Phi_{\mathrm{exp}}}= 
\begin{cases}
\sT(t), & t\in \left[\epsilon,1\right], \\
\frac{\sT(\epsilon)}{\epsilon}\, t, &  t\in \left[0,\epsilon\right).
\end{cases}
\end{align*}
Therefore, when $\epsilon=0$, $\sT_{\Phi_{\mathrm{exp}}}$ is convex, non-decreasing, invertible and satisfies that $\sT_{\Phi_{\mathrm{exp}}}(0)=0$. By Theorem~\ref{Thm:tightness}, we can choose $\Psi(t)=\sT_{\Phi_{\mathrm{exp}}}(t)$ in Theorem~\ref{Thm:excess_bounds_Psi_uniform}, or equivalently $\Gamma(t)=\sT_{\Phi_{\mathrm{exp}}}^{-1}(t)$ in Theorem~\ref{Thm:excess_bounds_Gamma_uniform}, which are optimal. To simplify the expression, using the fact that
\begin{align*}
1- \sqrt{1-t^2} & \geq \frac{t^2}{2}, \\
1-\frac{t+1}{2}e^{-B}-\frac{1-t}{2}e^B & = 1-1/2 e^B-1/2 e^{-B}+\frac{e^B-e^{-B}}2\, t,
\end{align*}
$\sT_{\Phi_{\mathrm{exp}}}$ can be lower bounded by
\begin{align*}
\wt{\sT}_{\Phi_{\mathrm{exp}}}(t)= \begin{cases}
\frac{t^2}{2},& t\leq \frac{e^{2B}-1}{e^{2B}+1},\\
\frac{1}{2}\paren*{\frac{e^{2B}-1}{e^{2B}+1}}\, t, & t> \frac{e^{2B}-1}{e^{2B}+1}.
\end{cases}   
\end{align*}
Thus, we adopt an upper bound of $\sT_{\Phi_{\mathrm{exp}}}^{-1}$ as follows:
\begin{align*}
\wt{\sT}_{\Phi_{\mathrm{exp}}}^{-1}(t)=
\begin{cases}
\sqrt{2t}, & t\leq \frac{1}{2}\paren*{\frac{e^{2B}-1}{e^{2B}+1}}^2,\\
2\paren*{\frac{e^{2B}+1}{e^{2B}-1}}\, t, & t> \frac{1}{2}\paren*{\frac{e^{2B}-1}{e^{2B}+1}}^2.
\end{cases}
\end{align*}
Therefore, by Theorem~\ref{Thm:excess_bounds_Psi_uniform} or Theorem~\ref{Thm:excess_bounds_Gamma_uniform}, setting $\e = 0$ yields the $\sH_{\mathrm{lin}}$-consistency estimation error bound for the exponential loss, valid for all $h \in \sH_{\mathrm{lin}}$:
\begin{multline}
\label{eq:exp-lin-est}
     \sR_{\ell_{0-1}}(h)-\sR_{\ell_{0-1},\sH_{\mathrm{lin}}}^*+\sM_{\ell_{0-1}, \sH_{\mathrm{lin}}}\\
     \leq 
     \begin{cases}
     \sqrt{2}\,\paren*{\sR_{\Phi_{\mathrm{exp}}}(h)- \sR_{\Phi_{\mathrm{exp}},\sH_{\mathrm{lin}}}^*+\sM_{\Phi_{\mathrm{exp}},\sH_{\mathrm{lin}}}}^{\frac12}, & \text{if } \sR_{\Phi_{\mathrm{exp}}}(h)- \sR_{\Phi_{\mathrm{exp}},\sH_{\mathrm{lin}}}^*\leq \frac{1}{2}\paren*{\frac{e^{2B}-1}{e^{2B}+1}}^2-\sM_{\Phi_{\mathrm{exp}},\sH_{\mathrm{lin}}},\\
     2\paren*{\frac{e^{2B}+1}{e^{2B}-1}}\paren*{\sR_{\Phi_{\mathrm{exp}}}(h)- \sR_{\Phi_{\mathrm{exp}},\sH_{\mathrm{lin}}}^*+\sM_{\Phi_{\mathrm{exp}},\sH_{\mathrm{lin}}}}, & \text{otherwise}.
     \end{cases}
\end{multline}
Since the $\paren*{\ell_{0-1},\sH_{\mathrm{lin}}}$-minimizability gap coincides with
the $\paren*{\ell_{0-1},\sH_{\mathrm{lin}}}$-approximation error and
$\paren*{\Phi_{\mathrm{exp}},\sH_{\mathrm{lin}}}$-minimizability gap coincides with
the $\paren*{\Phi_{\mathrm{exp}},\sH_{\mathrm{lin}}}$-approximation error for $B =\plus\infty$,
the inequality can be rewritten as follows:
\begin{multline*}
     \sR_{\ell_{0-1}}(h)- \sR_{\ell_{0-1},\sH_{\mathrm{all}}}^*\\
     \leq 
     \begin{cases}
      \sqrt{2}\,\bracket*{\sR_{\Phi_{\mathrm{exp}}}(h) - \sR_{\Phi_{\mathrm{exp}},\sH_{\mathrm{all}}}^*}^{\frac12} & \text{if } B = \plus\infty, \\
     \begin{cases}
    \sqrt{2}\,\bracket*{\sR_{\Phi_{\mathrm{exp}}}(h)- \sR_{\Phi_{\mathrm{exp}},\sH_{\mathrm{lin}}}^*+\sM_{\Phi_{\mathrm{exp}},\sH_{\mathrm{lin}}}}^{\frac12} & \text{if } \sR_{\Phi_{\mathrm{exp}}}(h)- \sR_{\Phi_{\mathrm{exp}},\sH_{\mathrm{lin}}}^*\leq \frac{1}{2}\paren*{\frac{e^{2B}-1}{e^{2B}+1}}^2-\sM_{\Phi_{\mathrm{exp}},\sH_{\mathrm{lin}}}, \\
    2\paren*{\frac{e^{2B}+1}{e^{2B}-1}}\paren*{\sR_{\Phi_{\mathrm{exp}}}(h)- \sR_{\Phi_{\mathrm{exp}},\sH_{\mathrm{lin}}}^*+\sM_{\Phi_{\mathrm{exp}},\sH_{\mathrm{lin}}}} & \text{otherwise}.
    \end{cases} & \text{otherwise}.
     \end{cases}
\end{multline*}
where the $\paren*{\Phi_{\mathrm{exp}},\sH_{\mathrm{lin}}}$-minimizability gap $\sM_{\Phi_{\mathrm{exp}},\sH_{\mathrm{lin}}}$ is characterized as below, which is less than
the $\paren*{\Phi_{\mathrm{exp}},\sH_{\mathrm{lin}}}$-approximation error when $B<\plus \infty$:
\begin{align*}
\sM_{\Phi_{\mathrm{exp}},\sH_{\mathrm{lin}}}
& = \sR_{\Phi_{\mathrm{exp}},\sH_{\mathrm{lin}}}^*-
\mathbb{E}_{X}\bracket*{2\sqrt{\eta(x)(1-\eta(x))}\mathds{1}_{1/2\log\abs*{\frac{\eta(x)}{1-\eta(x)}}\leq W\norm*{x}_p+B}}\\
& - \mathbb{E}_{X}\bracket*{\max\curl*{\eta(x),1-\eta(x)}e^{-(W\norm*{x}_p+B)}\mathds{1}_{1/2\log\abs*{\frac{\eta(x)}{1-\eta(x)}}> W\norm*{x}_p+B}}\\
&-\mathbb{E}_{X}\bracket*{\min\curl*{\eta(x),1-\eta(x)}e^{W\norm*{x}_p+B}\mathds{1}_{1/2\log\abs*{\frac{\eta(x)}{1-\eta(x)}}> W\norm*{x}_p+B}}\\
& < \sR_{\Phi_{\mathrm{exp}},\sH_{\mathrm{lin}}}^* - \mathbb{E}_{X}\bracket*{2\sqrt{\eta(x)(1-\eta(x))}}\\
& = \sR_{\Phi_{\mathrm{exp}},\sH_{\mathrm{lin}}}^* - \sR_{\Phi_{\mathrm{exp}},\sH_{\mathrm{all}}}^* .
\end{align*}
Therefore, the inequality for $B = \plus \infty$ coincides with the consistency excess error bound
known for the exponential loss \citep{Zhang2003,MohriRostamizadehTalwalkar2018} but the one for $B< \plus \infty$ is distinct and novel.

\subsubsection{Quadratic Loss}
For the quadratic loss $\Phi_{\mathrm{quad}}(\alpha)\colon=(1-\alpha)^2\mathds{1}_{\alpha\leq 1}$, for all $h\in \sH_{\mathrm{lin}}$ and $x\in \sX$:
\begin{equation*}
\begin{aligned}
\sC_{\Phi_{\mathrm{quad}}}(h,x,t)
&=t \Phi_{\mathrm{quad}}(h(x))+(1-t)\Phi_{\mathrm{quad}}(-h(x))\\
&=t\paren*{1-h(x)}^2\mathds{1}_{h(x)\leq 1}+(1-t)\paren*{1+h(x)}^2\mathds{1}_{h(x)\geq -1}.\\
\inf_{h\in\sH_{\mathrm{lin}}}\sC_{\Phi_{\mathrm{quad}}}(h,x,t)
&=\begin{cases}
4t(1-t), & \abs*{2t-1}\leq W\norm*{x}_p+B,\\
\max\curl*{t,1-t}\paren*{1-\paren*{W\norm*{x}_p+B}}^2
+\min\curl*{t,1-t}\paren*{1+W\norm*{x}_p+B}^2, &\abs*{2t-1}> W\norm*{x}_p+B.
\end{cases}
\end{aligned}
\end{equation*}
Therefore, the $\paren*{\Phi_{\mathrm{quad}},\sH_{\mathrm{lin}}}$-minimizability gap can be expressed as follows:
\begin{equation}
\begin{aligned}
\label{eq:M-quad-lin}
\sM_{\Phi_{\mathrm{quad}},\sH_{\mathrm{lin}}}
& = \sR_{\Phi_{\mathrm{quad}},\sH_{\mathrm{lin}}}^*-\mathbb{E}_{X}\bracket*{4\eta(x)(1-\eta(x))\mathds{1}_{\abs*{2\eta(x)-1}\leq W\norm*{x}_p+B}}\\
& - \mathbb{E}_{X}\bracket*{\max\curl*{\eta(x),1-\eta(x)}\paren*{1-\paren*{W\norm*{x}_p+B}}^2\mathds{1}_{\abs*{2\eta(x)-1}> W\norm*{x}_p+B}}\\
& - \mathbb{E}_{X}\bracket*{\min\curl*{\eta(x),1-\eta(x)}\paren*{1+\paren*{W\norm*{x}_p+B}}^2\mathds{1}_{\abs*{2\eta(x)-1}> W\norm*{x}_p+B}}
\end{aligned}
\end{equation}
Note $\paren*{\Phi_{\mathrm{quad}},\sH_{\mathrm{lin}}}$-minimizability gap coincides with
the $\paren*{\Phi_{\mathrm{quad}},\sH_{\mathrm{lin}}}$-approximation error $\sR_{\Phi_{\mathrm{quad}},\sH_{\mathrm{lin}}}^*-
\mathbb{E}_{X}\bracket*{4\eta(x)(1-\eta(x))}$ for $B \geq 1$.

For $\frac{1}2< t\leq1$, we have
\begin{align*}
\inf_{h\in\sH_{\mathrm{lin}}:h(x)<0}\sC_{\Phi_{\mathrm{quad}}}(h,x,t)
&=t+(1-t)\\
&=1\\
\inf_{x\in \sX}\inf_{h\in\sH_{\mathrm{lin}}:h(x)<0}\Delta\sC_{\Phi_{\mathrm{quad}},\sH_{\mathrm{lin}}}(h,x,t)
& =\inf_{x\in \sX}\paren*{\inf_{h\in\sH_{\mathrm{lin}}:h(x)<0}\sC_{\Phi_{\mathrm{quad}}}(h,x,t)-\inf_{h\in\sH_{\mathrm{lin}}}\sC_{\Phi_{\mathrm{quad}}}(h,x,t)}\\
&=\inf_{x\in \sX}\begin{cases}
1-4t(1-t),& 2t-1\leq W\norm*{x}_p+B,\\
1-t\paren*{1-\paren*{W\norm*{x}_p+B}}^2-(1-t)\paren*{1+W\norm*{x}_p+B}^2, & 2t-1> W\norm*{x}_p+B.
\end{cases}\\
&=\begin{cases}
1-4t(1-t),& 2t-1\leq B,\\
1-t\paren*{1-B}^2-(1-t)\paren*{1+B}^2,& 2t-1> B.
\end{cases}\\
&=\sT(2t-1)
\end{align*}
where $\sT$ is the increasing and convex function on $[0,1]$ defined by
\begin{align*}
\forall t\in[0,1], \quad
\sT(t)=\begin{cases}
t^2, & t\leq B,\\
2B \,t-B^2, & t> B.
\end{cases}
\end{align*}
By Definition~\ref{def:trans}, for any $\epsilon\geq 0$, the $\sH_{\mathrm{lin}}$-estimation error transformation of the quadratic loss is as follows:
\begin{align*}
\sT_{\Phi_{\mathrm{quad}}}= 
\begin{cases}
\sT(t), & t\in \left[\epsilon,1\right], \\
\frac{\sT(\epsilon)}{\epsilon}\, t, &  t\in \left[0,\epsilon\right).
\end{cases}
\end{align*}
Therefore, when $\epsilon=0$, $\sT_{\Phi_{\mathrm{quad}}}$ is convex, non-decreasing, invertible and satisfies that $\sT_{\Phi_{\mathrm{quad}}}(0)=0$. By Theorem~\ref{Thm:tightness}, we can choose $\Psi(t)=\sT_{\Phi_{\mathrm{quad}}}(t)$ in Theorem~\ref{Thm:excess_bounds_Psi_uniform}, or equivalently $\Gamma(t) = \sT_{\Phi_{\mathrm{quad}}}^{-1}(t)=
\begin{cases}
\sqrt{t}, & t \leq B^2 \\
\frac{t}{2B}+\frac{B}{2}, & t > B^2
\end{cases}$, in Theorem~\ref{Thm:excess_bounds_Gamma_uniform}, which are optimal. Thus, by Theorem~\ref{Thm:excess_bounds_Psi_uniform} or Theorem~\ref{Thm:excess_bounds_Gamma_uniform}, setting $\e = 0$ yields the $\sH_{\mathrm{lin}}$-consistency estimation error bound for the quadratic loss, valid for all $h \in \sH_{\mathrm{lin}}$:
\begin{multline}
\label{eq:quad-lin-est}
    \sR_{\ell_{0-1}}(h)- \sR_{\ell_{0-1},\sH_{\mathrm{lin}}}^*+\sM_{\ell_{0-1},\sH_{\mathrm{lin}}}\\
    \leq
    \begin{cases}
    \bracket*{\sR_{\Phi_{\mathrm{quad}}}(h)- \sR_{\Phi_{\mathrm{quad}},\sH_{\mathrm{lin}}}^*+\sM_{\Phi_{\mathrm{quad}},\sH_{\mathrm{lin}}}}^{\frac12}  & \text{if } \sR_{\Phi_{\mathrm{quad}}}(h)- \sR_{\Phi_{\mathrm{quad}},\sH_{\mathrm{lin}}}^*\leq B^2-\sM_{\Phi_{\mathrm{quad}},\sH_{\mathrm{lin}}} \\
    \frac{\sR_{\Phi_{\mathrm{quad}}}(h)- \sR_{\Phi_{\mathrm{quad}},\sH_{\mathrm{lin}}}^*+\sM_{\Phi_{\mathrm{quad}},\sH_{\mathrm{lin}}}}{2B}+\frac{B}{2} & \text{otherwise}
    \end{cases}
\end{multline}
Since the $\paren*{\ell_{0-1},\sH_{\mathrm{lin}}}$-minimizability gap coincides with
the $\paren*{\ell_{0-1},\sH_{\mathrm{lin}}}$-approximation error and
$\paren*{\Phi_{\mathrm{quad}},\sH_{\mathrm{lin}}}$-minimizability gap coincides with
the $\paren*{\Phi_{\mathrm{quad}},\sH_{\mathrm{lin}}}$-approximation error for $B \geq 1$,
the inequality can be rewritten as follows:
\begin{multline*}
     \sR_{\ell_{0-1}}(h)- \sR_{\ell_{0-1},\sH_{\mathrm{all}}}^*\\
     \leq 
     \begin{cases}
     \bracket*{\sR_{\Phi_{\mathrm{quad}}}(h) - \sR_{\Phi_{\mathrm{quad}},\sH_{\mathrm{all}}}^*}^{\frac12} & \text{if } B \geq 1\\
     \begin{cases}
    \bracket*{\sR_{\Phi_{\mathrm{quad}}}(h)- \sR_{\Phi_{\mathrm{quad}},\sH_{\mathrm{lin}}}^*+\sM_{\Phi_{\mathrm{quad}},\sH_{\mathrm{lin}}}}^{\frac12}  & \text{if } \sR_{\Phi_{\mathrm{quad}}}(h)- \sR_{\Phi_{\mathrm{quad}},\sH_{\mathrm{lin}}}^*\leq B^2-\sM_{\Phi_{\mathrm{quad}},\sH_{\mathrm{lin}}} \\
    \frac{\sR_{\Phi_{\mathrm{quad}}}(h)- \sR_{\Phi_{\mathrm{quad}},\sH_{\mathrm{lin}}}^*+\sM_{\Phi_{\mathrm{quad}},\sH_{\mathrm{lin}}}}{2B}+\frac{B}{2} & \text{otherwise}
    \end{cases} & \text{otherwise}
     \end{cases}
\end{multline*}
where the $\paren*{\Phi_{\mathrm{quad}},\sH_{\mathrm{lin}}}$-minimizability gap $\sM_{\Phi_{\mathrm{quad}},\sH_{\mathrm{lin}}}$ is characterized as below, which is less than
the $\paren*{\Phi_{\mathrm{quad}},\sH_{\mathrm{lin}}}$-approximation error when $B<1$:
\begin{align*}
\sM_{\Phi_{\mathrm{quad}},\sH_{\mathrm{lin}}}
& = \sR_{\Phi_{\mathrm{quad}},\sH_{\mathrm{lin}}}^*-\mathbb{E}_{X}\bracket*{4\eta(x)(1-\eta(x))\mathds{1}_{\abs*{2\eta(x)-1}\leq W\norm*{x}_p+B}}\\
& - \mathbb{E}_{X}\bracket*{\max\curl*{\eta(x),1-\eta(x)}\paren*{1-\paren*{W\norm*{x}_p+B}}^2\mathds{1}_{\abs*{2\eta(x)-1}> W\norm*{x}_p+B}}\\
& - \mathbb{E}_{X}\bracket*{\min\curl*{\eta(x),1-\eta(x)}\paren*{1+\paren*{W\norm*{x}_p+B}}^2\mathds{1}_{\abs*{2\eta(x)-1}> W\norm*{x}_p+B}}\\
& \leq \sR_{\Phi_{\mathrm{quad}},\sH_{\mathrm{lin}}}^* - \mathbb{E}_{X}\bracket*{4\eta(x)(1-\eta(x))}\\
& = \sR_{\Phi_{\mathrm{quad}},\sH_{\mathrm{lin}}}^* - \sR_{\Phi_{\mathrm{quad}},\sH_{\mathrm{all}}}^* .
\end{align*}
Therefore, the inequality for $B \geq 1$ coincides with the consistency excess error bound
known for the quadratic loss \citep{Zhang2003,bartlett2006convexity} but the one for $B< 1$ is distinct and novel.

\subsubsection{Sigmoid Loss}
\label{app:sig-lin}
For the sigmoid loss $\Phi_{\mathrm{sig}}(\alpha)\colon=1-\tanh(k\alpha),~k>0$,
for all $h\in \sH_{\mathrm{lin}}$ and $x\in \sX$:
\begin{equation*}
\begin{aligned}
\sC_{\Phi_{\mathrm{sig}}}(h,x,t)
&=t \Phi_{\mathrm{sig}}(h(x))+(1-t)\Phi_{\mathrm{sig}}(-h(x)),\\
&=t\paren*{1-\tanh(kh(x))}+(1-t)\paren*{1+\tanh(kh(x))}.\\
\inf_{h\in\sH_{\mathrm{lin}}}\sC_{\Phi_{\mathrm{sig}}}(h,x,t)
&=1-\abs*{1-2t}\tanh\paren*{k\paren*{W\norm*{x}_p+B}}
\end{aligned}
\end{equation*}
Therefore, the $\paren*{\Phi_{\mathrm{sig}},\sH_{\mathrm{lin}}}$-minimizability gap can be expressed as follows:
\begin{equation}
\begin{aligned}
\label{eq:M-sig-lin}
\sM_{\Phi_{\mathrm{sig}},\sH_{\mathrm{lin}}}
&= \sR_{\Phi_{\mathrm{sig}},\sH_{\mathrm{lin}}}^*-\mathbb{E}_{X}\bracket*{\inf_{h\in\sH_{\mathrm{lin}}}\sC_{\Phi_{\mathrm{sig}}}(h,x,\eta(x))}\\
&= \sR_{\Phi_{\mathrm{sig}},\sH_{\mathrm{lin}}}^*-\mathbb{E}_{X}\bracket*{1-\abs*{1-2\eta(x)}\tanh\paren*{k\paren*{W\norm*{x}_p+B}}}.
\end{aligned}
\end{equation}
Note $\paren*{\Phi_{\mathrm{sig}},\sH_{\mathrm{lin}}}$-minimizability gap coincides with
the $\paren*{\Phi_{\mathrm{sig}},\sH_{\mathrm{lin}}}$-approximation error $\sR_{\Phi_{\mathrm{sig}},\sH_{\mathrm{lin}}}^*-\mathbb{E}_{X}\bracket*{1-\abs*{1-2\eta(x)}}$ for $B = \plus \infty$.

For $\frac{1}2< t\leq1$, we have
\begin{align*}
\inf_{h\in\sH_{\mathrm{lin}}:h(x)<0}\sC_{\Phi_{\mathrm{sig}}}(h,x,t)
&=1-\abs*{1-2t}\tanh(0)\\
&=1.\\
\inf_{x\in \sX}\inf_{h\in\sH_{\mathrm{lin}}:h(x)<0}\Delta\sC_{\Phi_{\mathrm{sig}},\sH_{\mathrm{lin}}}(h,x,t)
&=\inf_{x\in \sX}\paren*{\inf_{h\in\sH_{\mathrm{lin}}:h(x)<0}\sC_{\Phi_{\mathrm{sig}}}(h,x,t)-\inf_{h\in\sH_{\mathrm{lin}}}\sC_{\Phi_{\mathrm{sig}}}(h,x,t)}\\
&=\inf_{x\in \sX}(2t-1)\tanh\paren*{k\paren*{W\norm*{x}_p+B}}\\
&=(2t-1)\tanh(kB)\\
&=\sT(2t-1)
\end{align*}
where $\sT$ is the increasing and convex function on $[0,1]$ defined by
\begin{align*}
\forall t\in[0,1],\; \sT(t)=\tanh(kB) \, t .
\end{align*}
By Definition~\ref{def:trans}, for any $\epsilon\geq 0$, the $\sH_{\mathrm{lin}}$-estimation error transformation of the sigmoid loss is as follows:
\begin{align*}
\sT_{\Phi_{\mathrm{sig}}}= \tanh(kB) \, t, \quad t \in [0,1],
\end{align*}
Therefore, $\sT_{\Phi_{\mathrm{sig}}}$ is convex, non-decreasing, invertible and satisfies that $\sT_{\Phi_{\mathrm{sig}}}(0)=0$. By Theorem~\ref{Thm:tightness}, we can choose $\Psi(t)=\tanh(kB)\,t$ in Theorem~\ref{Thm:excess_bounds_Psi_uniform}, or equivalently $\Gamma(t)=\frac{t}{\tanh(kB)}$ in Theorem~\ref{Thm:excess_bounds_Gamma_uniform}, which are optimal when $\e=0$.
Thus, by Theorem~\ref{Thm:excess_bounds_Psi_uniform} or Theorem~\ref{Thm:excess_bounds_Gamma_uniform}, setting $\e = 0$ yields the $\sH_{\mathrm{lin}}$-consistency estimation error bound for the sigmoid loss, valid for all $h \in \sH_{\mathrm{lin}}$:
\begin{align}
\label{eq:sig-lin-est}
     \sR_{\ell_{0-1}}(h)- \sR_{\ell_{0-1},\sH_{\mathrm{lin}}}^*\leq \frac{\sR_{\Phi_{\mathrm{sig}}}(h)- \sR_{\Phi_{\mathrm{sig}},\sH_{\mathrm{lin}}}^*+\sM_{\Phi_{\mathrm{sig}},\sH_{\mathrm{lin}}}}{\tanh(kB)}-\sM_{\ell_{0-1},\sH_{\mathrm{lin}}}.
\end{align}
Since the $\paren*{\ell_{0-1},\sH_{\mathrm{lin}}}$-minimizability gap coincides with
the $\paren*{\ell_{0-1},\sH_{\mathrm{lin}}}$-approximation error and
$\paren*{\Phi_{\mathrm{sig}},\sH_{\mathrm{lin}}}$-minimizability gap coincides with
the $\paren*{\Phi_{\mathrm{sig}},\sH_{\mathrm{lin}}}$-approximation error for $B = \plus \infty$,
the inequality can be rewritten as follows:
\begin{align}
\label{eq:sig-lin-est-2}
     \sR_{\ell_{0-1}}(h)- \sR_{\ell_{0-1},\sH_{\mathrm{all}}}^*
      \leq 
     \begin{cases}
     \sR_{\Phi_{\mathrm{sig}}}(h) - \sR_{\Phi_{\mathrm{sig}},\sH_{\mathrm{all}}}^* & \text{if } B = \plus \infty\\
     \frac{1}{\tanh(kB)} \bracket[\Big]{\sR_{\Phi_{\mathrm{sig}}}(h)
     - \mathbb{E}_{X}\bracket*{1-\abs*{1-2\eta(x)}\tanh\paren*{k\paren*{W\norm*{x}_p+B}}} } & \text{otherwise}.
     \end{cases}
\end{align}
The inequality for $B = \plus \infty$ coincides with the consistency excess error bound
known for the sigmoid loss \citep{Zhang2003,bartlett2006convexity,MohriRostamizadehTalwalkar2018} but the one for $B < \plus \infty$ is distinct and novel.  For $B<\plus \infty$, we have
\begin{align*}
\mathbb{E}_{X}\bracket*{1-\abs*{1-2\eta(x)}\tanh\paren*{k\paren*{W\norm*{x}_p+B}}}> \mathbb{E}_{X}\bracket*{1-\abs*{2\eta(x)-1}}= 2\mathbb{E}_X\bracket*{\min\curl*{\eta(x), 1 - \eta(x)}}= \sR_{\Phi_{\mathrm{hinge}},\sH_{\mathrm{all}}}^*.
\end{align*}
Therefore for $B<\plus \infty$,
\begin{align*}
\sR_{\Phi_{\mathrm{sig}}}(h) - \mathbb{E}_{X}\bracket*{1-\abs*{1-2\eta(x)}\tanh\paren*{k\paren*{W\norm*{x}_p+B}}} < \sR_{\Phi_{\mathrm{sig}}}(h) - \sR_{\Phi_{\mathrm{sig}},\sH_{\mathrm{all}}}^*.
\end{align*}
Note that: $\sR_{\Phi_{\mathrm{sig}},\sH_{\mathrm{all}}}^* = 2 \sR_{\ell_{0-1},\sH_{\mathrm{all}}}^* =2\mathbb{E}_X\bracket*{\min\curl*{\eta(x), 1 - \eta(x)}}$. Thus, the first
inequality (case $B = \plus \infty$) can be equivalently written as follows:
\begin{align*}
    \forall h \in \sH_{\mathrm{lin}}, \quad \sR_{\ell_{0-1}}(h) 
     \leq \sR_{\Phi_{\mathrm{sig}}}(h) - \mathbb{E}_X\bracket*{\min\curl*{\eta(x), 1 - \eta(x)}},
\end{align*}
which is a more informative upper bound than the standard
inequality $\sR_{\ell_{0-1}}(h) 
     \leq \sR_{\Phi_{\mathrm{sig}}}(h)$.

\subsubsection{\texorpdfstring{$\rho$}{rho}-Margin Loss}
\label{app:rho-lin}
For the $\rho$-margin loss $\Phi_{\rho}(\alpha)\colon=\min\curl*{1,\max\curl*{0,1-\frac{\alpha}{\rho}}},~\rho>0$,
for all $h\in \sH_{\mathrm{lin}}$ and $x\in \sX$:
\begin{equation*}
\begin{aligned}
\sC_{\Phi_{\rho}}(h,x,t)
&=t \Phi_{\rho}(h(x))+(1-t)\Phi_{\rho}(-h(x)),\\
&=t\min\curl*{1,\max\curl*{0,1-\frac{h(x)}{\rho}}}+(1-t)\min\curl*{1,\max\curl*{0,1+\frac{h(x)}{\rho}}}.\\
\inf_{h\in\sH_{\mathrm{lin}}}\sC_{\Phi_{\rho}}(h,x,t)
&=\min\curl*{t,1-t}+\max\curl*{t,1-t}\paren*{1-\frac{\min\curl*{W\norm*{x}_p+B,\rho}}{\rho}}.
\end{aligned}
\end{equation*}
Therefore, the $\paren*{\Phi_{\rho},\sH_{\mathrm{lin}}}$-minimizability gap can be expressed as follows:
\begin{equation}
\label{eq:M-rho-lin}
\begin{aligned}
\sM_{\Phi_{\rho},\sH_{\mathrm{lin}}}
& = \sR_{\Phi_{\rho},\sH_{\mathrm{lin}}}^*-\mathbb{E}_{X}\bracket*{\inf_{h\in\sH_{\mathrm{lin}}}\sC_{\Phi_{\rho}}(h,x,\eta(x))}\\
& = \sR_{\Phi_{\rho},\sH_{\mathrm{lin}}}^*-\mathbb{E}_{X}\bracket*{\min\curl*{\eta(x),1-\eta(x)}+\max\curl*{\eta(x),1-\eta(x)}\paren*{1-\frac{\min\curl*{W\norm*{x}_p+B,\rho}}{\rho}}}.
 \end{aligned}
\end{equation}
Note the $\paren*{\Phi_{\rho},\sH_{\mathrm{lin}}}$-minimizability gap coincides with
the $\paren*{\Phi_{\rho},\sH_{\mathrm{lin}}}$-approximation error $\sR_{\Phi_{\rho},\sH_{\mathrm{lin}}}^*-\mathbb{E}_{X}\bracket*{\min\curl*{\eta(x),1-\eta(x)}}$ for $B \geq \rho$.

For $\frac{1}2< t\leq1$, we have
\begin{align*}
\inf_{h\in\sH_{\mathrm{lin}}:h(x)<0}\sC_{\Phi_{\rho}}(h,x,t)
&=t+(1-t)\paren*{1-\frac{\min\curl*{W\norm*{x}_p+B,\rho}}{
\rho}}.\\
\inf_{x\in \sX}\inf_{h\in\sH_{\mathrm{lin}}:h(x)<0}\Delta\sC_{\Phi_{\rho},\sH_{\mathrm{lin}}}(h,x)
&=\inf_{x\in \sX}\paren*{\inf_{h\in\sH_{\mathrm{lin}}:h(x)<0}\sC_{\Phi_{\rho}}(h,x,t)-\inf_{h\in\sH_{\mathrm{lin}}}\sC_{\Phi_{\rho}}(h,x,t)}\\
&=\inf_{x\in \sX}(2t-1)\frac{\min\curl*{W\norm*{x}_p+B,\rho}}{\rho}\\
&=(2t-1)\frac{\min\curl*{B,\rho}}{\rho}\\
&=\sT(2t-1)
\end{align*}
where $\sT$ is the increasing and convex function on $[0,1]$ defined by
\begin{align*}
\forall t\in [0,1],\; \sT(t)=\frac{\min\curl*{B,\rho}}{\rho} \, t.    
\end{align*} 
By Definition~\ref{def:trans}, for any $\epsilon\geq 0$, the $\sH_{\mathrm{lin}}$-estimation error transformation of the $\rho$-margin loss is as follows:
\begin{align*}
\sT_{\Phi_{\rho}}= \frac{\min\curl*{B,\rho}}{\rho} \, t, \quad t \in [0,1],
\end{align*}
Therefore, $\sT_{\Phi_{\rho}}$ is convex, non-decreasing, invertible and satisfies that $\sT_{\Phi_{\rho}}(0)=0$. By Theorem~\ref{Thm:tightness}, we can choose $\Psi(t)=\frac{\min\curl*{B,\rho}}{\rho} \, t$ in Theorem~\ref{Thm:excess_bounds_Psi_uniform}, or equivalently $\Gamma(t)=\frac{\rho }{\min\curl*{B,\rho}} \, t$ in Theorem~\ref{Thm:excess_bounds_Gamma_uniform}, which are optimal when $\e=0$.
Thus, by Theorem~\ref{Thm:excess_bounds_Psi_uniform} or Theorem~\ref{Thm:excess_bounds_Gamma_uniform}, setting $\e = 0$ yields the $\sH_{\mathrm{lin}}$-consistency estimation error bound for the $\rho$-margin loss, valid for all $h \in \sH_{\mathrm{lin}}$:
\begin{align}
\label{eq:rho-lin-est}
     \sR_{\ell_{0-1}}(h)- \sR_{\ell_{0-1},\sH_{\mathrm{lin}}}^*\leq \frac{\rho\paren*{\sR_{\Phi_{\rho}}(h)- \sR_{\Phi_{\rho},\sH_{\mathrm{lin}}}^*+\sM_{\Phi_{\rho},\sH_{\mathrm{lin}}}}}{\min\curl*{B,\rho}}-\sM_{\ell_{0-1},\sH_{\mathrm{lin}}}.
\end{align}
Since the $\paren*{\ell_{0-1},\sH_{\mathrm{lin}}}$-minimizability gap coincides with
the $\paren*{\ell_{0-1},\sH_{\mathrm{lin}}}$-approximation error and
$\paren*{\Phi_{\rho},\sH_{\mathrm{lin}}}$-minimizability gap coincides with
the $\paren*{\Phi_{\rho},\sH_{\mathrm{lin}}}$-approximation error for $B \geq \rho$,
the inequality can be rewritten as follows:
\begin{align*}
     \sR_{\ell_{0-1}}(h)- \sR_{\ell_{0-1},\sH_{\mathrm{all}}}^*
     \leq 
     \begin{cases}
     \sR_{\Phi_{\rho}}(h) - \sR_{\Phi_{\rho},\sH_{\mathrm{all}}}^* & \text{if } B \geq \rho\\
     \frac{\rho \paren*{\sR_{\Phi_{\rho}}(h)
     - \mathbb{E}_{X}\bracket*{\min\curl*{\eta(x),1-\eta(x)}+\max\curl*{\eta(x),1-\eta(x)}\paren*{1-\frac{\min\curl*{W\norm*{x}_p+B,\rho}}{\rho}} } }}{B}  & \text{otherwise}. 
     \end{cases}
\end{align*}
Note that: $\sR_{\Phi_{\rho},\sH_{\mathrm{all}}}^* =  \sR_{\ell_{0-1},\sH_{\mathrm{all}}}^* =\mathbb{E}_X\bracket*{\min\curl*{\eta(x), 1 - \eta(x)}}$. Thus, the first
inequality (case $B \geq \rho$) can be equivalently written as follows:
\begin{align}
\label{eq:rho-lin-est-2}
    \forall h \in \sH_{\mathrm{lin}}, \quad \sR_{\ell_{0-1}}(h) 
     \leq \sR_{\Phi_{\rho}}(h).
\end{align}
The case $B \geq \rho$ is one of the ``trivial cases'' mentioned in Section~\ref{sec:general}, where the trivial inequality $\sR_{\ell_{0-1}}(h) \leq \sR_{\Phi_{\rho}}(h)$ can be obtained directly using the fact that $\ell_{0-1}$ is upper bounded by $\Phi_{\rho}$. This, however, does not imply that non-adversarial $\sH_{\mathrm{lin}}$-consistency estimation error bound for the $\rho$-margin loss is trivial when $B>\rho$ since it is optimal.

\subsection{One-Hidden-Layer ReLU Neural Network}
\label{app:derivation-NN}
As with the linear case, $\sH_{\mathrm{NN}}$ also satisfies the condition of Lemma~\ref{lemma:explicit_assumption_01} and thus the $\paren*{\ell_{0-1},\sH_{\mathrm{NN}}}$-minimizability gap coincides with
the $\paren*{\ell_{0-1},\sH_{\mathrm{NN}}}$-approximation error:
\begin{equation}
\begin{aligned}
\label{eq:M-01-NN}
\sM_{\ell_{0-1},\sH_{\mathrm{NN}}}
& = \sR_{\ell_{0-1},\sH_{\mathrm{NN}}}^*-\mathbb{E}_{X}\bracket*{\min\curl*{\eta(x),1-\eta(x)}} \\
&= \sR_{\ell_{0-1}, \sH_{\mathrm{NN}}}^* - \sR_{\ell_{0-1}, \sH_{\mathrm{all}}}^*.
\end{aligned}
\end{equation}
By the definition of $\sH_{\mathrm{NN}}$, for any $x \in \sX$, 
\begin{align*}
\curl*{h(x)\mid h\in\sH_{\mathrm{NN}}}= \bracket*{-\Lambda\paren*{W\norm*{x}_p+B}, \Lambda\paren*{W\norm*{x}_p+B}}.
\end{align*}

\subsubsection{Hinge Loss}
For the hinge loss $\Phi_{\mathrm{hinge}}(\alpha)\colon=\max\curl*{0,1 - \alpha}$, for all $h\in \sH_{\mathrm{NN}}$ and $x\in \sX$:
\begin{equation*}
\begin{aligned}
\sC_{\Phi_{\mathrm{hinge}}}(h,x,t)
& =t \Phi_{\mathrm{hinge}}(h(x))+(1-t)\Phi_{\mathrm{hinge}}(-h(x))\\
& =t\max\curl*{0,1-h(x)}+(1-t)\max\curl*{0,1+h(x)}.\\
\inf_{h\in\sH_{\mathrm{NN}}}\sC_{\Phi_{\mathrm{hinge}}}(h,x,t)
& = 1-\abs*{2t-1}\min\curl*{\Lambda W\norm*{x}_p+\Lambda B,1}.
\end{aligned}
\end{equation*}
Therefore, the $\paren*{\Phi_{\mathrm{hinge}},\sH_{\mathrm{NN}}}$-minimizability gap can be expressed as follows:
\begin{equation}
\begin{aligned}
\label{eq:M-hinge-NN}
\sM_{\Phi_{\mathrm{hinge}},\sH_{\mathrm{NN}}}
& = \sR_{\Phi_{\mathrm{hinge}},\sH_{\mathrm{NN}}}^*-\mathbb{E}_{X}\bracket*{1-\inf_{h\in\sH_{\mathrm{NN}}}\sC_{\Phi_{\mathrm{hinge}}}(h,x,\eta(x))}.\\
& = \sR_{\Phi_{\mathrm{hinge}},\sH_{\mathrm{NN}}}^*-\mathbb{E}_{X}\bracket*{1-\abs*{2\eta(x)-1}\min\curl*{\Lambda W\norm*{x}_p+\Lambda B,1}}.
\end{aligned}
\end{equation}
Note the $\paren*{\Phi_{\mathrm{hinge}},\sH_{\mathrm{NN}}}$-minimizability gap coincides with
the $\paren*{\Phi_{\mathrm{hinge}},\sH_{\mathrm{NN}}}$-approximation error $\sR_{\Phi_{\mathrm{hinge}},\sH_{\mathrm{NN}}}^*-\mathbb{E}_{X}\bracket*{1-\abs*{2\eta(x)-1}}$ for $\Lambda B \geq 1$.

For $\frac{1}2< t\leq1$, we have
\begin{align*}
\inf_{h\in\sH_{\mathrm{NN}}:h(x)<0}\sC_{\Phi_{\mathrm{hinge}}}(h,x,t)
& = t\max\curl*{0,1-0}+(1-t)\max\curl*{0,1+0}\\
& =1.\\
\inf_{x\in \sX} \inf_{h\in\sH_{\mathrm{NN}}:h(x)<0} \Delta\sC_{\Phi_{\mathrm{hinge}},\sH_{\mathrm{NN}}}(h,x,t)
& = \inf_{x\in \sX} \curl*{\inf_{h\in\sH_{\mathrm{NN}}:h(x)<0}\sC_{\Phi_{\mathrm{hinge}}}(h,x,t)-\inf_{h\in\sH_{\mathrm{NN}}}\sC_{\Phi_{\mathrm{hinge}}}(h,x,t)}\\
&=\inf_{x\in \sX}\paren*{2t-1}\min\curl*{\Lambda W\norm*{x}_p+\Lambda B,1}\\
&=(2t-1)\min\curl*{\Lambda B,1}\\
&=\sT(2t - 1),
\end{align*}
where $\sT$ is the increasing and convex function on $[0,1]$ defined by
\begin{align*}
\forall t \in [0,1],\; \sT(t) = \min \curl*{\Lambda B, 1} \, t .
\end{align*}
By Definition~\ref{def:trans}, for any $\epsilon\geq 0$, the $\sH_{\mathrm{NN}}$-estimation error transformation of the hinge loss is as follows:
\begin{align*}
\sT_{\Phi_{\mathrm{hinge}}}= \min \curl*{\Lambda B, 1} \, t, \quad t \in [0,1],
\end{align*}
Therefore, $\sT_{\Phi_{\mathrm{hinge}}}$ is convex, non-decreasing, invertible and satisfies that $\sT_{\Phi_{\mathrm{hinge}}}(0)=0$. By Theorem~\ref{Thm:tightness}, we can choose $\Psi(t)=\min\curl*{\Lambda B,1} \, t$ in Theorem~\ref{Thm:excess_bounds_Psi_uniform}, or, equivalently, $\Gamma(t) = \frac{t}{\min\curl*{\Lambda B, 1}}$ in Theorem~\ref{Thm:excess_bounds_Gamma_uniform}, which are optimal when $\e=0$. Thus, by Theorem~\ref{Thm:excess_bounds_Psi_uniform} or Theorem~\ref{Thm:excess_bounds_Gamma_uniform}, setting $\e = 0$ yields the $\sH_{\mathrm{NN}}$-consistency estimation error bound for the hinge loss, valid for all $h \in \sH_{\mathrm{NN}}$:
\begin{align}
\label{eq:hinge-NN-est}
     \sR_{\ell_{0-1}}(h)- \sR_{\ell_{0-1},\sH_{\mathrm{NN}}}^*
     \leq \frac{\sR_{\Phi_{\mathrm{hinge}}}(h)- \sR_{\Phi_{\mathrm{hinge}},\sH_{\mathrm{NN}}}^*+\sM_{\Phi_{\mathrm{hinge}},\sH_{\mathrm{NN}}}}{\min\curl*{\Lambda B,1}}-\sM_{\ell_{0-1}, \sH_{\mathrm{NN}}}.
\end{align}
Since the $\paren*{\ell_{0-1},\sH_{\mathrm{NN}}}$-minimizability gap coincides with
the $\paren*{\ell_{0-1},\sH_{\mathrm{NN}}}$-approximation error and
$\paren*{\Phi_{\mathrm{hinge}},\sH_{\mathrm{NN}}}$-minimizability gap coincides with
the $\paren*{\Phi_{\mathrm{hinge}},\sH_{\mathrm{NN}}}$-approximation error for $\Lambda B \geq 1$,
the inequality can be rewritten as follows:
\begin{align*}
     \sR_{\ell_{0-1}}(h)- \sR_{\ell_{0-1},\sH_{\mathrm{all}}}^*
      \leq 
     \begin{cases}
     \sR_{\Phi_{\mathrm{hinge}}}(h) - \sR_{\Phi_{\mathrm{hinge}},\sH_{\mathrm{all}}}^* & \text{if } \Lambda B \geq 1\\
     \frac{1}{\Lambda B} \bracket[\Big]{\sR_{\Phi_{\mathrm{hinge}}}(h)
     - \mathbb{E}_{X}\bracket*{1-\abs*{2\eta(x)-1}\min\curl*{\Lambda W\norm*{x}_p+\Lambda B,1}\
     } } & \text{otherwise}.
     \end{cases}
\end{align*}
The inequality for $\Lambda B \geq 1$ coincides with the consistency excess error bound
known for the hinge loss \citep{Zhang2003,bartlett2006convexity,MohriRostamizadehTalwalkar2018} but the one for $\Lambda B < 1$ is distinct and novel.  For $\Lambda B < 1$, we have
\begin{align*}
\mathbb{E}_{X}\bracket*{1-\abs*{2\eta(x)-1}\min\curl*{\Lambda W\norm*{x}_p+\Lambda B,1}}> \mathbb{E}_{X}\bracket*{1-\abs*{2\eta(x)-1}}= 2\mathbb{E}_X\bracket*{\min\curl*{\eta(x), 1 - \eta(x)}}= \sR_{\Phi_{\mathrm{hinge}},\sH_{\mathrm{all}}}^*.
\end{align*}
Therefore for $\Lambda B < 1$, 
\begin{align*}
\sR_{\Phi_{\mathrm{hinge}}}(h) - \mathbb{E}_{X}\bracket*{1-\abs*{2\eta(x)-1}\min\curl*{\Lambda W\norm*{x}_p+\Lambda B,1}} < \sR_{\Phi_{\mathrm{hinge}}}(h) - \sR_{\Phi_{\mathrm{hinge}},\sH_{\mathrm{all}}}^*.
\end{align*}
Note that: $\sR_{\Phi_{\mathrm{hinge}},\sH_{\mathrm{all}}}^* = 2 \sR_{\ell_{0-1},\sH_{\mathrm{all}}}^* =2\mathbb{E}_X\bracket*{\min\curl*{\eta(x), 1 - \eta(x)}}$. Thus, the first
inequality (case $\Lambda B \geq 1$) can be equivalently written as follows:
\begin{align*}
    \forall h \in \sH_{\mathrm{NN}}, \quad \sR_{\ell_{0-1}}(h) 
     \leq \sR_{\Phi_{\mathrm{hinge}}}(h) - \mathbb{E}_X\bracket*{\min\curl*{\eta(x), 1 - \eta(x)}},
\end{align*}
which is a more informative upper bound than the standard
inequality $\sR_{\ell_{0-1}}(h) 
     \leq \sR_{\Phi_{\mathrm{hinge}}}(h)$.

\subsubsection{Logistic Loss}

For the logistic loss $\Phi_{\mathrm{log}}(\alpha)\colon=\log_2(1+e^{-\alpha})$, for all $h\in \sH_{\mathrm{NN}}$ and $x\in \sX$:
\small
\begin{equation*}
\begin{aligned}
\sC_{\Phi_{\mathrm{log}}}(h,x,t)
& = t \Phi_{\mathrm{log}}(h(x))+(1-t)\Phi_{\mathrm{log}}(-h(x)),\\
& = t\log_2\paren*{1+e^{-h(x)}}+(1-t)\log_2\paren*{1+e^{h(x)}}.\\
\inf_{h\in\sH_{\mathrm{NN}}}\sC_{\Phi_{\mathrm{log}}}(h,x,t)
&=\begin{cases}
-t\log_2(t)-(1-t)\log_2(1-t) & \text{if }\log\abs*{\frac{t}{1-t}}\leq \Lambda W\norm*{x}_p+ \Lambda B,\\
\max\curl*{t,1-t}\log_2\paren*{1+e^{-(\Lambda W\norm*{x}_p+\Lambda B)}}+\min\curl*{t,1-t}\log_2\paren*{1+e^{\Lambda W\norm*{x}_p+\Lambda B}} &\text{if }\log\abs*{\frac{t}{1-t}}> \Lambda W\norm*{x}_p+\Lambda B.
\end{cases}
\end{aligned}
\end{equation*}
\normalsize
Therefore, the $\paren*{\Phi_{\mathrm{log}},\sH_{\mathrm{NN}}}$-minimizability gap can be expressed as follows:
\begin{equation}
\begin{aligned}
\label{eq:M-log-NN}
\sM_{\Phi_{\mathrm{log}},\sH_{\mathrm{NN}}}
& = \sR_{\Phi_{\mathrm{log}},\sH_{\mathrm{NN}}}^*-
\mathbb{E}_{X}\bracket*{\inf_{h\in\sH_{\mathrm{NN}}}\sC_{\Phi_{\mathrm{log}}}(h,x,\eta(x))}\\
& = \sR_{\Phi_{\mathrm{log}},\sH_{\mathrm{NN}}}^*-
\mathbb{E}_{X}\bracket*{-\eta(x)\log_2(\eta(x))-(1-\eta(x))\log_2(1-\eta(x))\mathds{1}_{\log\abs*{\frac{\eta(x)}{1-\eta(x)}}\leq \Lambda W\norm*{x}_p+\Lambda B}}\\
& - \mathbb{E}_{X}\bracket*{\max\curl*{\eta(x),1-\eta(x)}\log_2\paren*{1+e^{-(\Lambda W\norm*{x}_p+\Lambda B)}}\mathds{1}_{\log\abs*{\frac{\eta(x)}{1-\eta(x)}}> \Lambda W\norm*{x}_p+\Lambda B}}\\
&-\mathbb{E}_{X}\bracket*{\min\curl*{\eta(x),1-\eta(x)}\log_2\paren*{1+e^{\Lambda W\norm*{x}_p+\Lambda B}}\mathds{1}_{\log\abs*{\frac{\eta(x)}{1-\eta(x)}}> \Lambda W\norm*{x}_p+\Lambda B}}
\end{aligned}
\end{equation}
Note $\paren*{\Phi_{\mathrm{log}},\sH_{\mathrm{NN}}}$-minimizability gap coincides with
the $\paren*{\Phi_{\mathrm{log}},\sH_{\mathrm{NN}}}$-approximation error $\sR_{\Phi_{\mathrm{log}},\sH_{\mathrm{NN}}}^*-
\mathbb{E}_{X}\bracket*{-\eta(x)\log_2(\eta(x))-(1-\eta(x))\log_2(1-\eta(x))}$ for $\Lambda B =\plus\infty$.

For $\frac{1}2< t\leq1$, we have
\begin{align*}
\inf_{h\in\sH_{\mathrm{NN}}:h(x)<0}\sC_{\Phi_{\mathrm{log}}}(h,x,t)
& = t\log_2\paren*{1+e^{-0}}+(1-t)\log_2\paren*{1+e^{0}} \\
& = 1, \\
\inf_{x\in \sX}\inf_{h\in\sH_{\mathrm{NN}}:h(x)<0}\Delta\sC_{\Phi_{\mathrm{log}},\sH_{\mathrm{NN}}}(h,x,t)
&=\inf_{x\in \sX}\paren*{\inf_{h\in\sH_{\mathrm{NN}}:h(x)<0}\sC_{\Phi_{\mathrm{log}}}(h,x,t)-\inf_{h\in\sH_{\mathrm{NN}}}\sC_{\Phi_{\mathrm{log}}}(h,x,t)}\\
&=\inf_{x\in \sX}\begin{cases}
1+t\log_2(t)+(1-t)\log_2(1-t)\\
\text{if }\log\frac{t}{1-t}\leq \Lambda W\norm*{x}_p+\Lambda B,\\
1-t\log_2\paren*{1+e^{-(\Lambda W\norm*{x}_p+\Lambda B)}}-(1-t)\log_2\paren*{1+e^{\Lambda W\norm*{x}_p+\Lambda B}}\\
\text{if }\log\frac{t}{1-t}> \Lambda W\norm*{x}_p+\Lambda B.
\end{cases}\\
&=\begin{cases}
1+t\log_2(t)+(1-t)\log_2(1-t) & \text{if }\log\frac{t}{1-t}\leq \Lambda B,\\
1-t\log_2\paren*{1+e^{-\Lambda B}}-(1-t)\log_2\paren*{1+e^{\Lambda B}} & \text{if }\log\frac{t}{1-t}> \Lambda B.
\end{cases}\\
&=\sT(2t-1),
\end{align*}
where $\sT$ is the increasing and convex function on $[0,1]$ defined by
\begin{align*}
\forall t\in[0,1], \quad
\sT(t)=\begin{cases}
\frac{t+1}{2}\log_2(t+1)+\frac{1-t}{2}\log_2(1-t),\quad &  t\leq \frac{e^{\Lambda B}-1}{e^{\Lambda B}+1},\\
1-\frac{t+1}{2}\log_2(1+e^{-\Lambda B})-\frac{1-t}{2}\log_2(1+e^{\Lambda B}),\quad & t> \frac{e^{\Lambda B}-1}{e^{\Lambda B}+1}.
\end{cases}
\end{align*}
By Definition~\ref{def:trans}, for any $\epsilon\geq 0$, the $\sH_{\mathrm{NN}}$-estimation error transformation of the logistic loss is as follows:
\begin{align*}
\sT_{\Phi_{\mathrm{log}}}= 
\begin{cases}
\sT(t), & t\in \left[\epsilon,1\right], \\
\frac{\sT(\epsilon)}{\epsilon}\, t, &  t\in \left[0,\epsilon\right).
\end{cases}
\end{align*}
Therefore, when $\epsilon=0$, $\sT_{\Phi_{\mathrm{log}}}$ is convex, non-decreasing, invertible and satisfies that $\sT_{\Phi_{\mathrm{log}}}(0)=0$. By Theorem~\ref{Thm:tightness}, we can choose $\Psi(t)=\sT_{\Phi_{\mathrm{log}}}(t)$ in Theorem~\ref{Thm:excess_bounds_Psi_uniform}, or equivalently $\Gamma(t)=\sT_{\Phi_{\mathrm{log}}}^{-1}(t)$ in Theorem~\ref{Thm:excess_bounds_Gamma_uniform}, which are optimal. To simplify the expression, using the fact that
\begin{align*}
\frac{t+1}{2}\log_2(t+1)+\frac{1-t}{2}\log_2(1-t)
&=
1-\paren*{-\frac{t+1}{2}\log_2\paren*{\frac{t+1}{2}}-\frac{1-t}{2}\log_2\paren*{\frac{1-t}{2}}}\\
&\geq 1 - \sqrt{4\frac{1-t}{2} \frac{t+1}{2}}\\
& = 1- \sqrt{1-t^2}\\
& \geq \frac{t^2}{2},\\
1-\frac{t+1}{2}\log_2(1+e^{-\Lambda B})-\frac{1-t}{2}\log_2(1+e^{\Lambda B})
&=\frac{1}{2}\log_2\paren*{\frac{4}{2+e^{-\Lambda B}+e^{\Lambda B}}}+1/2\log_2\paren*{\frac{1+e^{\Lambda B}}{1+e^{-\Lambda B}}}\, t,
\end{align*}
$\sT_{\Phi_{\mathrm{log}}}$ can be lower bounded by
\begin{align*}
\wt{\sT}_{\Phi_{\mathrm{log}}}(t)= \begin{cases}
\frac{t^2}{2},& t\leq \frac{e^{\Lambda B}-1}{e^{\Lambda B}+1},\\
\frac{1}{2}\paren*{\frac{e^{\Lambda B}-1}{e^{\Lambda B}+1}}\, t, & t> \frac{e^{\Lambda B}-1}{e^{\Lambda B}+1}.
\end{cases}   
\end{align*}
Thus, we adopt an upper bound of $\sT_{\Phi_{\mathrm{log}}}^{-1}$ as follows:
\begin{align*}
\wt{\sT}_{\Phi_{\mathrm{log}}}^{-1}(t)=\begin{cases}
\sqrt{2t}, & t\leq \frac{1}{2}\paren*{\frac{e^{\Lambda B}-1}{e^{\Lambda B}+1}}^2,\\
2\paren*{\frac{e^{\Lambda B}+1}{e^{\Lambda B}-1}}\, t, & t> \frac{1}{2}\paren*{\frac{e^{\Lambda B}-1}{e^{\Lambda B}+1}}^2.
\end{cases}
\end{align*}
Therefore, by Theorem~\ref{Thm:excess_bounds_Psi_uniform} or Theorem~\ref{Thm:excess_bounds_Gamma_uniform}, setting $\e = 0$ yields the $\sH_{\mathrm{NN}}$-consistency estimation error bound for the logistic loss, valid for all $h \in \sH_{\mathrm{NN}}$:
\begin{multline}
\label{eq:log-NN-est}
     \sR_{\ell_{0-1}}(h)-\sR_{\ell_{0-1},\sH_{\mathrm{NN}}}^*+\sM_{\ell_{0-1}, \sH_{\mathrm{NN}}}\\
     \leq 
     \begin{cases}
     \sqrt{2}\,\paren*{\sR_{\Phi_{\mathrm{log}}}(h)- \sR_{\Phi_{\mathrm{log}},\sH_{\mathrm{NN}}}^*+\sM_{\Phi_{\mathrm{log}},\sH_{\mathrm{NN}}}}^{\frac12}, & \text{if } \sR_{\Phi_{\mathrm{log}}}(h)- \sR_{\Phi_{\mathrm{log}},\sH_{\mathrm{NN}}}^*\leq \frac{1}{2}\paren*{\frac{e^{\Lambda B}-1}{e^{\Lambda B}+1}}^2-\sM_{\Phi_{\mathrm{log}},\sH_{\mathrm{NN}}}\\
     2\paren*{\frac{e^{\Lambda B}+1}{e^{\Lambda B}-1}}\paren*{\sR_{\Phi_{\mathrm{log}}}(h)- \sR_{\Phi_{\mathrm{log}},\sH_{\mathrm{NN}}}^*+\sM_{\Phi_{\mathrm{log}},\sH_{\mathrm{NN}}}}, & \text{otherwise}
     \end{cases}
\end{multline}
Since the $\paren*{\ell_{0-1},\sH_{\mathrm{NN}}}$-minimizability gap coincides with
the $\paren*{\ell_{0-1},\sH_{\mathrm{NN}}}$-approximation error and
$\paren*{\Phi_{\mathrm{log}},\sH_{\mathrm{NN}}}$-minimizability gap coincides with
the $\paren*{\Phi_{\mathrm{log}},\sH_{\mathrm{NN}}}$-approximation error for $\Lambda B =\plus\infty$,
the inequality can be rewritten as follows:
\begin{align*}
     &\sR_{\ell_{0-1}}(h)- \sR_{\ell_{0-1},\sH_{\mathrm{all}}}^*\\
     &\quad \leq 
     \begin{cases}
      \sqrt{2}\,\bracket*{\sR_{\Phi_{\mathrm{log}}}(h) - \sR_{\Phi_{\mathrm{log}},\sH_{\mathrm{all}}}^*}^{\frac12} & \text{if } \Lambda B = \plus\infty \\
     \begin{cases}
    \sqrt{2}\,\bracket*{\sR_{\Phi_{\mathrm{log}}}(h)- \sR_{\Phi_{\mathrm{log}},\sH_{\mathrm{NN}}}^*+\sM_{\Phi_{\mathrm{log}},\sH_{\mathrm{NN}}}}^{\frac12}  & \text{if } \sR_{\Phi_{\mathrm{log}}}(h)- \sR_{\Phi_{\mathrm{log}},\sH_{\mathrm{NN}}}^*\leq \frac{1}{2}\paren*{\frac{e^{\Lambda B}-1}{e^{\Lambda B}+1}}^2-\sM_{\Phi_{\mathrm{log}},\sH_{\mathrm{NN}}} \\
    2\paren*{\frac{e^{\Lambda B}+1}{e^{\Lambda B}-1}}\paren*{\sR_{\Phi_{\mathrm{log}}}(h)- \sR_{\Phi_{\mathrm{log}},\sH_{\mathrm{NN}}}^*+\sM_{\Phi_{\mathrm{log}},\sH_{\mathrm{NN}}}} & \text{otherwise}
    \end{cases} & \text{otherwise}
     \end{cases}
\end{align*}
where the $\paren*{\Phi_{\mathrm{log}},\sH_{\mathrm{NN}}}$-minimizability gap $\sM_{\Phi_{\mathrm{log}},\sH_{\mathrm{NN}}}$ is characterized as below,which is less than
the $\paren*{\Phi_{\mathrm{log}},\sH_{\mathrm{NN}}}$-approximation error when $\Lambda B<\plus \infty$:
\begin{align*}
\sM_{\Phi_{\mathrm{log}},\sH_{\mathrm{NN}}}
& = \sR_{\Phi_{\mathrm{log}},\sH_{\mathrm{NN}}}^*-
\mathbb{E}_{X}\bracket*{-\eta(x)\log_2(\eta(x))-(1-\eta(x))\log_2(1-\eta(x))\mathds{1}_{\log_2\abs*{\frac{\eta(x)}{1-\eta(x)}}\leq \Lambda W\norm*{x}_p+\Lambda B}}\\
& - \mathbb{E}_{X}\bracket*{\max\curl*{\eta(x),1-\eta(x)}\log_2\paren*{1+e^{-(\Lambda W\norm*{x}_p+\Lambda B)}}\mathds{1}_{\log_2\abs*{\frac{\eta(x)}{1-\eta(x)}}> \Lambda W\norm*{x}_p+\Lambda B}}\\
&-\bracket*{\min\curl*{\eta(x),1-\eta(x)}\log_2\paren*{1+e^{\Lambda W\norm*{x}_p+\Lambda B}}\mathds{1}_{\log_2\abs*{\frac{\eta(x)}{1-\eta(x)}}> \Lambda W\norm*{x}_p+\Lambda B}}\\
& < \sR_{\Phi_{\mathrm{log}},\sH_{\mathrm{NN}}}^* - \mathbb{E}_{X}\bracket*{-\eta(x)\log_2(\eta(x))-(1-\eta(x))\log_2(1-\eta(x))}\\
& = \sR_{\Phi_{\mathrm{log}},\sH_{\mathrm{NN}}}^* - \sR_{\Phi_{\mathrm{log}},\sH_{\mathrm{all}}}^* .
\end{align*}
Therefore, the inequality for $\Lambda B = \plus \infty$ coincides with the consistency excess error bound
known for the logistic loss \citep{Zhang2003,MohriRostamizadehTalwalkar2018} but the one for $\Lambda B< \plus \infty$ is distinct and novel.

\subsubsection{Exponential Loss}
For the exponential loss $\Phi_{\mathrm{exp}}(\alpha)\colon=e^{-\alpha}$, for all $h\in \sH_{\mathrm{NN}}$ and $x\in \sX$:
\begin{equation*}
\begin{aligned}
\sC_{\Phi_{\mathrm{exp}}}(h,x,t)
&=t \Phi_{\mathrm{exp}}(h(x))+(1-t)\Phi_{\mathrm{exp}}(-h(x))\\
&=te^{-h(x)}+(1-t)e^{h(x)}.\\
\inf_{h\in\sH_{\mathrm{NN}}}\sC_{\Phi_{\mathrm{exp}}}(h,x,t)
&=\begin{cases}
2\sqrt{t(1-t)} & \text{if }1/2\log\abs*{\frac{t}{1-t}}\leq \Lambda W\norm*{x}_p+\Lambda B\\
\max\curl*{t,1-t}e^{-(\Lambda W\norm*{x}_p+ \Lambda B)}+\min\curl*{t,1-t}e^{\Lambda W\norm*{x}_p+\Lambda B} & \text{if }1/2\log\abs*{\frac{t}{1-t}}> \Lambda W\norm*{x}_p+ \Lambda B.
\end{cases}
\end{aligned}
\end{equation*}
Therefore, the $\paren*{\Phi_{\mathrm{exp}},\sH_{\mathrm{NN}}}$-minimizability gap can be expressed as follows:
\begin{equation}
\begin{aligned}
\label{eq:M-exp-NN}
\sM_{\Phi_{\mathrm{exp}},\sH_{\mathrm{NN}}}
& = \sR_{\Phi_{\mathrm{exp}},\sH_{\mathrm{NN}}}^*-
\mathbb{E}_{X}\bracket*{\inf_{h\in\sH_{\mathrm{NN}}}\sC_{\Phi_{\mathrm{exp}}}(h,x,\eta(x))}\\
& = \sR_{\Phi_{\mathrm{exp}},\sH_{\mathrm{NN}}}^*-
\mathbb{E}_{X}\bracket*{2\sqrt{\eta(x)(1-\eta(x))}\mathds{1}_{1/2\log\abs*{\frac{\eta(x)}{1-\eta(x)}}\leq \Lambda W\norm*{x}_p+\Lambda B}}\\
& - \mathbb{E}_{X}\bracket*{\max\curl*{\eta(x),1-\eta(x)}e^{-(\Lambda W\norm*{x}_p+\Lambda B)}\mathds{1}_{1/2\log\abs*{\frac{\eta(x)}{1-\eta(x)}}> \Lambda W\norm*{x}_p+\Lambda B}}\\
&-\mathbb{E}_{X}\bracket*{\min\curl*{\eta(x),1-\eta(x)}e^{\Lambda W\norm*{x}_p+\Lambda B}\mathds{1}_{1/2\log\abs*{\frac{\eta(x)}{1-\eta(x)}}>\Lambda W\norm*{x}_p+\Lambda B}}.
\end{aligned}
\end{equation}
Note $\paren*{\Phi_{\mathrm{exp}},\sH_{\mathrm{NN}}}$-minimizability gap coincides with
the $\paren*{\Phi_{\mathrm{exp}},\sH_{\mathrm{NN}}}$-approximation error $\sR_{\Phi_{\mathrm{exp}},\sH_{\mathrm{NN}}}^*-
\mathbb{E}_{X}\bracket*{2\sqrt{\eta(x)(1-\eta(x))}}$ for $\Lambda B =\plus\infty$.

For $\frac{1}2< t\leq1$, we have
\begin{align*}
\inf_{h\in\sH_{\mathrm{NN}}:h(x)<0}\sC_{\Phi_{\mathrm{exp}}}(h,x,t)
&=te^{-0}+(1-t)e^{0}\\
&=1.\\
\inf_{x\in \sX}\inf_{h\in\sH_{\mathrm{NN}}:h(x)<0}\Delta\sC_{\Phi_{\mathrm{exp}},\sH_{\mathrm{NN}}}(h,x,t)
&=\inf_{x\in \sX}\paren*{\inf_{h\in\sH_{\mathrm{NN}}:h(x)<0}\sC_{\Phi_{\mathrm{exp}}}(h,x,t)-\inf_{h\in\sH_{\mathrm{NN}}}\sC_{\Phi_{\mathrm{exp}}}(h,x,t)}\\
&=\inf_{x\in \sX}\begin{cases}
1-2\sqrt{t(1-t)} & \text{if }1/2\log\frac{t}{1-t}\leq \Lambda W\norm*{x}_p+\Lambda B,\\
1-te^{-(\Lambda W\norm*{x}_p+\Lambda B)}-(1-t)e^{\Lambda W\norm*{x}_p+\Lambda B} & \text{if }1/2\log\frac{t}{1-t}>\Lambda W\norm*{x}_p+\Lambda B.
\end{cases}\\
&=\begin{cases}
1-2\sqrt{t(1-t)}, & 1/2\log\frac{t}{1-t}\leq \Lambda B\\
1-te^{-\Lambda B}-(1-t)e^{\Lambda B}, & 1/2\log\frac{t}{1-t}> \Lambda B
\end{cases}\\
&=\sT(2t-1),
\end{align*}
where $\sT$ is the increasing and convex function on $[0,1]$ defined by
\begin{align*}
\forall t\in[0,1], \quad 
\sT(t)=\begin{cases}
1-\sqrt{1-t^2}, & t\leq \frac{e^{2\Lambda B}-1}{e^{2\Lambda B}+1},\\
1-\frac{t+1}{2}e^{-\Lambda B}-\frac{1-t}{2}e^{\Lambda B}, & t> \frac{e^{2\Lambda B}-1}{e^{2\Lambda B}+1}.
\end{cases}
\end{align*}
By Definition~\ref{def:trans}, for any $\epsilon\geq 0$, the $\sH_{\mathrm{lin}}$-estimation error transformation of the exponential loss is as follows:
\begin{align*}
\sT_{\Phi_{\mathrm{exp}}}= 
\begin{cases}
\sT(t), & t\in \left[\epsilon,1\right], \\
\frac{\sT(\epsilon)}{\epsilon}\, t, &  t\in \left[0,\epsilon\right).
\end{cases}
\end{align*}
Therefore, when $\epsilon=0$, $\sT_{\Phi_{\mathrm{exp}}}$ is convex, non-decreasing, invertible and satisfies that $\sT_{\Phi_{\mathrm{exp}}}(0)=0$. By Theorem~\ref{Thm:tightness}, we can choose $\Psi(t)=\sT_{\Phi_{\mathrm{exp}}}(t)$ in Theorem~\ref{Thm:excess_bounds_Psi_uniform}, or equivalently $\Gamma(t)=\sT_{\Phi_{\mathrm{exp}}}^{-1}(t)$ in Theorem~\ref{Thm:excess_bounds_Gamma_uniform}, which are optimal.  To simplify the expression, using the fact that
\begin{align*}
1- \sqrt{1-t^2} & \geq \frac{t^2}{2}, \\
1-\frac{t+1}{2}e^{-\Lambda B}-\frac{1-t}{2}e^{\Lambda B} & = 1-1/2 e^{\Lambda B}-1/2 e^{-\Lambda B}+\frac{e^{\Lambda B}-e^{-\Lambda B}}2\, t,
\end{align*}
$\sT_{\Phi_{\mathrm{exp}}}$ can be lower bounded by
\begin{align*}
\wt{\sT}_{\Phi_{\mathrm{exp}}}(t)= \begin{cases}
\frac{t^2}{2},& t\leq \frac{e^{2\Lambda B}-1}{e^{2\Lambda B}+1},\\
\frac{1}{2}\paren*{\frac{e^{2\Lambda B}-1}{e^{2\Lambda B}+1}}\, t, & t> \frac{e^{2\Lambda B}-1}{e^{2\Lambda B}+1}.
\end{cases}   
\end{align*}
Thus, we adopt an upper bound of $\sT_{\Phi_{\mathrm{exp}}}^{-1}$ as follows:
\begin{align*}
\wt{\sT}_{\Phi_{\mathrm{exp}}}^{-1}(t)=
\begin{cases}
\sqrt{2t}, & t\leq \frac{1}{2}\paren*{\frac{e^{2\Lambda B}-1}{e^{2B}+1}}^2,\\
2\paren*{\frac{e^{2\Lambda B}+1}{e^{2\Lambda B}-1}}\, t, & t> \frac{1}{2}\paren*{\frac{e^{2\Lambda B}-1}{e^{2\Lambda B}+1}}^2.
\end{cases}
\end{align*}
Therefore, by Theorem~\ref{Thm:excess_bounds_Psi_uniform} or Theorem~\ref{Thm:excess_bounds_Gamma_uniform}, setting $\e = 0$ yields the $\sH_{\mathrm{NN}}$-consistency estimation error bound for the exponential loss, valid for all $h \in \sH_{\mathrm{NN}}$:
\begin{multline}
\label{eq:exp-NN-est}
     \sR_{\ell_{0-1}}(h)-\sR_{\ell_{0-1},\sH_{\mathrm{NN}}}^*+\sM_{\ell_{0-1}, \sH_{\mathrm{NN}}}\\
     \leq 
     \begin{cases}
     \sqrt{2}\,\paren*{\sR_{\Phi_{\mathrm{exp}}}(h)- \sR_{\Phi_{\mathrm{exp}},\sH_{\mathrm{NN}}}^*+\sM_{\Phi_{\mathrm{exp}},\sH_{\mathrm{NN}}}}^{\frac12}, & \text{if } \sR_{\Phi_{\mathrm{exp}}}(h)- \sR_{\Phi_{\mathrm{exp}},\sH_{\mathrm{NN}}}^*\leq \frac{1}{2}\paren*{\frac{e^{2\Lambda B}-1}{e^{2\Lambda B}+1}}^2-\sM_{\Phi_{\mathrm{exp}},\sH_{\mathrm{NN}}},\\
     2\paren*{\frac{e^{2\Lambda B}+1}{e^{2\Lambda B}-1}}\paren*{\sR_{\Phi_{\mathrm{exp}}}(h)- \sR_{\Phi_{\mathrm{exp}},\sH_{\mathrm{NN}}}^*+\sM_{\Phi_{\mathrm{exp}},\sH_{\mathrm{NN}}}}, & \text{otherwise}.
     \end{cases}
\end{multline}
Since the $\paren*{\ell_{0-1},\sH_{\mathrm{NN}}}$-minimizability gap coincides with
the $\paren*{\ell_{0-1},\sH_{\mathrm{NN}}}$-approximation error and
$\paren*{\Phi_{\mathrm{log}},\sH_{\mathrm{NN}}}$-minimizability gap coincides with
the $\paren*{\Phi_{\mathrm{log}},\sH_{\mathrm{NN}}}$-approximation error for $\Lambda B =\plus\infty$,
the inequality can be rewritten as follows:
\begin{multline*}
     \sR_{\ell_{0-1}}(h)- \sR_{\ell_{0-1},\sH_{\mathrm{all}}}^*\\
     \leq 
     \begin{cases}
      \sqrt{2}\,\bracket*{\sR_{\Phi_{\mathrm{exp}}}(h) - \sR_{\Phi_{\mathrm{exp}},\sH_{\mathrm{all}}}^*}^{\frac12} & \text{if } B = \plus\infty, \\
     \begin{cases}
    \sqrt{2}\,\bracket*{\sR_{\Phi_{\mathrm{exp}}}(h)- \sR_{\Phi_{\mathrm{exp}},\sH_{\mathrm{NN}}}^*+\sM_{\Phi_{\mathrm{exp}},\sH_{\mathrm{NN}}}}^{\frac12} & \text{if } \sR_{\Phi_{\mathrm{exp}}}(h)- \sR_{\Phi_{\mathrm{exp}},\sH_{\mathrm{NN}}}^*\leq \frac{1}{2}\paren*{\frac{e^{2\Lambda B}-1}{e^{2\Lambda B}+1}}^2-\sM_{\Phi_{\mathrm{exp}},\sH_{\mathrm{NN}}}, \\
    2\paren*{\frac{e^{2\Lambda B}+1}{e^{2\Lambda B}-1}}\paren*{\sR_{\Phi_{\mathrm{exp}}}(h)- \sR_{\Phi_{\mathrm{exp}},\sH_{\mathrm{NN}}}^*+\sM_{\Phi_{\mathrm{exp}},\sH_{\mathrm{NN}}}} & \text{otherwise}.
    \end{cases} & \text{otherwise}.
     \end{cases}
\end{multline*}
where the $\paren*{\Phi_{\mathrm{exp}},\sH_{\mathrm{NN}}}$-minimizability gap $\sM_{\Phi_{\mathrm{exp}},\sH_{\mathrm{NN}}}$ is characterized as below, which is less than
the $\paren*{\Phi_{\mathrm{exp}},\sH_{\mathrm{NN}}}$-approximation error when $\Lambda B<\plus \infty$:
\begin{align*}
\sM_{\Phi_{\mathrm{exp}},\sH_{\mathrm{NN}}}
& = \sR_{\Phi_{\mathrm{exp}},\sH_{\mathrm{NN}}}^*-
\mathbb{E}_{X}\bracket*{2\sqrt{\eta(x)(1-\eta(x))}\mathds{1}_{1/2\log_2\abs*{\frac{\eta(x)}{1-\eta(x)}}\leq \Lambda W\norm*{x}_p+\Lambda B}}\\
& - \mathbb{E}_{X}\bracket*{\max\curl*{\eta(x),1-\eta(x)}e^{-(\Lambda W\norm*{x}_p+\Lambda B)}\mathds{1}_{1/2\log_2\abs*{\frac{\eta(x)}{1-\eta(x)}}> \Lambda W\norm*{x}_p+\Lambda B}}\\
&-\bracket*{\min\curl*{\eta(x),1-\eta(x)}e^{\Lambda W\norm*{x}_p+\Lambda B}\mathds{1}_{1/2\log_2\abs*{\frac{\eta(x)}{1-\eta(x)}}> \Lambda W\norm*{x}_p+\Lambda B}}\\
& < \sR_{\Phi_{\mathrm{exp}},\sH_{\mathrm{NN}}}^* - \mathbb{E}_{X}\bracket*{2\sqrt{\eta(x)(1-\eta(x))}}\\
& = \sR_{\Phi_{\mathrm{exp}},\sH_{\mathrm{NN}}}^* - \sR_{\Phi_{\mathrm{exp}},\sH_{\mathrm{all}}}^* .
\end{align*}
Therefore, the inequality for $\Lambda B = \plus \infty$ coincides with the consistency excess error bound
known for the exponential loss \citep{Zhang2003,MohriRostamizadehTalwalkar2018} but the one for $\Lambda B< \plus \infty$ is distinct and novel.

\subsubsection{Quadratic Loss}
For the quadratic loss $\Phi_{\mathrm{quad}}(\alpha)\colon=(1-\alpha)^2\mathds{1}_{\alpha\leq 1}$, for all $h\in \sH_{\mathrm{NN}}$ and $x\in \sX$:
\begin{equation*}
\begin{aligned}
\sC_{\Phi_{\mathrm{quad}}}(h,x,t)
&=t \Phi_{\mathrm{quad}}(h(x))+(1-t)\Phi_{\mathrm{quad}}(-h(x))\\
&=t\paren*{1-h(x)}^2\mathds{1}_{h(x)\leq 1}+(1-t)\paren*{1+h(x)}^2\mathds{1}_{h(x)\geq -1}.\\
\inf_{h\in\sH_{\mathrm{NN}}}\sC_{\Phi_{\mathrm{quad}}}(h,x,t)
&=\begin{cases}
4t(1-t), & \abs*{2t-1}\leq\Lambda W\norm*{x}_p+\Lambda B,\\
\max\curl*{t,1-t}\paren*{1-\paren*{\Lambda W\norm*{x}_p+\Lambda B}}^2
+\min\curl*{t,1-t}\paren*{1+\Lambda W\norm*{x}_p+\Lambda B}^2, &\abs*{2t-1}> \Lambda W\norm*{x}_p+\Lambda B.
\end{cases}
\end{aligned}
\end{equation*}
Therefore, the $\paren*{\Phi_{\mathrm{quad}},\sH_{\mathrm{NN}}}$-minimizability gap can be expressed as follows:
\begin{equation}
\begin{aligned}
\label{eq:M-quad-NN}
\sM_{\Phi_{\mathrm{quad}},\sH_{\mathrm{NN}}}
& = \sR_{\Phi_{\mathrm{quad}},\sH_{\mathrm{NN}}}^*-\mathbb{E}_{X}\bracket*{4\eta(x)(1-\eta(x))\mathds{1}_{\abs*{2\eta(x)-1}\leq \Lambda W\norm*{x}_p+\Lambda B}}\\
& - \mathbb{E}_{X}\bracket*{\max\curl*{\eta(x),1-\eta(x)}\paren*{1-\paren*{\Lambda W\norm*{x}_p+\Lambda B}}^2\mathds{1}_{\abs*{2\eta(x)-1}> \Lambda W\norm*{x}_p+\Lambda B}}\\
& - \mathbb{E}_{X}\bracket*{\min\curl*{\eta(x),1-\eta(x)}\paren*{1+\paren*{\Lambda W\norm*{x}_p+\Lambda B}}^2\mathds{1}_{\abs*{2\eta(x)-1}> \Lambda W\norm*{x}_p+\Lambda B}}
\end{aligned}
\end{equation}
Note $\paren*{\Phi_{\mathrm{quad}},\sH_{\mathrm{NN}}}$-minimizability gap coincides with
the $\paren*{\Phi_{\mathrm{quad}},\sH_{\mathrm{NN}}}$-approximation error $\sR_{\Phi_{\mathrm{quad}},\sH_{\mathrm{NN}}}^*-
\mathbb{E}_{X}\bracket*{4\eta(x)(1-\eta(x))}$ for $\Lambda B \geq 1$.

For $\frac{1}2< t\leq1$, we have
\small
\begin{align*}
\inf_{h\in\sH_{\mathrm{NN}}:h(x)<0}\sC_{\Phi_{\mathrm{quad}}}(h,x,t)
&=t+(1-t)\\
&=1\\
\inf_{x\in \sX}\inf_{h\in\sH_{\mathrm{NN}}:h(x)<0}\Delta\sC_{\Phi_{\mathrm{quad}},\sH_{\mathrm{NN}}}(h,x,t)
& =\inf_{x\in \sX}\paren*{\inf_{h\in\sH_{\mathrm{NN}}:h(x)<0}\sC_{\Phi_{\mathrm{quad}}}(h,x,t)-\inf_{h\in\sH_{\mathrm{NN}}}\sC_{\Phi_{\mathrm{quad}}}(h,x,t)}\\
&=\inf_{x\in \sX}\begin{cases}
1-4t(1-t),& 2t-1\leq \Lambda W\norm*{x}_p+\Lambda B,\\
1-t\paren*{1-\paren*{\Lambda W\norm*{x}_p+\Lambda B}}^2-(1-t)\paren*{1+\Lambda W\norm*{x}_p+\Lambda B}^2, & 2t-1> \Lambda W\norm*{x}_p+\Lambda B.
\end{cases}\\
&=\begin{cases}
1-4t(1-t),& 2t-1\leq \Lambda B,\\
1-t\paren*{1-\Lambda B}^2-(1-t)\paren*{1+\Lambda B}^2,& 2t-1> \Lambda B.
\end{cases}\\
&=\sT(2t-1)
\end{align*}
\normalsize
where $\sT$ is the increasing and convex function on $[0,1]$ defined by
\begin{align*}
\forall t\in[0,1], \quad
\sT(t)=\begin{cases}
t^2, & t\leq \Lambda B,\\
2\Lambda B \,t-(\Lambda B)^2, & t> \Lambda B.
\end{cases}
\end{align*}
By Definition~\ref{def:trans}, for any $\epsilon\geq 0$, the $\sH_{\mathrm{NN}}$-estimation error transformation of the quadratic loss is as follows:
\begin{align*}
\sT_{\Phi_{\mathrm{quad}}}= 
\begin{cases}
\sT(t), & t\in \left[\epsilon,1\right], \\
\frac{\sT(\epsilon)}{\epsilon}\, t, &  t\in \left[0,\epsilon\right).
\end{cases}
\end{align*}
Therefore, when $\epsilon=0$, $\sT_{\Phi_{\mathrm{quad}}}$ is convex, non-decreasing, invertible and satisfies that $\sT_{\Phi_{\mathrm{quad}}}(0)=0$. By Theorem~\ref{Thm:tightness}, we can choose $\Psi(t)=\sT_{\Phi_{\mathrm{quad}}}(t)$ in Theorem~\ref{Thm:excess_bounds_Psi_uniform}, or equivalently $\Gamma(t) = \sT_{\Phi_{\mathrm{quad}}}^{-1}(t)=
\begin{cases}
\sqrt{t}, & t \leq (\Lambda B)^2 \\
\frac{t}{2\Lambda B}+\frac{\Lambda B}{2}, & t > (\Lambda B)^2
\end{cases}$, in Theorem~\ref{Thm:excess_bounds_Gamma_uniform}, which are optimal. Thus, by Theorem~\ref{Thm:excess_bounds_Psi_uniform} or Theorem~\ref{Thm:excess_bounds_Gamma_uniform}, setting $\e = 0$ yields the $\sH_{\mathrm{NN}}$-consistency estimation error bound for the quadratic loss, valid for all $h \in \sH_{\mathrm{NN}}$:
\begin{multline}
\label{eq:quad-NN-est}
    \sR_{\ell_{0-1}}(h)- \sR_{\ell_{0-1},\sH_{\mathrm{NN}}}^*+\sM_{\ell_{0-1},\sH_{\mathrm{NN}}}\\
    \leq
    \begin{cases}
    \bracket*{\sR_{\Phi_{\mathrm{quad}}}(h)- \sR_{\Phi_{\mathrm{quad}},\sH_{\mathrm{NN}}}^*+\sM_{\Phi_{\mathrm{quad}},\sH_{\mathrm{NN}}}}^{\frac12}  & \text{if } \sR_{\Phi_{\mathrm{quad}}}(h)- \sR_{\Phi_{\mathrm{quad}},\sH_{\mathrm{NN}}}^*\leq (\Lambda B)^2-\sM_{\Phi_{\mathrm{quad}},\sH_{\mathrm{NN}}} \\
    \frac{\sR_{\Phi_{\mathrm{quad}}}(h)- \sR_{\Phi_{\mathrm{quad}},\sH_{\mathrm{NN}}}^*+\sM_{\Phi_{\mathrm{quad}},\sH_{\mathrm{NN}}}}{2\Lambda B}+\frac{\Lambda B}{2} & \text{otherwise}
    \end{cases}
\end{multline}
Since the $\paren*{\ell_{0-1},\sH_{\mathrm{NN}}}$-minimizability gap coincides with
the $\paren*{\ell_{0-1},\sH_{\mathrm{NN}}}$-approximation error and
$\paren*{\Phi_{\mathrm{quad}},\sH_{\mathrm{NN}}}$-minimizability gap coincides with
the $\paren*{\Phi_{\mathrm{quad}},\sH_{\mathrm{NN}}}$-approximation error for $\Lambda B \geq 1$, the inequality can be rewritten as follows:
\begin{multline*}
     \sR_{\ell_{0-1}}(h)- \sR_{\ell_{0-1},\sH_{\mathrm{all}}}^*\\
     \leq 
     \begin{cases}
     \bracket*{\sR_{\Phi_{\mathrm{quad}}}(h) - \sR_{\Phi_{\mathrm{quad}},\sH_{\mathrm{all}}}^*}^{\frac12} & \text{if } \Lambda B \geq 1\\
     \begin{cases}
    \bracket*{\sR_{\Phi_{\mathrm{quad}}}(h)- \sR_{\Phi_{\mathrm{quad}},\sH_{\mathrm{NN}}}^*+\sM_{\Phi_{\mathrm{quad}},\sH_{\mathrm{NN}}}}^{\frac12}  & \text{if } \sR_{\Phi_{\mathrm{quad}}}(h)- \sR_{\Phi_{\mathrm{quad}},\sH_{\mathrm{NN}}}^*\leq (\Lambda B)^2-\sM_{\Phi_{\mathrm{quad}},\sH_{\mathrm{NN}}} \\
    \frac{\sR_{\Phi_{\mathrm{quad}}}(h)- \sR_{\Phi_{\mathrm{quad}},\sH_{\mathrm{NN}}}^*+\sM_{\Phi_{\mathrm{quad}},\sH_{\mathrm{NN}}}}{2\Lambda B}+\frac{\Lambda B}{2} & \text{otherwise}
    \end{cases} & \text{otherwise}
     \end{cases}
\end{multline*}
where the $\paren*{\Phi_{\mathrm{quad}},\sH_{\mathrm{NN}}}$-minimizability gap $\sM_{\Phi_{\mathrm{quad}},\sH_{\mathrm{NN}}}$ is characterized as below,  which is less than
the $\paren*{\Phi_{\mathrm{quad}},\sH_{\mathrm{NN}}}$-approximation error when $\Lambda B<1$:
\begin{align*}
\sM_{\Phi_{\mathrm{quad}},\sH_{\mathrm{NN}}}
& = \sR_{\Phi_{\mathrm{quad}},\sH_{\mathrm{NN}}}^*-\mathbb{E}_{X}\bracket*{4\eta(x)(1-\eta(x))\mathds{1}_{\abs*{2\eta(x)-1}\leq \Lambda W\norm*{x}_p+\Lambda B}}\\
& - \mathbb{E}_{X}\bracket*{\max\curl*{\eta(x),1-\eta(x)}\paren*{1-\paren*{\Lambda W\norm*{x}_p+\Lambda B}}^2\mathds{1}_{\abs*{2\eta(x)-1}> \Lambda W\norm*{x}_p+\Lambda B}}\\
& - \mathbb{E}_{X}\bracket*{\min\curl*{\eta(x),1-\eta(x)}\paren*{1+\paren*{\Lambda W\norm*{x}_p+\Lambda B}}^2\mathds{1}_{\abs*{2\eta(x)-1}> \Lambda W\norm*{x}_p+\Lambda B}}\\
& < \sR_{\Phi_{\mathrm{quad}},\sH_{\mathrm{NN}}}^* - \mathbb{E}_{X}\bracket*{4\eta(x)(1-\eta(x))}\\
& = \sR_{\Phi_{\mathrm{quad}},\sH_{\mathrm{NN}}}^* - \sR_{\Phi_{\mathrm{quad}},\sH_{\mathrm{all}}}^* .
\end{align*}
Therefore, the inequality for $\Lambda B \geq 1$ coincides with the consistency excess error bound
known for the quadratic loss \citep{Zhang2003,bartlett2006convexity} but the one for $\Lambda B< 1$ is distinct and novel.

\subsubsection{Sigmoid Loss}
For the sigmoid loss $\Phi_{\mathrm{sig}}(\alpha)\colon=1-\tanh(k\alpha),~k>0$,
for all $h\in \sH_{\mathrm{NN}}$ and $x\in \sX$:
\begin{equation*}
\begin{aligned}
\sC_{\Phi_{\mathrm{sig}}}(h,x,t)
&=t \Phi_{\mathrm{sig}}(h(x))+(1-t)\Phi_{\mathrm{sig}}(-h(x)),\\
&=t\paren*{1-\tanh(kh(x))}+(1-t)\paren*{1+\tanh(kh(x))}.\\
\inf_{h\in\sH_{\mathrm{NN}}}\sC_{\Phi_{\mathrm{sig}}}(h,x,t)
&=1-\abs*{1-2t}\tanh\paren*{k\paren*{\Lambda W\norm*{x}_p+\Lambda B}}
\end{aligned}
\end{equation*}
Therefore, the $\paren*{\Phi_{\mathrm{sig}},\sH_{\mathrm{NN}}}$-minimizability gap can be expressed as follows:
\begin{equation}
\begin{aligned}
\label{eq:M-sig-NN}
\sM_{\Phi_{\mathrm{sig}},\sH_{\mathrm{NN}}}
&= \sR_{\Phi_{\mathrm{sig}},\sH_{\mathrm{NN}}}^*-\mathbb{E}_{X}\bracket*{\inf_{h\in\sH_{\mathrm{NN}}}\sC_{\Phi_{\mathrm{sig}}}(h,x,\eta(x))}\\
&= \sR_{\Phi_{\mathrm{sig}},\sH_{\mathrm{NN}}}^*-\mathbb{E}_{X}\bracket*{1-\abs*{1-2\eta(x)}\tanh\paren*{k\paren*{\Lambda W\norm*{x}_p+\Lambda B}}}.
\end{aligned}
\end{equation}
Note $\paren*{\Phi_{\mathrm{sig}},\sH_{\mathrm{NN}}}$-minimizability gap coincides with
the $\paren*{\Phi_{\mathrm{sig}},\sH_{\mathrm{NN}}}$-approximation error $\sR_{\Phi_{\mathrm{sig}},\sH_{\mathrm{NN}}}^*-\mathbb{E}_{X}\bracket*{1-\abs*{1-2\eta(x)}}$ for $\Lambda B = \plus \infty$.

For $\frac{1}2< t\leq1$, we have
\begin{align*}
\inf_{h\in\sH_{\mathrm{NN}}:h(x)<0}\sC_{\Phi_{\mathrm{sig}}}(h,x,t)
&=1-\abs*{1-2t}\tanh(0)\\
&=1.\\
\inf_{x\in \sX}\inf_{h\in\sH_{\mathrm{NN}}:h(x)<0}\Delta\sC_{\Phi_{\mathrm{sig}},\sH_{\mathrm{NN}}}(h,x,t)
&=\inf_{x\in \sX}\paren*{\inf_{h\in\sH_{\mathrm{NN}}:h(x)<0}\sC_{\Phi_{\mathrm{sig}}}(h,x,t)-\inf_{h\in\sH_{\mathrm{NN}}}\sC_{\Phi_{\mathrm{sig}}}(h,x,t)}\\
&=\inf_{x\in \sX}(2t-1)\tanh\paren*{k\paren*{\Lambda W\norm*{x}_p+\Lambda B}}\\
&=(2t-1)\tanh(k\Lambda B)\\
&=\sT(2t-1)
\end{align*}
where $\sT$ is the increasing and convex function on $[0,1]$ defined by
\begin{align*}
\forall t\in[0,1],\; \sT(t)=\tanh(k\Lambda B) \, t .
\end{align*}
By Definition~\ref{def:trans}, for any $\epsilon\geq 0$, the $\sH_{\mathrm{NN}}$-estimation error transformation of the sigmoid loss is as follows:
\begin{align*}
\sT_{\Phi_{\mathrm{sig}}}= \tanh(k\Lambda B) \, t, \quad t \in [0,1],
\end{align*}
Therefore, $\sT_{\Phi_{\mathrm{sig}}}$ is convex, non-decreasing, invertible and satisfies that $\sT_{\Phi_{\mathrm{sig}}}(0)=0$. By Theorem~\ref{Thm:tightness}, we can choose $\Psi(t)=\tanh(k\Lambda B)\,t$ in Theorem~\ref{Thm:excess_bounds_Psi_uniform}, or equivalently $\Gamma(t)=\frac{t}{\tanh(k\Lambda B)}$ in Theorem~\ref{Thm:excess_bounds_Gamma_uniform}, which are optimal when $\e=0$.
Thus, by Theorem~\ref{Thm:excess_bounds_Psi_uniform} or Theorem~\ref{Thm:excess_bounds_Gamma_uniform},  setting $\e = 0$ yields the $\sH_{\mathrm{NN}}$-consistency estimation error bound for the sigmoid loss, valid for all $h \in \sH_{\mathrm{NN}}$:
\begin{align}
\label{eq:sig-NN-est}
     \sR_{\ell_{0-1}}(h)- \sR_{\ell_{0-1},\sH_{\mathrm{NN}}}^*\leq \frac{\sR_{\Phi_{\mathrm{sig}}}(h)- \sR_{\Phi_{\mathrm{sig}},\sH_{\mathrm{NN}}}^*+\sM_{\Phi_{\mathrm{sig}},\sH_{\mathrm{NN}}}}{\tanh(k\Lambda B)}-\sM_{\ell_{0-1},\sH_{\mathrm{NN}}}.
\end{align}
Since the $\paren*{\ell_{0-1},\sH_{\mathrm{NN}}}$-minimizability gap coincides with
the $\paren*{\ell_{0-1},\sH_{\mathrm{NN}}}$-approximation error, and since
$\paren*{\Phi_{\mathrm{sig}},\sH_{\mathrm{NN}}}$-minimizability gap coincides with
the $\paren*{\Phi_{\mathrm{sig}},\sH_{\mathrm{NN}}}$-approximation error for $\Lambda B = \plus \infty$,
the inequality can be rewritten as follows:
\begin{align*}
     \sR_{\ell_{0-1}}(h)- \sR_{\ell_{0-1},\sH_{\mathrm{all}}}^*
      \leq 
     \begin{cases}
     \sR_{\Phi_{\mathrm{sig}}}(h) - \sR_{\Phi_{\mathrm{sig}},\sH_{\mathrm{all}}}^* & \text{if } \Lambda B = \plus \infty\\
     \frac{1}{\tanh(k\Lambda B)} \bracket[\Big]{\sR_{\Phi_{\mathrm{sig}}}(h)
     - \mathbb{E}_{X}\bracket*{1-\abs*{1-2\eta(x)}\tanh\paren*{k\paren*{\Lambda W\norm*{x}_p+\Lambda B}}\ } } & \text{otherwise}.
     \end{cases}
\end{align*}
The inequality for $\Lambda B = \plus \infty$ coincides with the consistency excess error bound
known for the sigmoid loss \citep{Zhang2003,bartlett2006convexity,MohriRostamizadehTalwalkar2018} but the one for $\Lambda B < \plus \infty$ is distinct and novel. 
For $\Lambda B<\plus \infty$, we have
\begin{align*}
\mathbb{E}_{X}\bracket*{1-\abs*{1-2\eta(x)}\tanh\paren*{k\paren*{\Lambda W\norm*{x}_p+\Lambda B}}}> \mathbb{E}_{X}\bracket*{1-\abs*{2\eta(x)-1}}= 2\mathbb{E}_X\bracket*{\min\curl*{\eta(x), 1 - \eta(x)}}= \sR_{\Phi_{\mathrm{hinge}},\sH_{\mathrm{all}}}^*.
\end{align*}
Therefore for $\Lambda B<\plus \infty$,
\begin{align*}
\sR_{\Phi_{\mathrm{sig}}}(h) - \mathbb{E}_{X}\bracket*{1-\abs*{1-2\eta(x)}\tanh\paren*{k\paren*{\Lambda W\norm*{x}_p+\Lambda B}}} < \sR_{\Phi_{\mathrm{sig}}}(h) - \sR_{\Phi_{\mathrm{sig}},\sH_{\mathrm{all}}}^*.
\end{align*}
Note that: $\sR_{\Phi_{\mathrm{sig}},\sH_{\mathrm{all}}}^* = 2 \sR_{\ell_{0-1},\sH_{\mathrm{all}}}^* =2\mathbb{E}_X\bracket*{\min\curl*{\eta(x), 1 - \eta(x)}}$. Thus, the first
inequality (case $\Lambda B = \plus \infty$) can be equivalently written as follows:
\begin{align*}
    \forall h \in \sH_{\mathrm{NN}},\; \sR_{\ell_{0-1}}(h) 
     \leq \sR_{\Phi_{\mathrm{sig}}}(h) - \mathbb{E}_X\bracket*{\min\curl*{\eta(x), 1 - \eta(x)}},
\end{align*}
which is a more informative upper bound than the standard
inequality $\sR_{\ell_{0-1}}(h) 
     \leq \sR_{\Phi_{\mathrm{sig}}}(h)$.

\subsubsection{\texorpdfstring{$\rho$}{rho}-Margin Loss}
For the $\rho$-margin loss $\Phi_{\rho}(\alpha)\colon=\min\curl*{1,\max\curl*{0,1-\frac{\alpha}{\rho}}},~\rho>0$,
for all $h\in \sH_{\mathrm{NN}}$ and $x\in \sX$:
\begin{equation*}
\begin{aligned}
\sC_{\Phi_{\rho}}(h,x,t)
&=t \Phi_{\rho}(h(x))+(1-t)\Phi_{\rho}(-h(x)),\\
&=t\min\curl*{1,\max\curl*{0,1-\frac{h(x)}{\rho}}}+(1-t)\min\curl*{1,\max\curl*{0,1+\frac{h(x)}{\rho}}}.\\
\inf_{h\in\sH_{\mathrm{NN}}}\sC_{\Phi_{\rho}}(h,x,t)
&=\min\curl*{t,1-t}+\max\curl*{t,1-t}\paren*{1-\frac{\min\curl*{\Lambda W\norm*{x}_p+\Lambda B,\rho}}{\rho}}.
\end{aligned}
\end{equation*}
Therefore, the $\paren*{\Phi_{\rho},\sH_{\mathrm{NN}}}$-minimizability gap can be expressed as follows:
\begin{equation}
\label{eq:M-rho-NN}
\begin{aligned}
\sM_{\Phi_{\rho},\sH_{\mathrm{NN}}}
& = \sR_{\Phi_{\rho},\sH_{\mathrm{NN}}}^*-\mathbb{E}_{X}\bracket*{\inf_{h\in\sH_{\mathrm{NN}}}\sC_{\Phi_{\rho}}(h,x,\eta(x))}\\
& = \sR_{\Phi_{\rho},\sH_{\mathrm{NN}}}^*-\mathbb{E}_{X}\bracket*{\min\curl*{\eta(x),1-\eta(x)}+\max\curl*{\eta(x),1-\eta(x)}\paren*{1-\frac{\min\curl*{\Lambda W\norm*{x}_p+\Lambda B,\rho}}{\rho}}}.
 \end{aligned}
\end{equation}
Note the $\paren*{\Phi_{\rho},\sH_{\mathrm{NN}}}$-minimizability gap coincides with
the $\paren*{\Phi_{\rho},\sH_{\mathrm{NN}}}$-approximation error $\sR_{\Phi_{\rho},\sH_{\mathrm{NN}}}^*-\mathbb{E}_{X}\bracket*{\min\curl*{\eta(x),1-\eta(x)}}$ for $\Lambda B \geq \rho$.

For $\frac{1}2< t\leq1$, we have
\begin{align*}
\inf_{h\in\sH_{\mathrm{NN}}:h(x)<0}\sC_{\Phi_{\rho}}(h,x,t)
&=t+(1-t)\paren*{1-\frac{\min\curl*{\Lambda W\norm*{x}_p+\Lambda B,\rho}}{
\rho}}.\\
\inf_{x\in \sX}\inf_{h\in\sH_{\mathrm{NN}}:h(x)<0}\Delta\sC_{\Phi_{\rho},\sH_{\mathrm{NN}}}(h,x)
&=\inf_{x\in \sX}\paren*{\inf_{h\in\sH_{\mathrm{NN}}:h(x)<0}\sC_{\Phi_{\rho}}(h,x,t)-\inf_{h\in\sH_{\mathrm{NN}}}\sC_{\Phi_{\rho}}(h,x,t)}\\
&=\inf_{x\in \sX}(2t-1)\frac{\min\curl*{\Lambda W\norm*{x}_p+\Lambda B,\rho}}{\rho}\\
&=(2t-1)\frac{\min\curl*{\Lambda B,\rho}}{\rho}\\
&=\sT(2t-1)
\end{align*}
where $\sT$ is the increasing and convex function on $[0,1]$ defined by
\begin{align*}
\forall t\in [0,1],\; \sT(t)=\frac{\min\curl*{\Lambda B,\rho}}{\rho} \, t.    
\end{align*} 
By Definition~\ref{def:trans}, for any $\epsilon\geq 0$, the $\sH_{\mathrm{NN}}$-estimation error transformation of the $\rho$-margin loss is as follows:
\begin{align*}
\sT_{\Phi_{\rho}}= \frac{\min\curl*{\Lambda B,\rho}}{\rho} \, t, \quad t \in [0,1],
\end{align*}
Therefore, $\sT_{\Phi_{\rho}}$ is convex, non-decreasing, invertible and satisfies that $\sT_{\Phi_{\rho}}(0)=0$. By Theorem~\ref{Thm:tightness}, we can choose $\Psi(t)=\frac{\min\curl*{\Lambda B,\rho}}{\rho} \, t$ in Theorem~\ref{Thm:excess_bounds_Psi_uniform}, or equivalently $\Gamma(t)=\frac{\rho }{\min\curl*{\Lambda B,\rho}} \, t$ in Theorem~\ref{Thm:excess_bounds_Gamma_uniform}, which are optimal when $\e=0$.
Thus, by Theorem~\ref{Thm:excess_bounds_Psi_uniform} or Theorem~\ref{Thm:excess_bounds_Gamma_uniform}, setting $\e = 0$ yields the $\sH_{\mathrm{NN}}$-consistency estimation error bound for the $\rho$-margin loss, valid for all $h \in \sH_{\mathrm{NN}}$:
\begin{align}
\label{eq:rho-NN-est}
     \sR_{\ell_{0-1}}(h)- \sR_{\ell_{0-1},\sH_{\mathrm{NN}}}^*\leq \frac{\rho\paren*{\sR_{\Phi_{\rho}}(h)- \sR_{\Phi_{\rho},\sH_{\mathrm{NN}}}^*+\sM_{\Phi_{\rho},\sH_{\mathrm{NN}}}}}{\min\curl*{\Lambda B,\rho}}-\sM_{\ell_{0-1},\sH_{\mathrm{NN}}}.
\end{align}
Since the $\paren*{\ell_{0-1},\sH_{\mathrm{NN}}}$-minimizability gap coincides with
the $\paren*{\ell_{0-1},\sH_{\mathrm{NN}}}$-approximation error and
$\paren*{\Phi_{\rho},\sH_{\mathrm{NN}}}$-minimizability gap coincides with
the $\paren*{\Phi_{\rho},\sH_{\mathrm{NN}}}$-approximation error for $\Lambda B \geq \rho$,
the inequality can be rewritten as follows:
\begin{align*}
     \sR_{\ell_{0-1}}(h)- \sR_{\ell_{0-1},\sH_{\mathrm{all}}}^*
      \leq 
     \begin{cases}
     \sR_{\Phi_{\rho}}(h) - \sR_{\Phi_{\rho},\sH_{\mathrm{all}}}^* & \text{if } \Lambda B \geq \rho\\
     \frac{\rho \paren*{\sR_{\Phi_{\rho}}(h)
     - \mathbb{E}_{X}\bracket*{\min\curl*{\eta(x),1-\eta(x)}+\max\curl*{\eta(x),1-\eta(x)}\paren*{1-\frac{\min\curl*{\Lambda W\norm*{x}_p+\Lambda B,\rho}}{\rho}} } }}{\Lambda B}  & \text{otherwise}. 
     \end{cases}
\end{align*}
Note that: $\sR_{\Phi_{\rho},\sH_{\mathrm{all}}}^* =  \sR_{\ell_{0-1},\sH_{\mathrm{all}}}^* =\mathbb{E}_X\bracket*{\min\curl*{\eta(x), 1 - \eta(x)}}$. Thus, the first
inequality (case $\Lambda B \geq \rho$) can be equivalently written as follows:
\begin{align*}
    \forall h \in \sH_{\mathrm{NN}}, \quad \sR_{\ell_{0-1}}(h) 
     \leq \sR_{\Phi_{\rho}}(h).
\end{align*}
The case $\Lambda B \geq \rho$ is one of the ``trivial cases'' mentioned in Section~\ref{sec:general}, where the trivial inequality $\sR_{\ell_{0-1}}(h) \leq \sR_{\Phi_{\rho}}(h)$ can be obtained directly using the fact that $\ell_{0-1}$ is upper bounded by $\Phi_{\rho}$. This, however, does not imply that non-adversarial $\sH_{\mathrm{NN}}$-consistency estimation error bound for the $\rho$-margin loss is trivial when $\Lambda B>\rho$ since it is optimal.

\section{Derivation of Adversarial \texorpdfstring{$\sH$}{H}-Estimation Error Bounds}
\label{app:derivation-adv}
\subsection{Linear Hypotheses}
\label{app:derivation-lin-adv}
By the definition of $\sH_{\mathrm{lin}}$, for any $x \in \sX$, 
\begin{align*}
&\uv h_\gamma(x)
  =w \cdot x-\gamma \|w\|_q+b \\
& \in 
  \begin{cases}
\bracket*{-W \norm*{x}_p-\gamma W-B, W\norm*{x}_p - \gamma W+ B}
&  \norm*{x}_p \geq \gamma\\
\bracket*{-W \norm*{x}_p-\gamma W-B, B}
&  \norm*{x}_p < \gamma
\end{cases},\\
&\ov h_\gamma(x)=w \cdot x+\gamma \|w\|_q+b \\
& \in 
\begin{cases}
\bracket*{-W \norm*{x}_p+\gamma W-B, W\norm*{x}_p + \gamma W+ B}
&  \norm*{x}_p \geq \gamma\\
\bracket*{-B, W\norm*{x}_p + \gamma W+ B}
&  \norm*{x}_p < \gamma
\end{cases}.
\end{align*}
Note $\sH_{\mathrm{lin}}$ is symmetric. For any $x\in \sX$, there exist $w=0$ and any $0<b\leq B$ such that $w \cdot x-\gamma \|w\|_q+b>0$. Thus by Lemma~\ref{lemma:explicit_assumption_01_adv}, for any $x\in \sX$, $\sC^*_{\ell_{\gamma},\sH_{\mathrm{lin}}}(x) =\min\curl*{\eta(x), 1 - \eta(x)}$. The $\paren*{\ell_{\gamma},\sH_{\mathrm{lin}}}$-minimizability gap can be expressed as follows:
\begin{align}
\label{eq:M-01-lin-adv}
\sM_{\ell_{\gamma},\sH_{\mathrm{lin}}}
& = \sR_{\ell_{\gamma},\sH_{\mathrm{lin}}}^*-\mathbb{E}_{X}\bracket*{\min\curl*{\eta(x),1-\eta(x)}}.
\end{align}

\subsubsection{Supremum-Based \texorpdfstring{$\rho$}{rho}-Margin Loss}
\label{app:rho-lin-adv}
For the supremum-based $\rho$-margin loss 
\begin{align*}
\wt{\Phi}_{\rho}\colon=\sup_{x'\colon \|x-x'\|_p\leq \gamma}\Phi_{\rho}(y h(x')),  \quad \text{where } \Phi_{\rho}(\alpha)=\min\curl*{1,\max\curl*{0,1-\frac{\alpha}{\rho}}},~\rho>0,   
\end{align*}
for all $h\in \sH_{\mathrm{lin}}$ and $x\in \sX$:
\begin{equation*}
\begin{aligned}
\sC_{\wt{\Phi}_{\rho}}(h,x,t) 
&=t \wt{\Phi}_{\rho}(h(x))+(1-t)\wt{\Phi}_{\rho}(-h(x))\\
& = t\Phi_{\rho}\paren*{\uv h_\gamma(x)}+(1-t)\Phi_{\rho}\paren*{-\ov h_\gamma(x)}\\
& =t\min\curl*{1,\max\curl*{0,1-\frac{\uv h_\gamma(x)}{\rho}}}+(1-t)\min\curl*{1,\max\curl*{0,1+\frac{\ov h_\gamma(x)}{\rho}}}.\\
\inf_{h\in\sH_{\mathrm{lin}}}\sC_{\wt{\Phi}_{\rho}}(h,x,t)
& =\max\curl*{t,1-t}\paren*{1-\frac{\min\curl*{W\max \curl*{\norm*{x}_p,\gamma}-\gamma W+B,\rho}}{\rho}} + \min\curl*{t,1-t}.
\end{aligned}
\end{equation*}
Therefore, the $\paren*{\wt{\Phi}_{\rho},\sH_{\mathrm{lin}}}$-minimizability gap can be expressed as follows:
\begin{equation}
\begin{aligned}
\label{eq:M-rho-lin-adv}
\sM_{\wt{\Phi}_{\rho},\sH_{\mathrm{lin}}} 
& = \sR_{\wt{\Phi}_{\rho},\sH_{\mathrm{lin}}}^*- \mathbb{E}_{X}\bracket*{\inf_{h\in\sH_{\mathrm{lin}}}\sC_{\wt{\Phi}_{\rho}}(h,x,\eta(x))}\\
& = \sR_{\wt{\Phi}_{\rho},\sH_{\mathrm{lin}}}^*- \mathbb{E}_{X}\bracket*{\max\curl*{\eta(x),1-\eta(x)}\paren*{1-\frac{\min\curl*{W\max\curl*{\norm*{x}_p,\gamma}-\gamma W+B,\rho}}{\rho}}}\\
& -\mathbb{E}_{X}\bracket*{\min\curl*{\eta(x),1-\eta(x)}}.    
\end{aligned}
\end{equation}
For $\frac{1}2< t\leq1$, we have
\begin{align*}
\inf_{h\in\sH_{\mathrm{lin}}:\uv h_\gamma(x)\leq 0 \leq \ov h_\gamma(x)}\sC_{\wt{\Phi}_{\rho}}(h,x,t)
& = t+(1-t)\\
& =1\\
\inf_{x\in \sX} \inf_{h\in\sH_{\mathrm{lin}}\colon \uv h_\gamma(x)\leq 0 \leq \ov h_\gamma(x)}\Delta\sC_{\wt{\Phi}_{\rho},\sH_{\mathrm{lin}}}(h,x,t)
& = \inf_{x\in \sX} \curl*{\inf_{h\in\sH_{\mathrm{lin}}:\uv h_\gamma(x)\leq 0 \leq \ov h_\gamma(x)}\sC_{\wt{\Phi}_{\rho}}(h,x,t)-\inf_{h\in\sH_{\mathrm{lin}}}\sC_{\wt{\Phi}_{\rho}}(h,x,t)}\\
&=\inf_{x\in \sX}\frac{\min\curl*{W\max \curl*{\norm*{x}_p,\gamma}-\gamma W+B,\rho}}{\rho}\,t\\
&=\frac{\min\curl*{B,\rho}}{\rho}\,t\\
&=\sT_1(t),
\end{align*}
where $\sT_1$ is the increasing and convex function on $\bracket*{0,1}$ defined by
\begin{align*}
\forall t \in \bracket*{0,1}, \quad \sT_1(t) = \frac{\min\curl*{B,\rho}}{\rho} \, t \,;
\end{align*}
\begin{align*}
\inf_{h\in\sH_{\mathrm{lin}}:\ov h_\gamma(x)<0}\sC_{\wt{\Phi}_{\rho}}(h,x,t)
& = t+(1-t)\paren*{1-\frac{\min\curl*{W\max \curl*{\norm*{x}_p,\gamma}-\gamma W+B,\rho}}{\rho}}\\
\inf_{x\in \sX} \inf_{h\in\sH_{\mathrm{lin}}\colon \ov h_\gamma(x)< 0}\Delta\sC_{\wt{\Phi}_{\rho},\sH_{\mathrm{lin}}}(h,x,t)
& = \inf_{x\in \sX} \curl*{\inf_{h\in\sH_{\mathrm{lin}}:\ov h_\gamma(x)< 0}\sC_{\wt{\Phi}_{\rho}}(h,x,t)-\inf_{h\in\sH_{\mathrm{lin}}}\sC_{\wt{\Phi}_{\rho}}(h,x,t)}\\
&=\inf_{x\in \sX}(2t-1)\frac{\min\curl*{W\max \curl*{\norm*{x}_p,\gamma}-\gamma W+B,\rho}}{\rho}\\
&=(2t-1)\frac{\min\curl*{B,\rho}}{\rho}\\
&=\sT_2(2t - 1),
\end{align*}
where $\sT_2$ is the increasing and convex function on $\bracket*{0,1}$ defined by
\begin{align*}
\forall t \in [0,1], \quad \sT_2(t) = \frac{\min\curl*{B,\rho}}{\rho} \, t \,;
\end{align*}
By Definition~\ref{def:trans-adv}, for any $\epsilon\geq 0$, the adversarial $\sH_{\mathrm{lin}}$-estimation error transformation of the supremum-based $\rho$-margin loss is as follows:
\begin{align*}
\sT_{\wt{\Phi}_{\rho}}= \frac{\min\curl*{B,\rho}}{\rho} \, t, \quad t \in [0,1],
\end{align*}
Therefore, $\sT_1=\sT_2$ and $\sT_{\wt{\Phi}_{\rho}}$ is convex, non-decreasing, invertible and satisfies that $\sT_{\wt{\Phi}_{\rho}}(0)=0$. By Theorem~\ref{Thm:tightness-adv}, we can choose $\Psi(t)=\frac{\min\curl*{B,\rho}}{\rho}\,t$ in Theorem~\ref{Thm:excess_bounds_Psi_uniform-adv}, or equivalently $\Gamma(t) = \frac{\rho}{\min\curl*{B,\rho}}\,t$ in Theorem~\ref{Thm:excess_bounds_Gamma_uniform-adv}, which are optimal when $\e=0$.
Thus, by Theorem~\ref{Thm:excess_bounds_Psi_uniform-adv} or Theorem~\ref{Thm:excess_bounds_Gamma_uniform-adv}, setting $\e = 0$ yields the adversarial $\sH_{\mathrm{lin}}$-consistency estimation error bound for the supremum-based $\rho$-margin loss, valid for all $h \in \sH_{\mathrm{lin}}$:
\begin{align}
\label{eq:rho-lin-est-adv}
     \sR_{\ell_{\gamma}}(h)- \sR_{\ell_{\gamma},\sH_{\mathrm{lin}}}^*
     \leq \frac{\rho\paren*{\sR_{\wt{\Phi}_{\rho}}(h)- \sR_{\wt{\Phi}_{\rho},\sH_{\mathrm{lin}}}^*+\sM_{\wt{\Phi}_{\rho},\sH_{\mathrm{lin}}}}}{\min\curl*{B,\rho}}-\sM_{\ell_{\gamma}, \sH_{\mathrm{lin}}}.
\end{align}
Since
\begin{align*}
\sM_{\ell_{\gamma},\sH_{\mathrm{lin}}}
& = \sR_{\ell_{\gamma},\sH_{\mathrm{lin}}}^*-\mathbb{E}_{X}\bracket*{\min\curl*{\eta(x),1-\eta(x)}},\\
\sM_{\wt{\Phi}_{\rho},\sH_{\mathrm{lin}}} 
& = \sR_{\wt{\Phi}_{\rho},\sH_{\mathrm{lin}}}^* - \mathbb{E}_{X}\bracket*{\max\curl*{\eta(x),1-\eta(x)}\paren*{1-\frac{\min\curl*{W\max \curl*{\norm*{x}_p,\gamma}-\gamma W+B,\rho}}{\rho}}}\\
& -\mathbb{E}_{X}\bracket*{\min\curl*{\eta(x),1-\eta(x)}},
\end{align*}
inequality \eqref{eq:rho-lin-est-adv} can be rewritten as follows:
\begin{align}
\label{eq:rho-lin-est-adv-2}
     \sR_{\ell_{\gamma}}(h) \leq
     \begin{cases}
     \sR_{\wt{\Phi}_{\rho}}(h)- \mathbb{E}_{X}\bracket*{\max\curl*{\eta(x),1-\eta(x)}\paren*{1-\frac{\min\curl*{W\max \curl*{\norm*{x}_p,\gamma}-\gamma W+B,\rho}}{\rho}}} & \text{if } B \geq \rho\\
    \frac{\rho \paren*{\sR_{\wt{\Phi}_{\rho}}(h)- \mathbb{E}_{X}\bracket*{\max\curl*{\eta(x),1-\eta(x)}\paren*{1-\frac{\min\curl*{W\max \curl*{\norm*{x}_p,\gamma}-\gamma W+B,\rho}}{\rho}}}}}{\min\curl*{B,\rho}}\\
    +\paren*{1-\frac{\rho}{\min\curl*{B,\rho}}}\mathbb{E}_{X}\bracket*{\min\curl*{\eta(x),1-\eta(x)}} & \text{otherwise}.
     \end{cases}
\end{align}
Note that: $\min\curl*{W\max \curl*{\norm*{x}_p,\gamma}-\gamma W+B,\rho} = \rho$ if $B \geq \rho$. Thus, the first
inequality (case $B \geq \rho$) can be equivalently written as follows:
\begin{align}
\label{eq:rho-lin-est-adv-3}
    \forall h \in \sH_{\mathrm{lin}}, \quad \sR_{\ell_{\gamma}}(h) 
     \leq \sR_{\wt{\Phi}_{\rho}}(h).
\end{align}
The case $B \geq \rho$ is one of the ``trivial cases'' mentioned in Section~\ref{sec:general}, where the trivial inequality $\sR_{\ell_{\gamma}}(h) \leq \sR_{\wt{\Phi}_{\rho}}(h)$ can be obtained directly using the fact that $\ell_{\gamma}$ is upper bounded by $\wt{\Phi}_{\rho}$. This, however, does not imply that adversarial $\sH_{\mathrm{lin}}$-consistency estimation error bound for the supremum-based $\rho$-margin loss is trivial when $B>\rho$ since it is optimal.

\subsection{One-Hidden-Layer ReLU Neural Networks}
\label{app:derivation-NN-adv}
By the definition of $\sH_{\mathrm{NN}}$, for any $x \in \sX$, 
\begin{align*}
&\uv h_\gamma(x)=\inf_{x'\colon \|x-x'\|_p\leq \gamma}\sum_{j = 1}^n u_j(w_j \cdot x'+b)_{+}\\
&\ov h_\gamma(x)=\sup_{x'\colon \|x-x'\|_p\leq \gamma}\sum_{j = 1}^n u_j(w_j \cdot x'+b)_{+} 
\end{align*}
Note $\sH_{\mathrm{NN}}$ is symmetric. For any $x\in \sX$, there exist $u=\paren*{\frac{1}{\Lambda},\ldots,\frac{1}{\Lambda}}$, $w=0$ and any $0<b\leq B$ satisfy that $\uv h_\gamma(x)>0$. Thus by Lemma~\ref{lemma:explicit_assumption_01_adv}, for any $x\in \sX$, $\sC^*_{\ell_{\gamma},\sH_{\mathrm{NN}}}(x) =\min\curl*{\eta(x), 1 - \eta(x)}$. The $\paren*{\ell_{\gamma},\sH_{\mathrm{NN}}}$-minimizability gap can be expressed as follows:
\begin{align}
\label{eq:M-01-NN-adv}
\sM_{\ell_{\gamma},\sH_{\mathrm{NN}}}
& = \sR_{\ell_{\gamma},\sH_{\mathrm{NN}}}^*-\mathbb{E}_{X}\bracket*{\min\curl*{\eta(x),1-\eta(x)}}.
\end{align}

\subsubsection{Supremum-Based \texorpdfstring{$\rho$}{rho}-Margin Loss}
\label{app:derivation-NN-adv-rho}
For the supremum-based $\rho$-margin loss 
\begin{align*}
\wt{\Phi}_{\rho}=\sup_{x'\colon \|x-x'\|_p\leq \gamma}\Phi_{\rho}(y h(x')),  \quad \text{where } \Phi_{\rho}(\alpha)=\min\curl*{1,\max\curl*{0,1-\frac{\alpha}{\rho}}},~\rho>0,   
\end{align*}
for all $h\in \sH_{\mathrm{NN}}$ and $x\in \sX$:
\begin{equation*}
\begin{aligned}
\sC_{\wt{\Phi}_{\rho}}(h,x,t) 
&=t \wt{\Phi}_{\rho}(h(x))+(1-t)\wt{\Phi}_{\rho}(-h(x))\\
& = t\Phi_{\rho}\paren*{\uv h_\gamma(x)}+(1-t)\Phi_{\rho}\paren*{-\ov h_\gamma(x)}\\
& =t\min\curl*{1,\max\curl*{0,1-\frac{\uv h_\gamma(x)}{\rho}}}+(1-t)\min\curl*{1,\max\curl*{0,1+\frac{\ov h_\gamma(x)}{\rho}}}.\\
\inf_{h\in\sH_{\mathrm{NN}}}\sC_{\wt{\Phi}_{\rho}}(h,x,t)
& =\max\curl*{t,1-t}\paren*{1-\frac{\min\curl*{\sup_{h\in\sH_{\mathrm{NN}}}\uv h_\gamma(x),\rho}}{\rho}} + \min\curl*{t,1-t}.
\end{aligned}
\end{equation*}
Therefore, the $\paren*{\wt{\Phi}_{\rho},\sH_{\mathrm{NN}}}$-minimizability gap can be expressed as follows:
\begin{equation}
\begin{aligned}
\label{eq:M-rho-NN-adv}
\sM_{\wt{\Phi}_{\rho},\sH_{\mathrm{NN}}}
& = \sR_{\wt{\Phi}_{\rho},\sH_{\mathrm{NN}}}^* - \mathbb{E}_{X}\bracket*{\inf_{h\in\sH_{\mathrm{NN}}}\sC_{\wt{\Phi}_{\rho}}(h,x,\eta(x))}\\
& = \sR_{\wt{\Phi}_{\rho},\sH_{\mathrm{NN}}}^* - \mathbb{E}_{X}\bracket*{\max\curl*{\eta(x),1-\eta(x)}\paren*{1-\frac{\min\curl*{\sup_{h\in\sH_{\mathrm{NN}}}\uv h_\gamma(x),\rho}}{\rho}}}\\
& -\mathbb{E}_{X}\bracket*{\min\curl*{\eta(x),1-\eta(x)}}.
\end{aligned}
\end{equation}
For $\frac{1}2<t\leq1$, we have
\begin{align*}
\inf_{h\in\sH_{\mathrm{NN}}:\uv h_\gamma(x)\leq 0 \leq \ov h_\gamma(x)}\sC_{\wt{\Phi}_{\rho}}(h,x,t) 
& = t+(1-t)\\
& = 1\\
\inf_{x\in \sX} \inf_{h\in\sH_{\mathrm{NN}}\colon \uv h_\gamma(x)\leq 0 \leq \ov h_\gamma(x)}\Delta\sC_{\wt{\Phi}_{\rho},\sH_{\mathrm{NN}}}(h,x,t)
& = \inf_{x\in \sX} \curl*{\inf_{h\in\sH_{\mathrm{NN}}:\uv h_\gamma(x)\leq 0 \leq \ov h_\gamma(x)}\sC_{\wt{\Phi}_{\rho}}(h,x,t)-\inf_{h\in\sH_{\mathrm{NN}}}\sC_{\wt{\Phi}_{\rho}}(h,x,t)}\\
&=\inf_{x\in \sX}\frac{\min\curl*{\sup_{h\in\sH_{\mathrm{NN}}}\uv h_\gamma(x),\rho}}{\rho}\,t\\
&=\frac{\min\curl*{\inf_{x\in\sX}\sup_{h\in\sH_{\mathrm{NN}}}\uv h_\gamma(x),\rho}}{\rho}\,t\\
&=\sT_1(\eta(x)),
\end{align*}
where $\sT_1$ is the increasing and convex function on $\bracket*{0,1}$ defined by
\begin{align*}
\forall t \in \bracket*{0,1}, \quad \sT_1(t) = \frac{\min\curl*{\inf_{x\in\sX}\sup_{h\in\sH_{\mathrm{NN}}}\uv h_\gamma(x),\rho}}{\rho} \, t \,;
\end{align*}
\begin{align*}
\inf_{h\in\sH_{\mathrm{NN}}:\ov h_\gamma(x)<0}\sC_{\wt{\Phi}_{\rho}}(h,x,t)
& = t+(1-t)\paren*{1-\frac{\min\curl*{\sup_{h\in\sH_{\mathrm{NN}}}\uv h_\gamma(x),\rho}}{\rho}}\\
\inf_{x\in \sX} \inf_{h\in\sH_{\mathrm{NN}}\colon \ov h_\gamma(x)< 0}\Delta\sC_{\wt{\Phi}_{\rho},\sH_{\mathrm{NN}}}(h,x,t)
& = \inf_{x\in \sX} \curl*{\inf_{h\in\sH_{\mathrm{NN}}:\ov h_\gamma(x)< 0}\sC_{\wt{\Phi}_{\rho}}(h,x,t)-\inf_{h\in\sH_{\mathrm{NN}}}\sC_{\wt{\Phi}_{\rho}}(h,x,t)}\\
&=\inf_{x\in \sX}(2t-1)\frac{\min\curl*{\sup_{h\in\sH_{\mathrm{NN}}}\uv h_\gamma(x),\rho}}{\rho}\\
&=(2t-1)\frac{\min\curl*{\inf_{x\in\sX}\sup_{h\in\sH_{\mathrm{NN}}}\uv h_\gamma(x),\rho}}{\rho}\\
&=\sT_2(2t - 1),
\end{align*}
where $\sT_2$ is the increasing and convex function on $\bracket*{0,1}$ defined by
\begin{align*}
\forall t \in [0,1], \quad \sT_2(t) = \frac{\min\curl*{\inf_{x\in\sX}\sup_{h\in\sH_{\mathrm{NN}}}\uv h_\gamma(x),\rho}}{\rho} \, t \,;
\end{align*}
By Definition~\ref{def:trans-adv}, for any $\epsilon\geq 0$, the adversarial $\sH_{\mathrm{NN}}$-estimation error transformation of the supremum-based $\rho$-margin loss is as follows:
\begin{align*}
\sT_{\wt{\Phi}_{\rho}}= \frac{\min\curl*{\inf_{x\in\sX}\sup_{h\in\sH_{\mathrm{NN}}}\uv h_\gamma(x),\rho}}{\rho}\,t, \quad t \in [0,1],
\end{align*}
Therefore, $\sT_1=\sT_2$ and $\sT_{\wt{\Phi}_{\rho}}$ is convex, non-decreasing, invertible and satisfies that $\sT_{\wt{\Phi}_{\rho}}(0)=0$. By Theorem~\ref{Thm:tightness-adv}, we can choose $\Psi(t)=\frac{\min\curl*{\inf_{x\in\sX}\sup_{h\in\sH_{\mathrm{NN}}}\uv h_\gamma(x),\rho}}{\rho}\,t$ in Theorem~\ref{Thm:excess_bounds_Psi_uniform-adv}, or equivalently $\Gamma(t) = \frac{\rho}{\min\curl*{\inf_{x\in\sX}\sup_{h\in\sH_{\mathrm{NN}}}\uv h_\gamma(x),\rho}} \, t$ in Theorem~\ref{Thm:excess_bounds_Gamma_uniform-adv}, which are optimal when $\e=0$.
Thus, by Theorem~\ref{Thm:excess_bounds_Psi_uniform-adv} or Theorem~\ref{Thm:excess_bounds_Gamma_uniform-adv}, setting $\e = 0$ yields the adversarial $\sH_{\mathrm{NN}}$-consistency estimation error bound for the supremum-based $\rho$-margin loss, valid for all $h \in \sH_{\mathrm{NN}}$:
\begin{align}
\label{eq:rho-NN-est-adv}
     \sR_{\ell_{\gamma}}(h)- \sR_{\ell_{\gamma},\sH_{\mathrm{NN}}}^*
     \leq \frac{\rho\paren*{\sR_{\wt{\Phi}_{\rho}}(h)- \sR_{\wt{\Phi}_{\rho},\sH_{\mathrm{NN}}}^*+\sM_{\wt{\Phi}_{\rho},\sH_{\mathrm{NN}}}}}{\min\curl*{\inf_{x\in\sX}\sup_{h\in\sH_{\mathrm{NN}}}\uv h_\gamma(x),\rho}}-\sM_{\ell_{\gamma}, \sH_{\mathrm{NN}}}.
\end{align}
Observe that
\begin{align*}
\inf_{x\in\sX}\sup_{h\in\sH_{\mathrm{NN}}}\uv h_\gamma(x)
&\geq
\sup_{h\in\sH_{\mathrm{NN}}}\inf_{x\in\sX}\uv h_\gamma(x)\\
& = \sup_{\|u \|_{1}\leq \Lambda,~\|w_j\|_q\leq W,~\abs*{b}\leq B}\inf_{x\in\sX}\inf_{\|s\|_p\leq \gamma}\sum_{j = 1}^n u_j(w_j \cdot x + w_j \cdot s +b)_{+}\\
& \geq \sup_{\|u \|_{1}\leq \Lambda,~\abs*{b}\leq B} \inf_{x\in\sX}\inf_{\|s\|_p\leq \gamma} \sum_{j = 1}^n u_j(0 \cdot x + 0 \cdot s +b)_{+}\\
& = \sup_{\|u \|_{1}\leq \Lambda,~\abs*{b}\leq B} \sum_{j = 1}^n u_j(b)_{+}\\
& = \Lambda B.
\end{align*}
Thus, the inequality can be relaxed as follows:
\begin{align}
\label{eq:rho-NN-est-adv-2}
     \sR_{\ell_{\gamma}}(h)- \sR_{\ell_{\gamma},\sH_{\mathrm{NN}}}^*
     \leq \frac{\rho\paren*{\sR_{\wt{\Phi}_{\rho}}(h)- \sR_{\wt{\Phi}_{\rho},\sH_{\mathrm{NN}}}^*+\sM_{\wt{\Phi}_{\rho},\sH_{\mathrm{NN}}}}}{\min\curl*{\Lambda B,\rho}}-\sM_{\ell_{\gamma}, \sH_{\mathrm{NN}}}.
\end{align}
Since
\begin{align*}
\sM_{\ell_{\gamma},\sH_{\mathrm{NN}}}
& = \sR_{\ell_{\gamma},\sH_{\mathrm{NN}}}^*-\mathbb{E}_{X}\bracket*{\min\curl*{\eta(x),1-\eta(x)}},\\
\sM_{\wt{\Phi}_{\rho},\sH_{\mathrm{NN}}}
& = \sR_{\wt{\Phi}_{\rho},\sH_{\mathrm{NN}}}^* - \mathbb{E}_{X}\bracket*{\max\curl*{\eta(x),1-\eta(x)}\paren*{1-\frac{\min\curl*{\sup_{h\in\sH_{\mathrm{NN}}}\uv h_\gamma(x),\rho}}{\rho}}}\\
& -\mathbb{E}_{X}\bracket*{\min\curl*{\eta(x),1-\eta(x)}},
\end{align*}
inequality \eqref{eq:rho-NN-est-adv} can be rewritten as follows:
\begin{align*}
\sR_{\ell_{\gamma}}(h)
     \leq
     \begin{cases}
     \sR_{\wt{\Phi}_{\rho}}(h)- \mathbb{E}_{X}\bracket*{\max\curl*{\eta(x),1-\eta(x)}\paren*{1-\frac{\min\curl*{\sup_{h\in\sH_{\mathrm{NN}}}\uv h_\gamma(x),\rho}}{\rho}}} & \text{if } \Lambda B \geq \rho\\
    \frac{\rho \paren*{\sR_{\wt{\Phi}_{\rho}}(h)- \mathbb{E}_{X}\bracket*{\max\curl*{\eta(x),1-\eta(x)}\paren*{1-\frac{\min\curl*{\sup_{h\in\sH_{\mathrm{NN}}}\uv h_\gamma(x),\rho}}{\rho}}}}}{\min\curl*{\Lambda B,\rho}}\\
    +\paren*{1-\frac{\rho}{\min\curl*{\Lambda B,\rho}}}\mathbb{E}_{X}\bracket*{\min\curl*{\eta(x),1-\eta(x)}} & \text{otherwise}.
     \end{cases}
\end{align*}
Observe that
\begin{align*}
\sup_{h\in\sH_{\mathrm{NN}}}\uv h_\gamma(x)
&=
\sup_{ \|u \|_{1}\leq \Lambda,~\|w_j\|_q\leq W,~\abs*{b}\leq B}\inf_{x'\colon \|x-x'\|_p\leq \gamma}\sum_{j = 1}^n u_j(w_j \cdot x'+b)_{+}\\
& \leq \inf_{x'\colon \|x-x'\|_p\leq \gamma} \sup_{ \|u \|_{1}\leq \Lambda,~\|w_j\|_q\leq W,~\abs*{b}\leq B} \sum_{j = 1}^n u_j(w_j \cdot x'+b)_{+} \\
& = \inf_{x'\colon \|x-x'\|_p\leq \gamma} \Lambda\paren*{W\norm*{x'}_p+B}\\
& =
\begin{cases}
\Lambda\paren*{W\norm*{x}_p-\gamma W + B} & \text{if } \norm*{x}_p \geq \gamma\\
\Lambda B & \text{if } \norm*{x}_p < \gamma
\end{cases}\\
& = \Lambda\paren*{W\max \curl*{\norm*{x}_p,\gamma}-\gamma W + B}.
\end{align*}
Thus, the inequality can be further relaxed as follows:
\begin{align}
\label{eq:rho-NN-est-adv-3}
\sR_{\ell_{\gamma}}(h)
     \leq
     \begin{cases}
     \sR_{\wt{\Phi}_{\rho}}(h)- \mathbb{E}_{X}\bracket*{\max\curl*{\eta(x),1-\eta(x)}\paren*{1-\frac{\min\curl*{\Lambda\paren*{W\max \curl*{\norm*{x}_p,\gamma}-\gamma W + B},\rho}}{\rho}}} & \text{if } \Lambda B \geq \rho\\
    \frac{\rho \paren*{\sR_{\wt{\Phi}_{\rho}}(h)- \mathbb{E}_{X}\bracket*{\max\curl*{\eta(x),1-\eta(x)}\paren*{1-\frac{\min\curl*{\Lambda\paren*{W\max \curl*{\norm*{x}_p,\gamma}-\gamma W + B},\rho}}{\rho}}}}}{\min\curl*{\Lambda B,\rho}}\\
    +\paren*{1-\frac{\rho}{\min\curl*{\Lambda B,\rho}}}\mathbb{E}_{X}\bracket*{\min\curl*{\eta(x),1-\eta(x)}} & \text{otherwise}.
     \end{cases}
\end{align}
Note the relaxed adversarial $\sH_{\mathrm{NN}}$-consistency estimation error bounds \eqref{eq:rho-NN-est-adv} and \eqref{eq:rho-NN-est-adv-3} for the supremum-based $\rho$-margin loss are identical to the bounds \eqref{eq:rho-lin-est-adv} and \eqref{eq:rho-lin-est-adv-2} in the linear case respectively  modulo the replacement of $B$ by $\Lambda B$.


\section{Derivation of Non-Adversarial
\texorpdfstring{$\sH_{\mathrm{all}}$}{all}-Estimation Error Bounds under Massart's Noise Condition}
\label{app:derivation-all_noise}
With Massart's noise condition, we introduce a modified $\sH$-estimation error transformation. 
We assume that $\epsilon=0$ throughout this section.
\begin{restatable}{proposition}{TightnessNoise}
\label{prop:prop-noise}
Under Massart's noise condition with $\beta$, the modified $\sH$-estimation error transformation of $\Phi$ for $\epsilon=0$ is defined on $t\in \left[0,1\right]$ by,
\begin{align*}
\sT^M_{\Phi}\paren*{t}= \sT(t)\mathds{1}_{t\in \left[2\beta,1\right]}+(\sT(2\beta)/2\beta)\,t\mathds{1}_{t\in \left[0,2\beta\right)},
\end{align*}
with $\sT(t)$ defined in Definition~\ref{def:trans}. Suppose that $\sH$ satisfies the condition of
Lemma~\ref{lemma:explicit_assumption_01} and $\wt{\sT}^M_{\Phi}$ is any lower bound of $\sT^M_{\Phi}$ such that $\wt{\sT}^M_{\Phi}\leq \sT^M_{\Phi}$. If
$\wt{\sT}^M_{\Phi}$ is convex with $\wt{\sT}^M_{\Phi}(0)=0$, then, for any hypothesis $h\in\sH$ and any distribution under Massart's noise condition with $\beta$,
\ifdim\columnwidth = \textwidth
{
\begin{equation*}
     \wt{\sT}^M_{\Phi}\paren*{\sR_{\ell_{0-1}}(h)- \sR_{\ell_{0-1},\sH}^*+\sM_{\ell_{0-1},\sH}}
     \leq  \sR_{\Phi}(h)-\sR_{\Phi,\sH}^*+\sM_{\Phi,\sH}.
\end{equation*}
}
\else
{
\begin{multline*}
     \sT^M_{\Phi}\paren*{\sR_{\ell_{0-1}}(h)- \sR_{\ell_{0-1},\sH}^*+\sM_{\ell_{0-1},\sH}}\\
     \leq  \sR_{\Phi}(h)-\sR_{\Phi,\sH}^*+\sM_{\Phi,\sH}.
\end{multline*}
}
\fi
\end{restatable}
\begin{proof}
Note the condition~\eqref{eq:condition_Psi_general} in Theorem~\ref{Thm:excess_bounds_Psi_01_general} is symmetric about $\Delta \eta(x)=0$. Thus, condition~\eqref{eq:condition_Psi_general} uniformly holds for all distributions is equivalent to the following holds for any $t\in\left[1/2+\beta,1\right]\colon$
\begin{align}
\label{eq:condition_Psi_general-massarts}
\Psi \paren*{\tri*{2t-1}_{\e}}\leq \inf_{x\in \sX,h\in\sH:h(x)<0}\Delta\sC_{\Phi,\sH}(h,x,t),
\end{align}
It is clear that any lower bound $\wt{\sT}^M_{\Phi}$ of the modified $\sH$-estimation error transformation verified condition~\eqref{eq:condition_Psi_general-massarts}. Then by Theorem~\ref{Thm:excess_bounds_Psi_01_general}, the proof is completed.
\end{proof}
\subsection{Quadratic Loss}
For the quadratic loss $\Phi_{\mathrm{quad}}(\alpha)\colon=(1-\alpha)^2\mathds{1}_{\alpha\leq 1}$, 
for all $h\in \sH_{\mathrm{all}}$ and $x\in \sX$:
\begin{align*}
\sC_{\Phi_{\mathrm{quad}}}(h,x,t)
&=t \Phi_{\mathrm{quad}}(h(x))+(1-t)\Phi_{\mathrm{quad}}(-h(x))\\
&=t\paren*{1-h(x)}^2\mathds{1}_{h(x)\leq 1}+(1-t)\paren*{1+h(x)}^2\mathds{1}_{h(x)\geq -1}.\\
\inf_{h\in\sH_{\mathrm{all}}}\sC_{\Phi_{\mathrm{quad}}}(h,x,t)&=4t(1-t)\\
\sM_{\Phi_{\mathrm{quad}},\sH_{\mathrm{all}}}
& = \sR_{\Phi_{\mathrm{quad}},\sH_{\mathrm{all}}}^*-\mathbb{E}_{X}\bracket*{\inf_{h\in\sH_{\mathrm{all}}}\sC_{\Phi_{\mathrm{quad}}}(h,x,\eta(x))}\\
& = \sR_{\Phi_{\mathrm{quad}},\sH_{\mathrm{all}}}^* - \mathbb{E}_{X}\bracket*{4\eta(x)(1-\eta(x))}\\
&=0
\end{align*}
Thus, for $\frac{1}2< t\leq1$, we have
\begin{align*}
\inf_{h\in\sH_{\mathrm{all}}:h(x)<0}\sC_{\Phi_{\mathrm{quad}}}(h,x,t)
&=t+(1-t)\\
&=1\\
\inf_{x\in \sX}\inf_{h\in\sH_{\mathrm{all}}:h(x)<0}\Delta\sC_{\Phi_{\mathrm{quad}},\sH_{\mathrm{all}}}(h,x,t)
& =\inf_{x\in \sX}\paren*{\inf_{h\in\sH_{\mathrm{all}}:h(x)<0}\sC_{\Phi_{\mathrm{quad}}}(h,x,t)-\inf_{h\in\sH_{\mathrm{all}}}\sC_{\Phi_{\mathrm{quad}}}(h,x,t)}\\
&=\inf_{x\in \sX}\paren*{1-4t(1-t)}\\
&=1-4t(1-t)\\
&=\sT(2t-1)
\end{align*}
where $\sT$ is the increasing and convex function on $[0,1]$ defined by
\begin{align*}
\forall t\in[0,1], \quad
\sT(t)=t^2.
\end{align*}
By
Proposition~\ref{prop:prop-noise}, for $\epsilon= 0$, the modified $\sH_{\mathrm{all}}$-estimation error transformation of the quadratic loss under Massart's noise condition with $\beta$ is as follows:
\begin{align*}
\sT^M_{\Phi_{\mathrm{quad}}}(t)= 
\begin{cases}
2\beta \, t, & t\in \left[0, 2\beta\right], \\
t^2, & t\in \left[2\beta,1\right].
\end{cases}
\end{align*}
Therefore, $\sT^M_{\Phi_{\mathrm{quad}}}$ is convex, non-decreasing, invertible and satisfies that $\sT^M_{\Phi_{\mathrm{quad}}}(0)=0$. By
Proposition~\ref{prop:prop-noise}, we obtain the $\sH_{\mathrm{all}}$-consistency estimation error bound for the quadratic loss, valid for all $h \in \sH_{\mathrm{all}}$ such that $\sR_{\Phi_{\mathrm{quad}}}(h)- \sR_{\Phi_{\mathrm{quad}},\sH_{\mathrm{all}}}^*\leq \sT(2\beta)=4\beta^2$ and distributions $\sD$ satisfies Massart's noise condition with $\beta$:
\begin{align}
    \sR_{\ell_{0-1}}(h)- \sR_{\ell_{0-1},\sH_{\mathrm{all}}}^* \leq
    \frac{\sR_{\Phi_{\mathrm{quad}}}(h)- \sR_{\Phi_{\mathrm{quad}},\sH_{\mathrm{all}}}^*}{2\beta}
\end{align}

\subsection{Logistic Loss}
For the logistic loss $\Phi_{\mathrm{log}}(\alpha)\colon=\log_2(1+e^{-\alpha})$, for all $h\in \sH_{\mathrm{all}}$ and $x\in \sX$:
\begin{align*}
\sC_{\Phi_{\mathrm{log}}}(h,x,t)
& = t \Phi_{\mathrm{log}}(h(x))+(1-t)\Phi_{\mathrm{log}}(-h(x)),\\
& = t\log_2\paren*{1+e^{-h(x)}}+(1-t)\log_2\paren*{1+e^{h(x)}}.\\
\inf_{h\in\sH_{\mathrm{all}}}\sC_{\Phi_{\mathrm{log}}}(h,x,t)
&=-t\log_2(t)-(1-t)\log_2(1-t)\\
\sM_{\Phi_{\mathrm{log}},\sH_{\mathrm{all}}}
& = \sR_{\Phi_{\mathrm{log}},\sH_{\mathrm{all}}}^*-
\mathbb{E}_{X}\bracket*{\inf_{h\in\sH_{\mathrm{all}}}\sC_{\Phi_{\mathrm{log}}}(h,x,\eta(x))}\\
& = \sR_{\Phi_{\mathrm{log}},\sH_{\mathrm{all}}}^*- \mathbb{E}_{X}\bracket*{-\eta(x)\log_2(\eta(x))-(1-\eta(x))\log_2(1-\eta(x))}\\
& = 0
\end{align*}
Thus, for $\frac{1}2< t\leq1$, we have
\begin{align*}
\inf_{h\in\sH_{\mathrm{all}}:h(x)<0}\sC_{\Phi_{\mathrm{log}}}(h,x,t)
& = t\log_2\paren*{1+e^{-0}}+(1-t)\log_2\paren*{1+e^{0}} \\
& = 1, \\
\inf_{x\in \sX}\inf_{h\in\sH_{\mathrm{all}}:h(x)<0}\Delta\sC_{\Phi_{\mathrm{log}},\sH_{\mathrm{all}}}(h,x,t)
&=\inf_{x\in \sX}\paren*{\inf_{h\in\sH_{\mathrm{all}}:h(x)<0}\sC_{\Phi_{\mathrm{log}}}(h,x,t)-\inf_{h\in\sH_{\mathrm{all}}}\sC_{\Phi_{\mathrm{log}}}(h,x,t)}\\
&=\inf_{x\in \sX}\paren*{1+t\log_2(t)+(1-t)\log_2(1-t}\\
&=1+t\log_2(t)+(1-t)\log_2(1-t)\\
&=\sT(2t-1),
\end{align*}
where $\sT$ is the increasing and convex function on $[0,1]$ defined by
\begin{align*}
\forall t\in[0,1], \quad
\sT(t)=\frac{t+1}{2}\log_2(t+1)+\frac{1-t}{2}\log_2(1-t)
\end{align*}
By Proposition~\ref{prop:prop-noise}, for $\epsilon= 0$, the modified $\sH_{\mathrm{all}}$-estimation error transformation of the logistic loss under Massart's noise condition with $\beta$ is as follows:
\begin{align*}
\sT^M_{\Phi_{\mathrm{log}}}= 
\begin{cases}
\sT(t), & t\in \left[2\beta,1\right], \\
\frac{\sT(2\beta)}{2\beta}\, t, &  t\in \left[0,2\beta\right).
\end{cases}
\end{align*}
Therefore, $\sT^M_{\Phi_{\mathrm{log}}}$ is convex, non-decreasing, invertible and satisfies that $\sT^M_{\Phi_{\mathrm{log}}}(0)=0$. By  Proposition~\ref{prop:prop-noise}, we obtain the $\sH_{\mathrm{all}}$-consistency estimation error bound for the logistic loss, valid for all $h \in \sH_{\mathrm{all}}$ such that $\sR_{\Phi_{\mathrm{log}}}(h)- \sR_{\Phi_{\mathrm{log}},\sH_{\mathrm{all}}}^*\leq \sT(2\beta)=\frac{2\beta+1}{2}\log_2(2\beta+1)+\frac{1-2\beta}{2}\log_2(1-2\beta)$ and distributions $\sD$ satisfies Massart's noise condition with $\beta$:
\begin{align}
     \sR_{\ell_{0-1}}(h)-\sR_{\ell_{0-1},\sH_{\mathrm{all}}}^* \leq 
     \frac{2\beta\paren*{\sR_{\Phi_{\mathrm{log}}}(h)- \sR_{\Phi_{\mathrm{log}},\sH_{\mathrm{all}}}^*}}{\frac{2\beta+1}{2}\log_2(2\beta+1)+\frac{1-2\beta}{2}\log_2(1-2\beta)}
\end{align}

\subsection{Exponential Loss}
For the exponential loss $\Phi_{\mathrm{exp}}(\alpha)\colon=e^{-\alpha}$, for all $h\in \sH_{\mathrm{all}}$ and $x\in \sX$:
\begin{align*}
\sC_{\Phi_{\mathrm{exp}}}(h,x,t)
&=t \Phi_{\mathrm{exp}}(h(x))+(1-t)\Phi_{\mathrm{exp}}(-h(x))\\
&=t e^{-h(x)}+(1-t)e^{h(x)}.\\
\inf_{h\in\sH_{\mathrm{all}}}\sC_{\Phi_{\mathrm{exp}}}(h,x,t)
&=2\sqrt{t(1-t)}\\
\sM_{\Phi_{\mathrm{exp}},\sH_{\mathrm{all}}}
& = \sR_{\Phi_{\mathrm{exp}},\sH_{\mathrm{all}}}^*-
\mathbb{E}_{X}\bracket*{\inf_{h\in\sH_{\mathrm{all}}}\sC_{\Phi_{\mathrm{exp}}}(h,x,\eta(x))}\\
& = \sR_{\Phi_{\mathrm{exp}},\sH_{\mathrm{all}}}^*-
\mathbb{E}_{X}\bracket*{2\sqrt{\eta(x)(1-\eta(x))}}\\
& = 0.
\end{align*}
Thus, for $\frac{1}2< t\leq1$, we have
\begin{align*}
\inf_{h\in\sH_{\mathrm{all}}:h(x)<0}\sC_{\Phi_{\mathrm{exp}}}(h,x,t)
&=te^{-0}+(1-t)e^{0}\\
&=1.\\
\inf_{x\in \sX}\inf_{h\in\sH_{\mathrm{all}}:h(x)<0}\Delta\sC_{\Phi_{\mathrm{exp}},\sH_{\mathrm{all}}}(h,x)
&=\inf_{x\in \sX}\paren*{\inf_{h\in\sH_{\mathrm{all}}:h(x)<0}\sC_{\Phi_{\mathrm{exp}}}(h,x)-\inf_{h\in\sH_{\mathrm{all}}}\sC_{\Phi_{\mathrm{exp}}}(h,x)}\\
&=\inf_{x\in \sX}\paren*{1-2\sqrt{t(1-t)}}\\
&=1-2\sqrt{t(1-t)}\\
&=\sT(2t-1),
\end{align*}
where $\sT$ is the increasing and convex function on $[0,1]$ defined by
\begin{align*}
\forall t\in[0,1], \quad 
\sT(t)=1-\sqrt{1-t^2}.
\end{align*}
By Proposition~\ref{prop:prop-noise}, for $\epsilon= 0$, the modified $\sH_{\mathrm{all}}$-estimation error transformation of the exponential loss under Massart's noise condition with $\beta$ is as follows:
\begin{align*}
\sT^M_{\Phi_{\mathrm{exp}}}= 
\begin{cases}
\sT(t), & t\in \left[2\beta,1\right], \\
\frac{\sT(2\beta)}{2\beta}\, t, &  t\in \left[0,2\beta\right).
\end{cases}
\end{align*}
Therefore, $\sT^M_{\Phi_{\mathrm{exp}}}$ is convex, non-decreasing, invertible and satisfies that $\sT^M_{\Phi_{\mathrm{exp}}}(0)=0$. By Proposition~\ref{prop:prop-noise}, we obtain the $\sH_{\mathrm{all}}$-consistency estimation error bound for the exponential loss, valid for all $h \in \sH_{\mathrm{all}}$ such that $\sR_{\Phi_{\mathrm{exp}}}(h)- \sR_{\Phi_{\mathrm{exp}},\sH_{\mathrm{all}}}^*\leq \sT(2\beta)=1-\sqrt{1-4\beta^2}$ and distributions $\sD$ satisfies Massart's noise condition with $\beta$:
\begin{align}
     &\sR_{\ell_{0-1}}(h)-\sR_{\ell_{0-1},\sH_{\mathrm{all}}}^*\leq 
     \frac{2\beta\paren*{\sR_{\Phi_{\mathrm{exp}}}(h)- \sR_{\Phi_{\mathrm{exp}},\sH_{\mathrm{all}}}^*}}{1-\sqrt{1-4\beta^2}}
\end{align}

\section{Derivation of Adversarial
\texorpdfstring{$\sH$}{H}-Estimation Error Bounds under Massart's Noise Condition}
\label{app:derivation-adv_noise}
As with the non-adversarial scenario in Section~\ref{sec:noise-non-adv},  we
introduce a modified adversarial $\sH$-estimation error transformation. We assume that $\epsilon=0$ throughout this section.
\begin{restatable}{proposition}{TransNoiseAdv}
\label{prop-adv-noise}
Under Massart's noise condition with $\beta$, the modified adversarial $\sH$-estimation error transformation of $\wt{\Phi}$ for $\epsilon=0$ is defined on $t\in \left[0,1\right]$ by 
\begin{align*}
\sT^M_{\wt{\Phi}}\paren*{t}=
\min\curl*{\sT^M_1(t), \sT^M_2(t)},
\end{align*}
where 
\begin{align*}
    &\sT^M_1(t):=\h{\sT}_1(t)\mathds{1}_{t\in \left[\frac{1}{2}+\beta,1\right]}+ 2/(1+2\beta)\,\h{\sT}_1\paren*{\frac{1}{2}+\beta}\, t\mathds{1}_{t\in \left[0,\frac{1}{2}+\beta\right)},\\
    &\sT^M_2(t):=\h{\sT}_2(t)\mathds{1}_{t\in \left[2\beta,1\right]}+ \frac{\h{\sT}_2(2\beta)}{2\beta}\,t\mathds{1}_{t\in \left[0,2\beta\right)},
\end{align*}
with $\h{\sT}_1(t)$ and $\h{\sT}_2(t)$ defined in Definition~\ref{def:trans-adv}. Suppose that $\sH$ is symmetric and $\wt{\sT}^M_{\wt{\Phi}}$ is any lower bound of $\sT^M_{\wt{\Phi}}$ such that $\wt{\sT}^M_{\wt{\Phi}}\leq \sT^M_{\wt{\Phi}}$. If
$\wt{\sT}^M_{\wt{\Phi}}$ is convex with
$\wt{\sT}^M_{\wt{\Phi}}(0)=0$, then, for any hypothesis $h\in\sH$ and any distribution under Massart's noise condition with $\beta$,
\ifdim\columnwidth=\textwidth
{
\begin{equation*}
     \wt{\sT}^M_{\wt{\Phi}}\paren*{\sR_{\ell_{\gamma}}(h) - \sR_{\ell_{\gamma},\sH}^* + \sM_{\ell_{\gamma},\sH}}
     \leq  \sR_{\wt{\Phi}}(h)
     - \sR_{\wt{\Phi},\sH}^* + \sM_{\wt{\Phi},\sH}.
\end{equation*}
}
\else
{
\begin{multline*}
    \sT^M_{\wt{\Phi}}\paren*{\sR_{\ell_{\gamma}}(h)- \sR_{\ell_{\gamma},\sH}^*+\sM_{\ell_{\gamma},\sH}}\\
     \leq  \sR_{\wt{\Phi}}(h)-\sR_{\wt{\Phi},\sH}^*
     +\sM_{\wt{\Phi},\sH}.
\end{multline*}
}
\fi
\end{restatable}
\begin{proof}
Note the condition~\eqref{eq:condition_Psi_general_adv} in Theorem~\ref{Thm:excess_bounds_Psi_01_general_adv} is symmetric about $\Delta \eta(x)=0$. Thus, condition~\eqref{eq:condition_Psi_general_adv} uniformly holds for all distributions under Massart's noise condition with $\beta$ is equivalent to the following holds for any $t\in\left[1/2+\beta,1\right]\colon$
\begin{equation}
\label{eq:condition_Psi_general_adv-massarts}
\begin{aligned}
&\Psi\paren*{\tri*{t}_{\e}}  \leq \inf_{x\in\sX,h\in \ov \sH_\gamma\subsetneqq \sH}\Delta\sC_{\wt{\Phi},\sH}(h,x,t),\\
&\Psi\paren*{\tri*{2t-1}_{\e}} \leq \inf_{x\in \sX,h\in\sH\colon  \ov h_\gamma(x)< 0}\Delta\sC_{\wt{\Phi},\sH}(h,x,t),
\end{aligned}
\end{equation}
It is clear that any lower bound $\wt{\sT}^M_{\wt{\Phi}}$ of the modified adversarial $\sH$-estimation error transformation verified condition~\eqref{eq:condition_Psi_general_adv-massarts}. Then by Theorem~\ref{Thm:excess_bounds_Psi_01_general_adv}, the proof is completed.
\end{proof}
\subsection{Linear Hypotheses}
\label{app:derivation-lin-adv_noise}
By the definition of $\sH_{\mathrm{lin}}$, for any $x \in \sX$, 
\begin{align*}
&\uv h_\gamma(x)
  =w \cdot x-\gamma \|w\|_q+b \\
& \in 
  \begin{cases}
\bracket*{-W \norm*{x}_p-\gamma W-B, W\norm*{x}_p - \gamma W+ B}
&  \norm*{x}_p \geq \gamma\\
\bracket*{-W \norm*{x}_p-\gamma W-B, B}
&  \norm*{x}_p < \gamma
\end{cases},\\
&\ov h_\gamma(x)=w \cdot x+\gamma \|w\|_q+b \\
& \in 
\begin{cases}
\bracket*{-W \norm*{x}_p+\gamma W-B, W\norm*{x}_p + \gamma W+ B}
&  \norm*{x}_p \geq \gamma\\
\bracket*{-B, W\norm*{x}_p + \gamma W+ B}
&  \norm*{x}_p < \gamma
\end{cases}.
\end{align*}
Note $\sH_{\mathrm{lin}}$ is symmetric. For any $x\in \sX$, there exist $w=0$ and any $0<b\leq B$ such that $w \cdot x-\gamma \|w\|_q+b>0$. Thus by Lemma~\ref{lemma:explicit_assumption_01_adv}, for any $x\in \sX$, $\sC^*_{\ell_{\gamma},\sH_{\mathrm{lin}}}(x) =\min\curl*{\eta(x), 1 - \eta(x)}$. The $\paren*{\ell_{\gamma},\sH_{\mathrm{lin}}}$-minimizability gap can be expressed as follows:
\begin{align*}
\sM_{\ell_{\gamma},\sH_{\mathrm{lin}}}
& = \sR_{\ell_{\gamma},\sH_{\mathrm{lin}}}^*-\mathbb{E}_{X}\bracket*{\min\curl*{\eta(x),1-\eta(x)}}.
\end{align*}
\subsubsection{Supremum-Based Hinge Loss}
For the supremum-based hinge loss 
\begin{align*}
\wt{\Phi}_{\mathrm{hinge}}\colon=\sup_{x'\colon \|x-x'\|_p\leq \gamma}\Phi_{\mathrm{hinge}}(y h(x')),  \quad \text{where } \Phi_{\mathrm{hinge}}(\alpha)=\max\curl*{0,1 - \alpha},  
\end{align*}
for all $h\in \sH_{\mathrm{lin}}$ and $x\in \sX$:
\small
\begin{align*}
\sC_{\wt{\Phi}_{\mathrm{hinge}}}(h,x,t) 
&=t \wt{\Phi}_{\mathrm{hinge}}(h(x))+(1-t)\wt{\Phi}_{\mathrm{hinge}}(-h(x))\\
& = t\Phi_{\mathrm{hinge}}\paren*{\uv h_\gamma(x)}+(1-t)\Phi_{\mathrm{hinge}}\paren*{-\ov h_\gamma(x)}\\
& =t\max\curl*{0,1-\uv h_\gamma(x)} +(1-t)\max\curl*{0,1+\ov h_\gamma(x)}\\
& \geq \bracket*{ t\max\curl*{0,1-\ov h_\gamma(x)} +(1-t)\max\curl*{0,1+\ov h_\gamma(x)} }\wedge \bracket*{ t\max\curl*{0,1-\uv h_\gamma(x)} +(1-t)\max\curl*{0,1+\uv h_\gamma(x)}} \\
\inf_{h\in\sH_{\mathrm{lin}}}\sC_{\wt{\Phi}_{\mathrm{hinge}}}(h,x,t) &\geq \inf_{h\in\sH_{\mathrm{lin}}}\bracket*{ t\max\curl*{0,1-\ov h_\gamma(x)} +(1-t)\max\curl*{0,1+\ov h_\gamma(x)} }\wedge \inf_{h\in\sH_{\mathrm{lin}}}\bracket*{ t\max\curl*{0,1-\uv h_\gamma(x)} +(1-t)\max\curl*{0,1+\uv h_\gamma(x)}} \\
& = 1-\abs*{2t-1}\min\curl*{W\max\curl*{\norm*{x}_p,\gamma}-\gamma W+B,1}\\
\inf_{h\in\sH_{\mathrm{lin}}}\sC_{\wt{\Phi}_{\mathrm{hinge}}}(h,x,t)&=\inf_{h\in\sH_{\mathrm{lin}}}\bracket*{t\max\curl*{0,1-\uv h_\gamma(x)} +(1-t)\max\curl*{0,1+\ov h_\gamma(x)}}\\
& = \inf_{h\in\sH_{\mathrm{lin}}}
\bracket*{t\max\curl*{0,1-w \cdot x+\gamma \|w\|_q-b}+(1-t)\max\curl*{0,1+w \cdot x+\gamma \|w\|_q+b}}\\
& \leq \inf_{b\in [-B,B]}\bracket*{t\max\curl*{0,1-b}+(1-t)\max\curl*{0,1+b}}\\
&= 1-\abs*{2t-1}\min\curl*{B,1}\\
\sM_{\wt{\Phi}_{\mathrm{hinge}},\sH_{\mathrm{lin}}} 
& = \sR_{\wt{\Phi}_{\mathrm{hinge}},\sH_{\mathrm{lin}}}^* - \mathbb{E}\bracket*{\inf_{h\in\sH_{\mathrm{lin}}}\sC_{\wt{\Phi}_{\mathrm{hinge}}}(h,x,\eta(x))}\\
& \leq \sR_{\wt{\Phi}_{\mathrm{hinge}},\sH_{\mathrm{lin}}}^* - \mathbb{E}\bracket*{1-\abs*{2\eta(x)-1}\min\curl*{W\max\curl*{\norm*{x}_p,\gamma}-\gamma W+B,1}}
\end{align*}
\normalsize
Thus, for $\frac{1}2< t\leq1$, we have
\begin{align*}
\inf_{h\in\sH_{\mathrm{lin}}:\uv h_\gamma(x)\leq 0 \leq \ov h_\gamma(x)}\sC_{\wt{\Phi}_{\mathrm{hinge}}}(h,x,t)
& = t+(1-t)\\
& =1\\
\inf_{x\in \sX} \inf_{h\in\sH_{\mathrm{lin}}\colon \uv h_\gamma(x)\leq 0 \leq \ov h_\gamma(x)}\Delta\sC_{\wt{\Phi}_{\mathrm{hinge}},\sH_{\mathrm{lin}}}(h,x,t)
& = \inf_{x\in \sX} \curl*{1-\inf_{h\in\sH_{\mathrm{lin}}}\sC_{\wt{\Phi}_{\mathrm{hinge}}}(h,x,t)}\\
&\geq\inf_{x\in \sX}\paren*{2t-1}\min\curl*{B,1}\\
&=\paren*{2t-1}\min\curl*{B,1}\\
&=\sT_1(t),
\end{align*}
where $\sT_1$ is the increasing and convex function on $\bracket*{0,1}$ defined by
\begin{align*}
\sT_1(t) =
\begin{cases}
\min\curl*{B,1} \, (2t-1), & t \in \bracket*{1/2+\beta,1},\\ \min\curl*{B,1}\frac{4\beta}{1+2\beta}\, t, & t\in \left[0,1/2+\beta\right).
\end{cases}
\end{align*}
\begin{align*}
\inf_{h\in\sH_{\mathrm{lin}}:\ov h_\gamma(x)<0}\sC_{\wt{\Phi}_{\mathrm{hinge}}}(h,x,t)
& \geq \inf_{h\in\sH_{\mathrm{lin}}:\ov h_\gamma(x)<0}\, \bracket*{t\max\curl*{0,1-\ov h_\gamma(x)} +(1-t)\max\curl*{0,1+\ov h_\gamma(x)}}\\
& = t\max\curl*{0,1-0}+(1-t)\max\curl*{0,1+0}\\
& =1\\
\inf_{x\in \sX} \inf_{h\in\sH_{\mathrm{lin}}\colon \ov h_\gamma(x)< 0}\Delta\sC_{\wt{\Phi}_{\mathrm{hinge}},\sH_{\mathrm{lin}}}(h,x,t)
& = \inf_{x\in \sX} \curl*{\inf_{h\in\sH_{\mathrm{lin}}:\ov h_\gamma(x)< 0}\sC_{\wt{\Phi}_{\mathrm{hinge}}}(h,x,t)-\inf_{h\in\sH_{\mathrm{lin}}}\sC_{\wt{\Phi}_{\mathrm{hinge}}}(h,x,t)}\\
&\geq\inf_{x\in \sX}\paren*{2t-1}\min\curl*{B,1}\\
&=\paren*{2t-1}\min\curl*{B,1}\\
&=\sT_2(2t - 1),
\end{align*}
where $\sT_2$ is the increasing and convex function on $\bracket*{0,1}$ defined by
\begin{align*}
\forall t \in [0,1], \quad \sT_2(t) = \min\curl*{B,1} \, t.
\end{align*}
By Proposition~\ref{prop-adv-noise}, for $\epsilon= 0$, the modified adversarial $\sH_{\mathrm{lin}}$-estimation error transformation of the supremum-based hinge loss under Massart's noise condition with $\beta$ is lower bounded as follows:
\begin{align*}
\sT^M_{\wt{\Phi}_{\mathrm{hinge}}}\geq\wt{\sT}^M_{\wt{\Phi}_{\mathrm{hinge}}}:=\min\curl*{\sT_1,\sT_2}=\begin{cases}
\min\curl*{B,1} \, (2t-1), & t \in \bracket*{1/2+\beta,1},\\ \min\curl*{B,1}\frac{4\beta}{1+2\beta}\, t, & t\in \left[0,1/2+\beta\right).
\end{cases}
\end{align*}
Note $\wt{\sT}^M_{\wt{\Phi}_{\mathrm{hinge}}}$ is convex, non-decreasing, invertible and satisfies that $\wt{\sT}^M_{\wt{\Phi}_{\mathrm{hinge}}}(0)=0$. By Proposition~\ref{prop-adv-noise}, using the fact that
$\wt{\sT}^M_{\wt{\Phi}_{\mathrm{hinge}}}\geq \min\curl*{B,1}\frac{4\beta}{1+2\beta}\, t$ yields the adversarial $\sH_{\mathrm{lin}}$-consistency estimation error bound for the supremum-based hinge loss, valid for all $h \in \sH_{\mathrm{lin}}$ and distributions $\sD$ satisfies Massart's noise condition with $\beta$:
\begin{align}
\label{eq:hinge-lin-est-adv}
     & \sR_{\ell_{\gamma}}(h)- \sR_{\ell_{\gamma},\sH_{\mathrm{lin}}}^* \leq 
     \frac{1+2\beta}{4\beta}\frac{\sR_{\wt{\Phi}_{\mathrm{hinge}}}(h)-\sR_{\wt{\Phi}_{\mathrm{hinge}},\sH_{\mathrm{lin}}}^*+\sM_{\wt{\Phi}_{\mathrm{hinge}},\sH_{\mathrm{lin}}}}{\min\curl*{B,1}}-\sM_{\ell_{\gamma}, \sH_{\mathrm{lin}}}
\end{align}
Since
\begin{align*}
\sM_{\ell_{\gamma},\sH_{\mathrm{lin}}}
& = \sR_{\ell_{\gamma},\sH_{\mathrm{lin}}}^*-\mathbb{E}_{X}\bracket*{\min\curl*{\eta(x),1-\eta(x)}},\\
\sM_{\wt{\Phi}_{\mathrm{hinge}},\sH_{\mathrm{lin}}} 
& \leq \sR_{\wt{\Phi}_{\mathrm{hinge}},\sH_{\mathrm{lin}}}^* - \mathbb{E}\bracket*{1-\abs*{2\eta(x)-1}\min\curl*{W\max\curl*{\norm*{x}_p,\gamma}-\gamma W+B,1}},
\end{align*}
the inequality can be relaxed as follows:
\begin{align*}
     \sR_{\ell_{\gamma}}(h) 
     \leq 
     \frac{1+2\beta}{4\beta}\frac{\sR_{\wt{\Phi}_{\mathrm{hinge}}}(h)}{\min\curl*{B,1}}+\mathbb{E}_{X}\bracket*{\min\curl*{\eta(x),1-\eta(x)}}-\frac{1+2\beta}{4\beta}\frac{\mathbb{E}\bracket*{1-\abs*{2\eta(x)-1}\min\curl*{W\max\curl*{\norm*{x}_p,\gamma}-\gamma W+B,1}}}{\min\curl*{B,1}}
\end{align*}
Note that: $\min\curl*{W\max\curl*{\norm*{x}_p,\gamma}-\gamma W+B,1}\leq 1$ and $1-\abs*{1-2\eta(x)}=2\min\curl*{\eta(x),1-\eta(x)}$. Thus the inequality can be further relaxed as follows:
\begin{align*}
     \sR_{\ell_{\gamma}}(h)
     \leq \frac{1+2\beta}{4\beta}\frac{\sR_{\wt{\Phi}_{\mathrm{hinge}}}(h)}{ \min\curl*{B,1}}-\paren*{\frac{1+2\beta}{2\beta \min\curl*{B,1}}-1}\,\mathbb{E}_{X}\bracket*{\min\curl*{\eta(x),1-\eta(x)}}.
\end{align*}
When $B\geq 1$, it can be equivalently written as follows:
\begin{align}
\label{eq:hinge-lin-est-adv-2}
     \sR_{\ell_{\gamma}}(h)
     \leq \frac{1+2\beta}{4\beta}\,\sR_{\wt{\Phi}_{\mathrm{hinge}}}(h)-\frac{1}{2\beta }\,\mathbb{E}_{X}\bracket*{\min\curl*{\eta(x),1-\eta(x)}}.
\end{align}

\subsubsection{Supremum-Based Sigmoid Loss}
For the supremum-based sigmoid loss 
\begin{align*}
\wt{\Phi}_{\mathrm{sig}}\colon=\sup_{x'\colon \|x-x'\|_p\leq \gamma}\Phi_{\mathrm{sig}}(y h(x')),  \quad \text{where } \Phi_{\mathrm{sig}}(\alpha)=1-\tanh(k\alpha),~k>0,   
\end{align*}
for all $h\in \sH_{\mathrm{lin}}$ and $x\in \sX$:
\begin{align*}
\sC_{\wt{\Phi}_{\mathrm{sig}}}(h,x,t) 
&=t \wt{\Phi}_{\mathrm{sig}}(h(x))+(1-t)\wt{\Phi}_{\mathrm{sig}}(-h(x))\\
& = t\Phi_{\mathrm{sig}}\paren*{\uv h_\gamma(x)}+(1-t)\Phi_{\mathrm{sig}}\paren*{-\ov h_\gamma(x)}\\
& =t\paren*{1-\tanh(k \uv h_\gamma(x))} +(1-t)\paren*{1+\tanh(k \ov h_\gamma(x))}\\
& \geq \max\curl*{1+\paren{1-2t}\tanh(k\ov h_\gamma(x)), 1+\paren{1-2t}\tanh(k\uv h_\gamma(x))}\\
\inf_{h\in\sH_{\mathrm{lin}}}\sC_{\wt{\Phi}_{\mathrm{sig}}}(h,x,t)
& \geq \max\curl*{\inf_{h\in\sH_{\mathrm{lin}}}\bracket*{1+\paren{1-2t}\tanh(k\ov h_\gamma(x))}, \inf_{h\in\sH_{\mathrm{lin}}}\bracket*{1+\paren{1-2t}\tanh(k\uv h_\gamma(x))}}\\
& = 1 - \abs*{1-2t}\tanh\paren*{k\paren*{W\max\curl*{\norm*{x}_p,\gamma}-\gamma W+B}}\\
\inf_{h\in\sH_{\mathrm{lin}}}\sC_{\wt{\Phi}_{\mathrm{sig}}}(h,x,t)
&= \inf_{h\in\sH_{\mathrm{lin}}}\bracket*{t\paren*{1-\tanh(k \paren*{w \cdot x-\gamma \|w\|_q+b})} +(1-t)\paren*{1+\tanh(k\paren*{ w \cdot x+\gamma \|w\|_q+b})}}\\
&\leq \inf_{b\in [-B,B]}\bracket*{t\paren*{1-\tanh(kb)} +(1-t)\paren*{1+\tanh(kb)}}\\
& = \max\curl*{t,1-t}\paren*{1-\tanh\paren*{k B }}+\min\curl*{t,1-t}\paren*{1+\tanh\paren*{k B}}\\
& = 1 - \abs*{1-2t}\tanh\paren*{k B}\\
\sM_{\wt{\Phi}_{\mathrm{sig}},\sH_{\mathrm{lin}}} 
& = \sR_{\wt{\Phi}_{\mathrm{sig}},\sH_{\mathrm{lin}}}^* - \mathbb{E}\bracket*{\inf_{h\in\sH_{\mathrm{lin}}}\sC_{\wt{\Phi}_{\mathrm{sig}}}(h,x,\eta(x))}\\
& \leq \sR_{\wt{\Phi}_{\mathrm{sig}},\sH_{\mathrm{lin}}}^* - \mathbb{E}\bracket*{1 - \abs*{1-2\eta(x)}\tanh\paren*{k\paren*{W\max\curl*{\norm*{x}_p,\gamma}-\gamma W+B}}}
\end{align*}
Thus, for $\frac{1}2< t\leq1$, we have
\begin{align*}
\inf_{h\in\sH_{\mathrm{lin}}:\uv h_\gamma(x)\leq 0 \leq \ov h_\gamma(x)}\sC_{\wt{\Phi}_{\mathrm{sig}}}(h,x,t) 
& = t+(1-t)\\
& =1\\
\inf_{x\in \sX} \inf_{h\in\sH_{\mathrm{lin}}\colon \uv h_\gamma(x)\leq 0 \leq \ov h_\gamma(x)}\Delta\sC_{\wt{\Phi}_{\mathrm{sig}},\sH_{\mathrm{lin}}}(h,x,t)
& = \inf_{x\in \sX} \curl*{1-\inf_{h\in\sH_{\mathrm{lin}}}\sC_{\wt{\Phi}_{\mathrm{sig}}}(h,x,t)}\\
&\geq \inf_{x\in \sX} (2t-1)\tanh\paren*{k B}\\
&=(2t-1)\tanh\paren*{k B}\\
&=\sT_1(t),
\end{align*}
where $\sT_1$ is the increasing and convex function on $\bracket*{0,1}$ defined by
\begin{align*}
 \sT_1(t) = \begin{cases}
\tanh\paren*{k B} \frac{4\beta}{1+2\beta} \, t, & t\in\bracket*{0,1/2+\beta},\\
\tanh\paren*{k B} \, (2t-1),  & t \in \bracket*{1/2+\beta,1}.
\end{cases}    
\end{align*}
\begin{align*}
\inf_{h\in\sH_{\mathrm{lin}}:\ov h_\gamma(x)<0}\sC_{\wt{\Phi}_{\mathrm{sig}}}(h,x,t)
& \geq \inf_{h\in\sH_{\mathrm{lin}}:\ov h_\gamma(x)<0} 
\bracket*{1+\paren{1-2t}\tanh(k\ov h_\gamma(x))}\\
& = 1\\
\inf_{x\in \sX} \inf_{h\in\sH_{\mathrm{lin}}\colon \ov h_\gamma(x)< 0}\Delta\sC_{\wt{\Phi}_{\mathrm{sig}},\sH_{\mathrm{lin}}}(h,x,t)
& = \inf_{x\in \sX} \curl*{\inf_{h\in\sH_{\mathrm{lin}}:\ov h_\gamma(x)< 0}\sC_{\wt{\Phi}_{\mathrm{sig}}}(h,x,t)-\inf_{h\in\sH_{\mathrm{lin}}}\sC_{\wt{\Phi}_{\mathrm{sig}}}(h,x,t)}\\
&\geq \inf_{x\in \sX}(2t-1)\tanh\paren*{k B}\\
&=(2t-1)\tanh\paren*{k B}\\
&=\sT_2(2t-1),
\end{align*}
where $\sT_2$ is the increasing and convex function on $\bracket*{2\beta,1}$ defined by
\begin{align*}
\forall t \in [0,1], \quad \sT_2(t) = \tanh\paren*{k B} \, t \,;
\end{align*}
By Proposition~\ref{prop-adv-noise}, for $\epsilon= 0$, the modified adversarial $\sH_{\mathrm{lin}}$-estimation error transformation of the supremum-based sigmoid loss under Massart's noise condition with $\beta$ is lower bounded as follows:
\begin{align*}
\sT^M_{\wt{\Phi}_{\mathrm{sig}}}\geq \wt{\sT}^M_{\wt{\Phi}_{\mathrm{sig}}}=\min\curl*{\sT_1,\sT_2}=\begin{cases}
\tanh\paren*{k B} \frac{4\beta}{1+2\beta} \, t, & t\in\bracket*{0,1/2+\beta},\\
\tanh\paren*{k B} \, (2t-1),  & t \in \bracket*{1/2+\beta,1}.
\end{cases}    
\end{align*}
Note $\wt{\sT}^M_{\wt{\Phi}_{\mathrm{sig}}}$ is convex, non-decreasing, invertible and satisfies that $\wt{\sT}^M_{\wt{\Phi}_{\mathrm{sig}}}(0)=0$. By Proposition~\ref{prop-adv-noise}, using the fact that
$\wt{\sT}^M_{\wt{\Phi}_{\mathrm{sig}}}\geq \tanh\paren*{k B} \frac{4\beta}{1+2\beta} \, t$ yields the adversarial $\sH_{\mathrm{lin}}$-consistency estimation error bound for the supremum-based sigmoid loss, valid for all $h \in \sH_{\mathrm{lin}}$ and distributions $\sD$ satisfies Massart's noise condition with $\beta$:
\begin{align}
\label{eq:sig-lin-est-adv}
     \sR_{\ell_{\gamma}}(h)- \sR_{\ell_{\gamma},\sH_{\mathrm{lin}}}^* 
     \leq 
     \frac{1+2\beta}{4\beta}\frac{\sR_{\wt{\Phi}_{\mathrm{sig}}}(h)-\sR_{\wt{\Phi}_{\mathrm{sig}},\sH_{\mathrm{lin}}}^*+\sM_{\wt{\Phi}_{\mathrm{sig}},\sH_{\mathrm{lin}}}}{\tanh\paren*{k B}}-\sM_{\ell_{\gamma}, \sH_{\mathrm{lin}}}.
\end{align}
Since
\begin{align*}
\sM_{\ell_{\gamma},\sH_{\mathrm{lin}}}
& = \sR_{\ell_{\gamma},\sH_{\mathrm{lin}}}^*-\mathbb{E}_{X}\bracket*{\min\curl*{\eta(x),1-\eta(x)}},\\
\sM_{\wt{\Phi}_{\mathrm{sig}},\sH_{\mathrm{lin}}} 
& \leq \sR_{\wt{\Phi}_{\mathrm{sig}},\sH_{\mathrm{lin}}}^* - \mathbb{E}\bracket*{1 - \abs*{1-2\eta(x)}\tanh\paren*{k\paren*{W\max\curl*{\norm*{x}_p,\gamma}-\gamma W+B}}},
\end{align*}
the inequality can be relaxed as follows:
\begin{align*}
     \sR_{\ell_{\gamma}}(h) 
     \leq 
     \frac{1+2\beta}{4\beta}\frac{\sR_{\wt{\Phi}_{\mathrm{sig}}}(h)}{\tanh\paren*{k B}}+\mathbb{E}_{X}\bracket*{\min\curl*{\eta(x),1-\eta(x)}}-\frac{1+2\beta}{4\beta}\frac{\mathbb{E}\bracket*{1 - \abs*{1-2\eta(x)}\tanh\paren*{k\paren*{W\max\curl*{\norm*{x}_p,\gamma}-\gamma W+B}}}}{\tanh\paren*{k B}}
\end{align*}
Note that: $\tanh\paren*{k\paren*{W\max\curl*{\norm*{x}_p,\gamma}-\gamma W+B}}\leq 1$ and $1-\abs*{1-2\eta(x)}=2\min\curl*{\eta(x),1-\eta(x)}$. Thus the inequality can be further relaxed as follows:
\begin{align*}
    \sR_{\ell_{\gamma}}(h)
     \leq \frac{1+2\beta}{4\beta}\frac{\sR_{\wt{\Phi}_{\mathrm{sig}}}(h)}{\tanh\paren*{k B}}-\paren*{\frac{1+2\beta}{2\beta \tanh(kB)}-1}\,\mathbb{E}_{X}\bracket*{\min\curl*{\eta(x),1-\eta(x)}}.
\end{align*}
When $B=\plus \infty$, it can be equivalently written as follows:
\begin{align}
\label{eq:sig-lin-est-adv-2}
    \sR_{\ell_{\gamma}}(h)
     \leq \frac{1+2\beta}{4\beta}\,\sR_{\wt{\Phi}_{\mathrm{sig}}}(h)-\frac{1}{2\beta}\,\mathbb{E}_{X}\bracket*{\min\curl*{\eta(x),1-\eta(x)}}.
\end{align}

\subsection{One-Hidden-Layer ReLU Neural Networks}
\label{app:derivation-NN-adv_noise}

By the definition of $\sH_{\mathrm{NN}}$, for any $x \in \sX$, 
\begin{align*}
&\uv h_\gamma(x)=\inf_{x'\colon \|x-x'\|_p\leq \gamma}\sum_{j = 1}^n u_j(w_j \cdot x'+b)_{+}\\
&\ov h_\gamma(x)=\sup_{x'\colon \|x-x'\|_p\leq \gamma}\sum_{j = 1}^n u_j(w_j \cdot x'+b)_{+} 
\end{align*}
Note $\sH_{\mathrm{NN}}$ is symmetric. For any $x\in \sX$, there exist $u=\paren*{\frac{1}{\Lambda},\ldots,\frac{1}{\Lambda}}$, $w=0$ and any $0<b\leq B$ satisfy that $\uv h_\gamma(x)>0$. Thus by Lemma~\ref{lemma:explicit_assumption_01_adv}, for any $x\in \sX$, $\sC^*_{\ell_{\gamma},\sH_{\mathrm{NN}}}(x) =\min\curl*{\eta(x), 1 - \eta(x)}$. The $\paren*{\ell_{\gamma},\sH_{\mathrm{NN}}}$-minimizability gap can be expressed as follows:
\begin{align*}
\sM_{\ell_{\gamma},\sH_{\mathrm{NN}}}
& = \sR_{\ell_{\gamma},\sH_{\mathrm{NN}}}^*-\mathbb{E}_{X}\bracket*{\min\curl*{\eta(x),1-\eta(x)}}.
\end{align*}

\subsubsection{Supremum-Based Hinge Loss}
For the supremum-based hinge loss 
\begin{align*}
\wt{\Phi}_{\mathrm{hinge}}\colon=\sup_{x'\colon \|x-x'\|_p\leq \gamma}\Phi_{\mathrm{hinge}}(y h(x')),  \quad \text{where } \Phi_{\mathrm{hinge}}(\alpha)=\max\curl*{0,1 - \alpha},  
\end{align*}
for all $h\in \sH_{\mathrm{NN}}$ and $x\in \sX$:
\small
\begin{align*}
\sC_{\wt{\Phi}_{\mathrm{hinge}}}(h,x,t) 
&=t \wt{\Phi}_{\mathrm{hinge}}(h(x))+(1-t)\wt{\Phi}_{\mathrm{hinge}}(-h(x))\\
& = t\Phi_{\mathrm{hinge}}\paren*{\uv h_\gamma(x)}+(1-t)\Phi_{\mathrm{hinge}}\paren*{-\ov h_\gamma(x)}\\
& =t\max\curl*{0,1-\uv h_\gamma(x)} +(1-t)\max\curl*{0,1+\ov h_\gamma(x)}\\
& \geq \bracket*{ t\max\curl*{0,1-\ov h_\gamma(x)} +(1-t)\max\curl*{0,1+\ov h_\gamma(x)} }\wedge \bracket*{ t\max\curl*{0,1-\uv h_\gamma(x)} +(1-t)\max\curl*{0,1+\uv h_\gamma(x)}} \\
\inf_{h\in\sH_{\mathrm{NN}}}\sC_{\wt{\Phi}_{\mathrm{hinge}}}(h,x,t)
& \geq \inf_{h\in\sH_{\mathrm{NN}}}\bracket*{ t\max\curl*{0,1-\ov h_\gamma(x)} +(1-t)\max\curl*{0,1+\ov h_\gamma(x)} }\wedge \inf_{h\in\sH_{\mathrm{NN}}}\bracket*{ t\max\curl*{0,1-\uv h_\gamma(x)} +(1-t)\max\curl*{0,1+\uv h_\gamma(x)}} \\
& = 1-\abs*{2t-1}\min\curl*{\sup_{h\in\sH_{\mathrm{NN}}}\uv h_\gamma(x),1}\\
\inf_{h\in\sH_{\mathrm{NN}}}\sC_{\wt{\Phi}_{\mathrm{hinge}}}(h,x,t)
& \leq \inf_{h\in\sH_{\mathrm{NN}:w=0}}\sC_{\wt{\Phi}_{\mathrm{hinge}}}(h,x,t)\\
& = 1-\abs*{2t-1}\min\curl*{\Lambda B,1}\\
\sM_{\wt{\Phi}_{\mathrm{hinge}},\sH_{\mathrm{NN}}} 
& = \sR_{\wt{\Phi}_{\mathrm{hinge}},\sH_{\mathrm{NN}}}^* - \mathbb{E}\bracket*{\inf_{h\in\sH_{\mathrm{NN}}}\sC_{\wt{\Phi}_{\mathrm{hinge}}}(h,x,\eta(x))}\\
& \leq \sR_{\wt{\Phi}_{\mathrm{hinge}},\sH_{\mathrm{NN}}}^* - \mathbb{E}\bracket*{1-\abs*{2\eta(x)-1}\min\curl*{\sup_{h\in\sH_{\mathrm{NN}}}\uv h_\gamma(x),1}}
\end{align*}
\normalsize
Thus, for $\frac{1}2< t\leq1$, we have
\begin{align*}
\inf_{h\in\sH_{\mathrm{NN}}:\uv h_\gamma(x)\leq 0 \leq \ov h_\gamma(x)}\sC_{\wt{\Phi}_{\mathrm{hinge}}}(h,x,t)
& = t+(1-t)\\
& =1\\
\inf_{x\in \sX} \inf_{h\in\sH_{\mathrm{NN}}\colon \uv h_\gamma(x)\leq 0 \leq \ov h_\gamma(x)}\Delta\sC_{\wt{\Phi}_{\mathrm{hinge}},\sH_{\mathrm{NN}}}(h,x,t)
& = \inf_{x\in \sX} \curl*{1-\inf_{h\in\sH_{\mathrm{NN}}}\sC_{\wt{\Phi}_{\mathrm{hinge}}}(h,x,t)}\\
&\geq\inf_{x\in \sX}\paren*{2t-1}\min\curl*{\Lambda B,1}\\
&=\paren*{2t-1}\min\curl*{\Lambda B,1}\\
&=\sT_1(t),
\end{align*}
where $\sT_1$ is the increasing and convex function on $\bracket*{0,1}$ defined by
\begin{align*}
\sT_1(t) = \begin{cases}
\min\curl*{\Lambda B,1} \frac{4\beta}{1+2\beta} \, t, & t\in\bracket*{0,1/2+\beta},\\
\min\curl*{\Lambda B,1} \, (2t-1),  & t \in \bracket*{1/2+\beta,1}.
\end{cases}   
\end{align*}
\begin{align*}
\inf_{h\in\sH_{\mathrm{NN}}:\ov h_\gamma(x)<0}\sC_{\wt{\Phi}_{\mathrm{hinge}}}(h,x,t)
& \geq \inf_{h\in\sH_{\mathrm{NN}}:\ov h_\gamma(x)<0}\, [t\max\curl*{0,1-\ov h_\gamma(x)}t +(1-t)\max\curl*{0,1+\ov h_\gamma(x)}]\\
& = t\max\curl*{0,1-0}+(1-t)\max\curl*{0,1+0}\\
& =1\\
\inf_{x\in \sX} \inf_{h\in\sH_{\mathrm{NN}}\colon \ov h_\gamma(x)< 0}\Delta\sC_{\wt{\Phi}_{\mathrm{hinge}},\sH_{\mathrm{NN}}}(h,x,t)
& = \inf_{x\in \sX} \curl*{\inf_{h\in\sH_{\mathrm{NN}}:\ov h_\gamma(x)< 0}\sC_{\wt{\Phi}_{\mathrm{hinge}}}(h,x,t)-\inf_{h\in\sH_{\mathrm{NN}}}\sC_{\wt{\Phi}_{\mathrm{hinge}}}(h,x,t)}\\
&\geq\inf_{x\in \sX}\paren*{2t-1}\min\curl*{\Lambda B,1}\\
&=\paren*{2t-1}\min\curl*{\Lambda B,1}\\
&=\sT_2(2t - 1),
\end{align*}
where $\sT_2$ is the increasing and convex function on $\bracket*{0,1}$ defined by
\begin{align*}
\forall t \in [0,1], \quad \sT_2(t) = \min\curl*{\Lambda B,1} \, t \,;
\end{align*}
By Proposition~\ref{prop-adv-noise}, for $\epsilon= 0$, the modified adversarial $\sH_{\mathrm{NN}}$-estimation error transformation of the supremum-based hinge loss under Massart's noise condition with $\beta$ is lower bounded as follows:
\begin{align*}
\sT^M_{\wt{\Phi}_{\mathrm{hinge}}}\geq\wt{\sT}^M_{\wt{\Phi}_{\mathrm{hinge}}}:=\min\curl*{\sT_1,\sT_2}=\begin{cases}
\min\curl*{\Lambda B,1} \, (2t-1), & t \in \bracket*{1/2+\beta,1},\\ \min\curl*{\Lambda B,1}\frac{4\beta}{1+2\beta}\, t, & t\in \left[0,1/2+\beta\right).
\end{cases}
\end{align*}
Note $\wt{\sT}^M_{\wt{\Phi}_{\mathrm{hinge}}}$ is convex, non-decreasing, invertible and satisfies that $\wt{\sT}^M_{\wt{\Phi}_{\mathrm{hinge}}}(0)=0$. By Proposition~\ref{prop-adv-noise}, using the fact that
$\wt{\sT}^M_{\wt{\Phi}_{\mathrm{hinge}}}\geq \min\curl*{\Lambda B,1}\frac{4\beta}{1+2\beta}\, t$ yields the adversarial $\sH_{\mathrm{NN}}$-consistency estimation error bound for the supremum-based hinge loss, valid for all $h \in \sH_{\mathrm{NN}}$ and distributions $\sD$ satisfies Massart's noise condition with $\beta$
\begin{align}
\label{eq:hinge-NN-est-adv}
     \sR_{\ell_{\gamma}}(h)- \sR_{\ell_{\gamma},\sH_{\mathrm{NN}}}^* 
     \leq 
      \frac{1+2\beta}{4\beta}\frac{\sR_{\wt{\Phi}_{\mathrm{hinge}}}(h)-\sR_{\wt{\Phi}_{\mathrm{hinge}},\sH_{\mathrm{NN}}}^*+\sM_{\wt{\Phi}_{\mathrm{hinge}},\sH_{\mathrm{NN}}}}{\min\curl*{\Lambda B,1}}-\sM_{\ell_{\gamma}, \sH_{\mathrm{NN}}}
\end{align}
Since
\begin{align*}
\sM_{\ell_{\gamma},\sH_{\mathrm{NN}}}
& = \sR_{\ell_{\gamma},\sH_{\mathrm{NN}}}^*-\mathbb{E}_{X}\bracket*{\min\curl*{\eta(x),1-\eta(x)}},\\
\sM_{\wt{\Phi}_{\mathrm{hinge}},\sH_{\mathrm{NN}}} 
& \leq \sR_{\wt{\Phi}_{\mathrm{hinge}},\sH_{\mathrm{NN}}}^* - \mathbb{E}\bracket*{1-\abs*{2\eta(x)-1}\min\curl*{\sup_{h\in\sH_{\mathrm{NN}}}\uv h_\gamma(x),1}},
\end{align*}
the inequality can be relaxed as follows:
\begin{align*}
     \sR_{\ell_{\gamma}}(h) 
     \leq 
      \frac{1+2\beta}{4\beta}\frac{\sR_{\wt{\Phi}_{\mathrm{hinge}}}(h)}{\min\curl*{\Lambda B,1}}+\mathbb{E}_{X}\bracket*{\min\curl*{\eta(x),1-\eta(x)}}-\frac{1+2\beta}{4\beta}\frac{\mathbb{E}\bracket*{1-\abs*{2\eta(x)-1}\min\curl*{\sup_{h\in\sH_{\mathrm{NN}}}\uv h_\gamma(x),1}}}{\min\curl*{\Lambda B,1}}
\end{align*}
Observe that
\begin{align*}
\sup_{h\in\sH_{\mathrm{NN}}}\uv h_\gamma(x)
&=
\sup_{ \|u \|_{1}\leq \Lambda,~\|w_j\|_q\leq W,~\abs*{b}\leq B}\inf_{x'\colon \|x-x'\|_p\leq \gamma}\sum_{j = 1}^n u_j(w_j \cdot x'+b)_{+}\\
& \leq \inf_{x'\colon \|x-x'\|_p\leq \gamma} \sup_{ \|u \|_{1}\leq \Lambda,~\|w_j\|_q\leq W,~\abs*{b}\leq B} \sum_{j = 1}^n u_j(w_j \cdot x'+b)_{+} \\
& = \inf_{x'\colon \|x-x'\|_p\leq \gamma} \Lambda\paren*{W\norm*{x'}_p+B}\\
& =
\begin{cases}
\Lambda\paren*{W\norm*{x}_p-\gamma W + B} & \text{if } \norm*{x}_p \geq \gamma\\
\Lambda B & \text{if } \norm*{x}_p < \gamma
\end{cases}\\
& = \Lambda\paren*{W\max \curl*{\norm*{x}_p,\gamma}-\gamma W + B}.
\end{align*}
Thus, the inequality can be further relaxed as follows:
\begin{align*}
     \sR_{\ell_{\gamma}}(h) 
     \leq 
     \frac{1+2\beta}{4\beta}\frac{\sR_{\wt{\Phi}_{\mathrm{hinge}}}(h)}{\min\curl*{\Lambda B,1}}+\mathbb{E}_{X}\bracket*{\min\curl*{\eta(x),1-\eta(x)}}-\frac{1+2\beta}{4\beta}\frac{\mathbb{E}\bracket*{1-\abs*{2\eta(x)-1}\min\curl*{ \Lambda\paren*{W\max \curl*{\norm*{x}_p,\gamma}-\gamma W + B},1}}}{\min\curl*{\Lambda B,1}}
\end{align*}
Note that: $\min\curl*{ \Lambda\paren*{W\max \curl*{\norm*{x}_p,\gamma}-\gamma W + B},1}\leq 1$ and $1-\abs*{1-2\eta(x)}=2\min\curl*{\eta(x),1-\eta(x)}$. Thus the  inequality can be further relaxed as follows:
\begin{align}
\label{eq:hinge-NN-est-adv-2}
     \sR_{\ell_{\gamma}}(h)
     \leq \frac{1+2\beta}{4\beta}\frac{\sR_{\wt{\Phi}_{\mathrm{hinge}}}(h)}{ \min\curl*{\Lambda B,1}}-\paren*{\frac{1+2\beta}{2\beta \min\curl*{\Lambda B,1}}-1}\,\mathbb{E}_{X}\bracket*{\min\curl*{\eta(x),1-\eta(x)}}.
\end{align}
When $\Lambda B\geq 1$, it can be equivalently written as follows:
\begin{align*}
     \sR_{\ell_{\gamma}}(h)
     \leq \frac{1+2\beta}{4\beta}\,\sR_{\wt{\Phi}_{\mathrm{hinge}}}(h)-\frac{1}{2\beta }\,\mathbb{E}_{X}\bracket*{\min\curl*{\eta(x),1-\eta(x)}}.
\end{align*}

\subsubsection{Supremum-Based Sigmoid Loss}
For the supremum-based sigmoid loss 
\begin{align*}
\wt{\Phi}_{\mathrm{sig}}\colon=\sup_{x'\colon \|x-x'\|_p\leq \gamma}\Phi_{\mathrm{sig}}(y h(x')),  \quad \text{where } \Phi_{\mathrm{sig}}(\alpha)=1-\tanh(k\alpha),~k>0,   
\end{align*}
for all $h\in \sH_{\mathrm{NN}}$ and $x\in \sX$:
\begin{align*}
\sC_{\wt{\Phi}_{\mathrm{sig}}}(h,x,t) 
&=t \wt{\Phi}_{\mathrm{sig}}(h(x))+(1-t)\wt{\Phi}_{\mathrm{sig}}(-h(x))\\
& = t\Phi_{\mathrm{sig}}\paren*{\uv h_\gamma(x)}+(1-t)\Phi_{\mathrm{sig}}\paren*{-\ov h_\gamma(x)}\\
& =t\paren*{1-\tanh(k \uv h_\gamma(x))} +(1-t)\paren*{1+\tanh(k \ov h_\gamma(x))}\\
& \geq \max\curl*{1+\paren{1-2t}\tanh(k\ov h_\gamma(x)), 1+\paren{1-2t}\tanh(k\uv h_\gamma(x))}\\
\inf_{h\in\sH_{\mathrm{NN}}}\sC_{\wt{\Phi}_{\mathrm{sig}}}(h,x,t)
& \geq \max\curl*{\inf_{h\in\sH_{\mathrm{NN}}}\bracket*{1+\paren{1-2t}\tanh(k\ov h_\gamma(x))}, \inf_{h\in\sH_{\mathrm{NN}}}\bracket*{1+\paren{1-2t}\tanh(k\uv h_\gamma(x))}}\\
& = 1 - \abs*{1-2t}\tanh\paren*{k\sup_{h\in\sH_{\mathrm{NN}}}\uv h_\gamma(x)}\\
\inf_{h\in\sH_{\mathrm{NN}}}\sC_{\wt{\Phi}_{\mathrm{sig}}}(h,x,t)
& \leq \max\curl*{t,1-t}\paren*{1-\tanh\paren*{k \Lambda B }}+\min\curl*{t,1-t}\paren*{1+\tanh\paren*{k \Lambda B}}\\
& = 1 - \abs*{1-2t}\tanh\paren*{k \Lambda B}\\
\sM_{\wt{\Phi}_{\mathrm{sig}},\sH_{\mathrm{NN}}} 
& = \sR_{\wt{\Phi}_{\mathrm{sig}},\sH_{\mathrm{NN}}}^* - \mathbb{E}\bracket*{\inf_{h\in\sH_{\mathrm{NN}}}\sC_{\wt{\Phi}_{\mathrm{sig}}}(h,x,\eta(x))}\\
& \leq \sR_{\wt{\Phi}_{\mathrm{sig}},\sH_{\mathrm{NN}}}^* - \mathbb{E}\bracket*{1 - \abs*{1-2\eta(x)}\tanh\paren*{k\sup_{h\in\sH_{\mathrm{NN}}}\uv h_\gamma(x)}}
\end{align*}
For $\frac{1}2< t\leq1$, we have
\begin{align*}
\inf_{h\in\sH_{\mathrm{NN}}:\uv h_\gamma(x)\leq 0 \leq \ov h_\gamma(x)}\sC_{\wt{\Phi}_{\mathrm{sig}}}(h,x,t) 
& = t+(1-t)\\
& =1\\
\inf_{x\in \sX} \inf_{h\in\sH_{\mathrm{NN}}\colon \uv h_\gamma(x)\leq 0 \leq \ov h_\gamma(x)}\Delta\sC_{\wt{\Phi}_{\mathrm{sig}},\sH_{\mathrm{NN}}}(h,x,t)
& = \inf_{x\in \sX} \curl*{1-\inf_{h\in\sH_{\mathrm{NN}}}\sC_{\wt{\Phi}_{\mathrm{sig}}}(h,x,t)}\\
&\geq \inf_{x\in \sX} (2t-1)\tanh\paren*{k \Lambda B}\\
&=(2t-1)\tanh\paren*{k \Lambda B}\\
&=\sT_1(t),
\end{align*}
where $\sT_1$ is the increasing and convex function on $\bracket*{0,1}$ defined by
\begin{align*}
\sT_1(t) = \begin{cases}
\tanh\paren*{k \Lambda B} \frac{4\beta}{1+2\beta} \, t, & t\in\bracket*{0,1/2+\beta},\\
\tanh\paren*{k \Lambda B} \, (2t-1),  & t \in \bracket*{1/2+\beta,1}.
\end{cases}    
\end{align*}
\begin{align*}
\inf_{h\in\sH_{\mathrm{NN}}:\ov h_\gamma(x)<0}\sC_{\wt{\Phi}_{\mathrm{sig}}}(h,x,t)
& \geq \inf_{h\in\sH_{\mathrm{NN}}:\ov h_\gamma(x)<0} 1+\paren{1-2t}\tanh(k\ov h_\gamma(x))\\
& = 1\\
\inf_{x\in \sX} \inf_{h\in\sH_{\mathrm{NN}}\colon \ov h_\gamma(x)< 0}\Delta\sC_{\wt{\Phi}_{\mathrm{sig}},\sH_{\mathrm{NN}}}(h,x)
& = \inf_{x\in \sX} \curl*{\inf_{h\in\sH_{\mathrm{NN}}:\ov h_\gamma(x)< 0}\sC_{\wt{\Phi}_{\mathrm{sig}}}(h,x,t)-\inf_{h\in\sH_{\mathrm{NN}}}\sC_{\wt{\Phi}_{\mathrm{sig}}}(h,x,t)}\\
&\geq \inf_{x\in \sX}(2t-1)\tanh\paren*{k \Lambda B}\\
&=(2t-1)\tanh\paren*{k \Lambda B}\\
&=\sT_2(2t-1),
\end{align*}
where $\sT_2$ is the increasing and convex function on $\bracket*{0,1}$ defined by
\begin{align*}
\forall t \in [0,1], \quad \sT_2(t) = \tanh\paren*{k \Lambda B} \, t \,;
\end{align*}
By Proposition~\ref{prop-adv-noise}, for $\epsilon= 0$, the modified adversarial $\sH_{\mathrm{NN}}$-estimation error transformation of the supremum-based sigmoid loss under Massart's noise condition with $\beta$ is lower bounded as follows:
\begin{align*}
\sT^M_{\wt{\Phi}_{\mathrm{sig}}}\geq \wt{\sT}^M_{\wt{\Phi}_{\mathrm{sig}}}=\min\curl*{\sT_1,\sT_2}=\begin{cases}
\tanh\paren*{k \Lambda B} \frac{4\beta}{1+2\beta} \, t, & t\in\bracket*{0,1/2+\beta},\\
\tanh\paren*{k \Lambda B} \, (2t-1),  & t \in \bracket*{1/2+\beta,1}.
\end{cases}    
\end{align*}
Note $\wt{\sT}^M_{\wt{\Phi}_{\mathrm{sig}}}$ is convex, non-decreasing, invertible and satisfies that $\wt{\sT}^M_{\wt{\Phi}_{\mathrm{sig}}}(0)=0$. By Proposition~\ref{prop-adv-noise}, using the fact that
$\wt{\sT}^M_{\wt{\Phi}_{\mathrm{sig}}}\geq \tanh\paren*{k \Lambda B} \frac{4\beta}{1+2\beta} \, t$ yields the adversarial $\sH_{\mathrm{NN}}$-consistency estimation error bound for the supremum-based sigmoid loss, valid for all $h \in \sH_{\mathrm{NN}}$ and distributions $\sD$ satisfies Massart's noise condition with $\beta$:
\begin{align}
\label{eq:sig-NN-est-adv}
     \sR_{\ell_{\gamma}}(h)- \sR_{\ell_{\gamma},\sH_{\mathrm{NN}}}^* 
     \leq 
     \frac{1+2\beta}{4\beta}\frac{\sR_{\wt{\Phi}_{\mathrm{sig}}}(h)-\sR_{\wt{\Phi}_{\mathrm{sig}},\sH_{\mathrm{NN}}}^*+\sM_{\wt{\Phi}_{\mathrm{sig}},\sH_{\mathrm{NN}}}}{\tanh\paren*{k \Lambda B}}-\sM_{\ell_{\gamma}, \sH_{\mathrm{NN}}}
\end{align}
Since
\begin{align*}
\sM_{\ell_{\gamma},\sH_{\mathrm{NN}}}
& = \sR_{\ell_{\gamma},\sH_{\mathrm{NN}}}^*-\mathbb{E}_{X}\bracket*{\min\curl*{\eta(x),1-\eta(x)}},\\
\sM_{\wt{\Phi}_{\mathrm{sig}},\sH_{\mathrm{NN}}} 
& \leq \sR_{\wt{\Phi}_{\mathrm{sig}},\sH_{\mathrm{NN}}}^* - \mathbb{E}\bracket*{1 - \abs*{1-2\eta(x)}\tanh\paren*{k\sup_{h\in\sH_{\mathrm{NN}}}\uv h_\gamma(x)}},
\end{align*}
the inequality can be relaxed as follows:
\begin{align*}
     \sR_{\ell_{\gamma}}(h) 
     \leq 
     \frac{1+2\beta}{4\beta}\frac{\sR_{\wt{\Phi}_{\mathrm{sig}}}(h)}{\tanh\paren*{k \Lambda B}}+\mathbb{E}_{X}\bracket*{\min\curl*{\eta(x),1-\eta(x)}}-\frac{1+2\beta}{4\beta}\frac{\mathbb{E}\bracket*{1 - \abs*{1-2\eta(x)}\tanh\paren*{k\sup_{h\in\sH_{\mathrm{NN}}}\uv h_\gamma(x)}}}{\tanh\paren*{k \Lambda B}}\
\end{align*}
Observe that
\begin{align*}
\sup_{h\in\sH_{\mathrm{NN}}}\uv h_\gamma(x)
&=
\sup_{ \|u \|_{1}\leq \Lambda,~\|w_j\|_q\leq W,~\abs*{b}\leq B}\inf_{x'\colon \|x-x'\|_p\leq \gamma}\sum_{j = 1}^n u_j(w_j \cdot x'+b)_{+}\\
& \leq \inf_{x'\colon \|x-x'\|_p\leq \gamma} \sup_{ \|u \|_{1}\leq \Lambda,~\|w_j\|_q\leq W,~\abs*{b}\leq B} \sum_{j = 1}^n u_j(w_j \cdot x'+b)_{+} \\
& = \inf_{x'\colon \|x-x'\|_p\leq \gamma} \Lambda\paren*{W\norm*{x'}_p+B}\\
& =
\begin{cases}
\Lambda\paren*{W\norm*{x}_p-\gamma W + B} & \text{if } \norm*{x}_p \geq \gamma\\
\Lambda B & \text{if } \norm*{x}_p < \gamma
\end{cases}\\
& = \Lambda\paren*{W\max \curl*{\norm*{x}_p,\gamma}-\gamma W + B}.
\end{align*}
Thus, the inequality can be further relaxed as follows:
\begin{align*}
     \sR_{\ell_{\gamma}}(h) 
     \leq 
     \frac{1+2\beta}{4\beta}\frac{\sR_{\wt{\Phi}_{\mathrm{sig}}}(h)}{\tanh\paren*{k \Lambda B}}+\mathbb{E}_{X}\bracket*{\min\curl*{\eta(x),1-\eta(x)}}-\frac{1+2\beta}{4\beta}\frac{\mathbb{E}\bracket*{1 - \abs*{1-2\eta(x)}\tanh\paren*{k\Lambda\paren*{W\max \curl*{\norm*{x}_p,\gamma}-\gamma W + B}}}}{\tanh\paren*{k \Lambda B}}
\end{align*}
Note that: $\tanh\paren*{k\Lambda\paren*{W\max \curl*{\norm*{x}_p,\gamma}-\gamma W + B}}\leq 1$ and $1-\abs*{1-2\eta(x)}=2\min\curl*{\eta(x),1-\eta(x)}$. Thus the inequality can be further relaxed as follows:
\begin{align}
\label{eq:sig-NN-est-adv-2}
     \sR_{\ell_{\gamma}}(h)
     \leq \frac{1+2\beta}{4\beta}\frac{\sR_{\wt{\Phi}_{\mathrm{sig}}}(h)}{\tanh\paren*{k\Lambda B}}-\paren*{\frac{1+2\beta}{2\beta \tanh(k\Lambda B)}-1}\,\mathbb{E}_{X}\bracket*{\min\curl*{\eta(x),1-\eta(x)}}.
\end{align}
When $\Lambda B=\plus \infty$, it can be equivalently written as follows:
\begin{align*}
     \sR_{\ell_{\gamma}}(h)
     \leq \frac{1+2\beta}{4\beta}\,\sR_{\wt{\Phi}_{\mathrm{sig}}}(h)-\frac{1}{2\beta}\,\mathbb{E}_{X}\bracket*{\min\curl*{\eta(x),1-\eta(x)}}.
\end{align*}

\end{document}